\ifdefined\useorstyle

\documentclass[opre,accepted]{informs3}

\DoubleSpacedXI % Made default 4/4/2014 at request
\usepackage{endnotes}
\let\footnote=\endnote

\usepackage{natbib}
 \bibpunct[, ]{(}{)}{,}{a}{}{,}%
 \def\bibsep{\smallskipamount}%
\TheoremsNumberedThrough     % Preferred (Theorem 1, Lemma 1, Theorem 2)
\ECRepeatTheorems

\EquationsNumberedThrough    % Default: (1), (2), ...
\MANUSCRIPTNO{OPRE-2020-07-444.R1}
\usepackage[utf8]{inputenc} % allow utf-8 input
\usepackage[T1]{fontenc}    % use 8-bit T1 fonts
\usepackage{xcolor}
\definecolor{darkblue}{rgb}{0,0,.5}
\usepackage[colorlinks=true,allcolors=darkblue]{hyperref}       % hyperlinks
\usepackage{url}            % simple URL typesetting
\usepackage{booktabs}       % professional-quality tables
\usepackage{amsfonts}       % blackboard math symbols
\usepackage{nicefrac}       % compact symbols for 1/2, etc.
\usepackage{microtype}      % microtypography
\usepackage{todonotes}
\usepackage{xfrac}

\usepackage{subcaption}
\usepackage{color}
\usepackage{optidef}
\usepackage{listings}

\newcommand{\red}[1]{\textcolor{red}{#1}}
\definecolor{deepmagenta}{rgb}{0.8, 0.0, 0.8}

\definecolor{innerboxcolor}{rgb}{.9,.95,1}
\definecolor{outerlinecolor}{rgb}{.6,0,.2}

\newcommand{\jd}[1]{\todo[linecolor=blue,backgroundcolor=blue!25,bordercolor=blue]{#1}}
\newcommand{\jdi}[1]{\todo[inline,linecolor=blue,backgroundcolor=blue!25,bordercolor=blue]{#1}}

\newcommand{\doin}{\mbox{\normalfont do}}

\usepackage{statistics-macros-or}

\renewcommand{\theassumption}{A\arabic{assumption}}

\begin{document}
%%%%%%%%%%%%%%%%

%\setcounter{page}{1}
%\DOI{10.1287/xxxx.0000.0000}%

\RUNAUTHOR{Duchi, Hashimoto and Namkoong}

\RUNTITLE{Distributionally Robust Losses for Latent Covariate Mixtures}

\TITLE{Distributionally Robust Losses for Latent Covariate Mixtures}
%\TITLE{}

  \ARTICLEAUTHORS{%
    \AUTHOR{John C. Duchi} \AFF{Departments of Electrical Engineering and
      Statistics, Stanford University, \EMAIL{jduchi@stanford.edu},
      \URL{}}

    \AUTHOR{Tatsunori Hashimoto} \AFF{Department of Computer Science,
      Stanford University, \EMAIL{thashim@stanford.edu}, \URL{}}

    \AUTHOR{Hongseok Namkoong} \AFF{Decision, Risk, and Operations Division,
      Columbia Business School, \EMAIL{namkoong@gsb.columbia.edu}, \URL{}}
} % end of the block

\ABSTRACT{%
  % We develop a method for controlling the worst-case loss over unknown, latent groups in a heterogeneous dataset, where the groups may represent minority demographics or difficult subsets of the data. Casting this problem as a minimax estimation problem, we obtain a simple, convex surrogate loss which upper bounds the worst case objective, and show that this surrogate is asymptotically optimal.
% 
% \jdi{Make more active, alive} Statistical models are often \jd{Passive}
% trained on heterogeneous data, such as text from multiple corpora, or data
% from multiple demographic groups. A basic goal for such models is to ensure
% uniformly high accuracy across subpopulations in the dataset.  We derive a
% loss function which accounts for heterogeneous data by studying worst-case
% losses over all subpopulations of a certain size. Our proposed loss is a
% computationally tractable convex loss with a finite-sample guarantee of
% performance, and converges at the nonparametric rate\jd{A little ???y
% (OCR)}. Empirically, we observe on lexical similarity and recidivism
% prediction tasks that our worst-case procedure learns models that do well
% against unseen subpopulations.

%Modern large-scale datasets are often collected from heterogenous populations
%such as multiple demographic groups, or text from multiple corpora. For models
%trained on heterogeneous data, ensuring uniformly high accuracy across
%subpopulations is a basic goal.

While modern large-scale datasets often consist of heterogeneous
subpopulations---for example, multiple demographic groups or multiple text
corpora---the standard practice of minimizing average loss fails to
guarantee uniformly low losses across all subpopulations.  We propose a
convex procedure that controls the worst-case performance over all
subpopulations of a given size. Our procedure comes with finite-sample
(nonparametric) convergence guarantees on the worst-off
subpopulation. Empirically, we observe on lexical similarity, wine quality,
and recidivism prediction tasks that our worst-case procedure learns models
that do well against unseen subpopulations.

%%% Local Variables:
%%% TeX-master: "main.tex"
%%% End:

}%

%\KEYWORDS{deterministic inventory theory; infinite linear programming duality; 
%  existence of optimal policies; semi-Markov decision process; cyclic schedule}
%\MSCCLASS{Primary: 90B05; secondary: 90C40, 90C90}
%\ORMSCLASS{Primary: Inventory/production: deterministic multi-item;
%  secondary: dynamic programming/optimal control: deterministic 
%  semi-Markov; programming: infinite dimensional}
%\HISTORY{Received November 20, 2003; revised March 8, 2004, and March 26, 2004.}

 \MSCCLASS{} \ORMSCLASS{Primary: Statistics ; secondary: Programming:Stochastic}
\HISTORY{}

\maketitle
%%%%%%%%%%%%%%%%%%%%%%%%%%%%%%%%%%%%%%%%%%%%%%%%%%%%%%%%%%%%%%%%%%%%%%
% Samples of sectioning (and labeling) in MOOR.
% NOTE: (1) all section levels end with a period,
%       (2) capitalization is as shown (sentence style, not title style).
%
% \section{Introduction.}\label{intro} %%1.
%\subsection{Duality and the classical EOQ problem.}\label{class-EOQ} %% 1.1.
%\subsection{Outline.}\label{outline1} %% 1.2.
%\subsubsection{Cyclic schedules for the general deterministic SMDP.}
%  \label{cyclic-schedules} %% 1.2.1
%\section{Problem description.}\label{problemdescription} %% 2.

\else

\documentclass[11pt]{article}
\usepackage[numbers]{natbib}
\usepackage[top=2.54cm,bottom=2.54cm,left=3cm,right=3cm]{geometry}

%\pdfminorversion=4
% NOTE: To produce blinded version, replace "0" with "1" below.
%

\usepackage[utf8]{inputenc} % allow utf-8 input
\usepackage[T1]{fontenc}    % use 8-bit T1 fonts
\usepackage{xcolor}
\definecolor{darkblue}{rgb}{0,0,.5}
\usepackage[colorlinks=true,allcolors=darkblue]{hyperref}       % hyperlinks
\usepackage{url}            % simple URL typesetting
\usepackage{booktabs}       % professional-quality tables
\usepackage{amsfonts}       % blackboard math symbols
\usepackage{nicefrac}       % compact symbols for 1/2, etc.
\usepackage{microtype}      % microtypography
\usepackage{todonotes}
\usepackage{xfrac}

\usepackage{subcaption}
\usepackage{setspace}
\captionsetup[sub]{font={small, stretch=1.0}}

\usepackage{color}
\usepackage{optidef}
\usepackage{listings}

\newcommand{\red}[1]{\textcolor{red}{#1}}
\definecolor{innerboxcolor}{rgb}{.9,.95,1}
\definecolor{outerlinecolor}{rgb}{.6,0,.2}
\newcommand{\hno}[1]{\fcolorbox{outerlinecolor}{innerboxcolor}{
    \begin{minipage}{.9\textwidth}
      \red{\bf Hong:} {#1}
  \end{minipage}} \\
}

\newcommand{\jd}[1]{\todo[linecolor=blue,backgroundcolor=blue!25,bordercolor=blue]{#1}}
\newcommand{\jdi}[1]{\todo[inline,linecolor=blue,backgroundcolor=blue!25,bordercolor=blue]{#1}}

\newcommand{\doin}{\mbox{\normalfont do}}

\usepackage{statistics-macros}

\renewcommand{\theassumption}{A\arabic{assumption}}

\begin{document}

% ------------------------------------------------------------------------
% Default title and authorship
% ------------------------------------------------------------------------
\begin{center}
  {\huge  Distributionally Robust Losses \\ \vspace{.3cm}  for Latent Covariate Mixtures} \\
  \vspace{.5cm}
  {\Large John C.\ Duchi$^{1}$ ~~ Tatsunori Hashimoto$^2$ ~~ Hongseok Namkoong$^3$} \\
  \vspace{.2cm}
  Departments of $^1$Statistics,
  Electrical Engineering, and $^2$Computer Science, Stanford University \\
   $^3$Decision, Risk, and Operations Division, Columbia Business School \\
  \vspace{.2cm}
  \texttt{\{jduchi,thashim\}@stanford.edu, namkoong@gsb.columbia.edu}
\end{center}

\begin{abstract}
  
\end{abstract}
\fi

%\maketitle

% \todo{Note - should we be including the RKHS stuff? if so, how much of it?}

\newcommand{\latent}{Z_i}
\newcommand{\latdist}{\text{Bernoulli}(\alpha)}
\newcommand{\xdom}{\mathcal{X}}
\newcommand{\ydom}{\mathcal{Y}}
\newcommand{\confmixdist}{\mathcal{P}_{\alpha_0,(X,C)}}
\newcommand{\mixdist}{\mathcal{P}_{\alpha_0,X}}
\newcommand{\mixdistq}{\mathcal{P}_{\Delta,X,q}}
\newcommand{\jointmixdist}{\mathcal{P}_{\alpha_0,(X,Y)}}
\newcommand{\lossvar}{\loss(\param; (X,Y))}
\newcommand{\condrisk}{\E[\loss(\param; (X,Y)) \mid X]}
\newcommand{\confcondrisk}{\E[\loss(\param; (X,Y)) \mid X, C]}
\newcommand{\partcondrisk}{\E[\loss(\param; (X,X',Y)) \mid X]}
\newcommand{\confthresh}{\delta}
\newcommand{\lipspace}{\mc{H}_L}
\newcommand{\EX}[1]{\underset{#1}{\E}}
\newcommand{\pworst}{Q_0}
\newcommand{\pnuisance}{Q_1}
\newcommand{\pmarg}{P_X}
\newcommand{\pcond}{P_{Y \mid X}}
\newcommand{\pdist}{P}

\newcommand{\empparam}{\what{\theta}^{\rm rob}_{n, \epsilon}}
\newcommand{\lbound}{M}
\newcommand{\empcond}{\what{P}_{n}|x}
\newcommand{\diam}{\mathop{\rm diam}}
\newcommand{\rpos}{\mathbb{R}_+}

\newcommand{\bddspace}{\mc{F}_{\delta, \kdual}}

\newcommand{\kexp}{q}
\newcommand{\kexpinv}{{1/q}}
\newcommand{\kdual}{{p}}
\newcommand{\kdualinv}{{1/p}}

\newcommand{\vopt}{R_\kdual^*}
\newcommand{\empobj}{\what{R}_{\kdual, \epsilon, L}(\theta, \eta, B)}
\newcommand{\empobjshort}{\what{R}_{\kdual, \epsilon, L}(\theta, \eta)}
\newcommand{\empupperbd}{\what{R}_{\kdual, \epsilon, L_n(\gamma)} (\theta, \eta)}
\newcommand{\empobjnoarg}{\what{R}_{\kdual, \epsilon, L}}
\newcommand{\popobj}{R_{\kdual, \epsilon, L}}
\newcommand{\cpopobjone}{R_{1, \kdual, \epsilon, L}}
\newcommand{\cpopobj}{R_{c, \kdual, \epsilon, L}}
\newcommand{\conepopobj}{R_{c_1, \kdual, \epsilon, L}}
\newcommand{\ctwopopobj}{R_{c_2, \kdual, \epsilon, L}}
\newcommand{\confpopobj}{R_{\kdual, \epsilon, L, \delta}}
\newcommand{\confpopobjshort}{R_{\kdual, \epsilon, L, \delta}(\theta, \eta)} 
\newcommand{\confempobj}{\what{R}_{\kdual, \epsilon, L, \delta}(\theta, \eta, B)} 
\newcommand{\confempobjshort}{\what{R}_{\kdual, \epsilon, L, \delta}(\theta, \eta)}
\newcommand{\confempupperbd}{\what{R}_{\kdual, \epsilon, L_n(\gamma), \delta_n(\gamma)} (\theta, \eta)}

\newcommand{\holderspace}{\mc{H}_{L, \alpha}}
\newcommand{\holderbound}{\mc{H}_{L, \kdual}}
\newcommand{\emplipspace}{\what{\mc{H}}_{L}}
\newcommand{\empholderspace}{\what{\mc{H}}_{\alpha, L}}
\newcommand{\empholderbound}{\what{\mc{H}}_{L, \kdual}}

\newcommand{\holderball}[1]{\Lambda^{#1}} % ball with Holder-norm
\newcommand{\holdersmooth}{\beta} % Holder smoothness parameter

\ifdefined\useorstyle
\vspace{-20pt}
\else
\fi
\section{Introduction}
\label{section:intro}

%As statistical models increasingly get applied in high-stakes decision making
%problems, it is imperative that they maintain reliable performance when
%deployed in the wild. Ideally, we want our models to have some guaranteed
%level of performance against interventions on covariates, potentially even in
%presence of confounding. In this work, we are interested in a particular type
%of intervention that yield subpopulations which we call mixture covariate
%shifts. We study worst-case procedures that can provide a uniform level of
%performance against such interventions.

%% \jcdcomment{Throughout let us change notation to
%%   \begin{equation*}
%%     L(\theta \mid X) \defeq \condrisk
%%   \end{equation*}
%%   and do some careful editing to make it have $x$ or $X$.
%% }
  
When we train models over heterogeneous data, a basic goal is to train
models that perform uniformly well across all subpopulations instead of just
on average.  For example, in natural language processing (NLP), large-scale
corpora often consist of data from multiple domains, each domain varying in
difficulty and frequently containing large proportions of easy
examples~\citep{CerDiAgLoLu17}.  Standard approaches
optimize average performance, however, and yield models that accurately
predict easy examples but sacrifice predictive performance on hard
subpopulations~\citep{RajpurkarJiLi18}.

The growing use of machine learning systems in socioeconomic decision-making
problems, such as loan-servicing and recidivism prediction, highlights the
importance of models that perform well over different demographic
groups~\citep{BarocasSe16}. In the face of this need, a number of authors
observe that optimizing average performance often yields models that perform
poorly on minority subpopulations~\citep{AmodeiAnAnBaBaCaCaCaChCh16,
  GrotherQuPh10, HovySo15, BlodgettGrOc16, SapiezynskiKaWi17, Tatman17}.
When datasets contain demographic information, a natural approach is to
optimize worst-case group loss or equalize losses over groups.  But in many
tasks---such as language identification or video
analysis~\citep{Tatman17,BlodgettGrOc16}---privacy concerns preclude
recording demographic or other sensitive information, limiting the
applicability of methods that require knowledge of demographic
identities. For example, lenders in the United States are prohibited from
asking loan applicants for racial information unless it is to demonstrate
compliance with anti-discrimination regulation~\citep{CFPB14,
  ChenKaMaSvUd19}.

To address these challenges, we seek models that perform well on each
subpopulation rather than those that achieve good (average) performance by
focusing on the easy examples and domains.  Thus, in this paper we develop
procedures that control performance over \emph{all} large enough
subpopulations, agnostic to the distribution of each subpopulation.  We
study a worst-case formulation over large enough subpopulations in the data,
providing procedures that automatically focus on the difficult subsets of
the dataset. Our procedure guarantees a uniform level of performance across
subpopulations by hedging against unseen covariate shifts, potentially even
in the presence of confounding.

% Before we formally describe our procedure, we give two concrete examples where
% data heterogeneity arises naturally.

% While controlling the worst-case performance over pre-defined groups is a
% natural approach to ensure uniform performance across subpopulations, the
% groups of interest are often difficult or impossible to identify a
% priori. Instead,

% heterogeneous datasets, uniform performance
% Mixture covariate shift is a natural concept for problems where
% the training data consists of aggregates over different populations or data
% sources, and the goal is to find models which performs uniformly well.
% We motivate mixture covariate shift with two concrete problem settings.

% Ensuring uniform performance across subpopulations is the \emph{mixture
% covariate shift} problem \todo{Maybe use names from shift literature},

In classical statistical learning and prediction problems, we wish to
predict a target $Y \in \mc{Y}$ from a covariate vector $X \in \mc{X}
\subset \R^d$ drawn from an underlying population $(X, Y) \sim P$, measuring
performance of a predictor $\theta$ via the loss $\loss : \paramdomain
\times (\mc{X} \times \mc{Y}) \to \R_+$.  The standard approach is to
minimize the population expectation $\E_P[\loss(\theta; (X, Y))]$.  In
contrast, we consider an elaborated setting in which the observed data comes
from a mixture model, and we evaluate model losses on a component
(subpopulation) from this mixture. More precisely, we assume that
for some mixing proportion $\alpha \in (0, 1)$, the data
$X$ are marginally distributed as
% \begin{equation*}
  $X \sim P_X \defeq \alpha \pworst + (1 - \alpha) \pnuisance$,
% \end{equation*}
while the
subpopulations $\pworst$ and $\pnuisance$ are unknown. The
classical formulation does little to ensure equitable performance for
data $X$ from both $Q_0$ and $Q_1$, especially for small $\alpha$.
Thus for a fixed
conditional distribution $\pcond$,
we instead seek $\theta \in \paramdomain$ that
minimizes the expected loss under the latent subpopulation $\pworst$
\begin{equation}
  \label{eq:latmix}
  \minimize_{\param\in\paramdomain} ~
  \E_{X\sim \pworst} [\condrisk].
\end{equation}
We call this loss minimization under \emph{mixture covariate shifts}.

As the latent mixture weight and components are unknown, it is impossible
to compute the loss~\eqref{eq:latmix} from observed data. Thus, we postulate
a lower bound $\alpha_0 \in (0, \half)$ on the subpopulation proportion
$\alpha$ and consider the set of potential minority subpopulations
\begin{equation*}
  \mixdist \defeq
  \left\{\pworst: \pmarg = \alpha \pworst + (1-\alpha)\pnuisance ~~\mbox{for
      some}~
  \alpha \ge \alpha_0
  ~\mbox{and distribution}~\pnuisance~\mbox{on}~\mc{X}
  \right\}.
\end{equation*}
Concretely, our goal is to minimize worst-case subpopulation
risk $\risk$,
\begin{equation}
  \label{eq:grouploss}
  \minimize_{\param\in\paramdomain}
  \left\{ \risk(\theta)
  \defeq \sup_{\pworst \in \mixdist}\EX{X\sim \pworst}[ \condrisk]
  \right\}.
\end{equation}
The worst-case formulation~\eqref{eq:grouploss} is a distributionally robust
optimization (DRO) problem~\citep{Ben-TalGhNe09, ShapiroDeRu09} where we
consider the worst-case loss over mixture covariate shifts $Q_0 \in \mixdist$,
and we term the methodology we develop around this formulation \emph{marginal}
distributionally robust optimization, as we seek robustness only to shifts in
the marginals over the covariates $X$. For datasets with heterogeneous
subpopulations (e.g.\ natural language processing corpora), the worst-case subpopulation corresponds
to a group that is ``hard'' under the current model $\theta$. As we detail in
the related work section, the approach~\eqref{eq:grouploss} has connections
with covariate shift problems, distributional robustness,
fairness, and causal inference. In particular, the dual form
of~\eqref{eq:grouploss} corresponds to the conditional-value-at-risk (CVaR) of
the conditional risk $\condrisk$.

In some instances, the worst-case subpopulation~\eqref{eq:grouploss} may be
too conservative; the distribution of $X$ may shift only on some components,
or we may only care to achieve uniform performance across one variable.  As
an example, popular computer-vision datasets draw images mostly from western
Europe and the United States~\citep{ShankarHaBrAtWiSc17}, but one may wish
for models that perform uniformly well over different geographic
locations. In such cases, when one wishes to consider distributional shifts
only on a subset of variables $X_1$ (e.g.\ geographic location) of the
covariate vector $X=(X_1,X_2)$, we may simply redefine $X$ as $X_1$, and $Y$
as $(X_2,Y)$ in the problem~\eqref{eq:grouploss}. All of our subsequent
discussion generalizes to such scenarios.

On the other hand, because of confounding, the assumption that the conditional
distribution $\pcond$ does not change across groups may be too
optimistic. While the assumption is appropriate for machine learning tasks
where human annotators use $X$ to generate the label $Y$, many problems
include \emph{unmeasured} confounding variables that affect the label $Y$ and
vary across subpopulations.  For example, in a recidivism prediction task, the
feature $X$ may be the type of crime, the label $Y$ represents re-offending,
and the subgroup may be race; without measuring unobserved variables, such as
income or location, $\pcond$ is likely to differ between groups. To address
this issue, in Section~\ref{section:confounding} we generalize our proposed
worst-case loss~\eqref{eq:grouploss} to incorporate worst-case confounding
shifts, providing finite-sample upper bounds on worst-case loss whose
tightness depends on the effect of the unmeasured confounders on the
conditional risk $\condrisk$.

\subsection{Overview of results}

In the rest of the paper, we construct a tractable finite sample approximation
to the worst-case problem~\eqref{eq:grouploss}, and show that it allows
learning models $\theta \in \Theta$ that perform \emph{uniformly well} over
subpopulations. Our starting point is the duality result (see
Section~\ref{section:cvar})
\begin{align*}
  \risk(\theta)
  & \defeq
  \sup_{\pworst \in \mixdist} \mathop{\E}_{X\sim \pworst}[ \condrisk ] \nonumber
  = \inf_{\eta} \left\{ 
  \frac{1}{\alpha_0} \mathop{\E}_{X\sim \pmarg}[\hinge{\condrisk - \eta}]
  + \eta \right\}. \nonumber
\end{align*}
For convex losses, the dual form yields a single convex loss minimization in
the variables $(\theta, \eta)$ for minimizing $\risk(\theta)$. When we
(approximately) know the conditional risk $\condrisk$---for example,
when we have access to replicate observations $Y$ for each $X$---it
is reasonably straightforward to develop estimators for the
risk~\eqref{eq:grouploss} (see Section~\ref{section:cert}).

Estimating the conditional risk via replication is infeasible in scenarios in
which $X$ corresponds to a unique individual (similar to issues in estimation
of conditional treatment effects~\citep{ImbensRu15}). Alternative procedures
that depend on parametric assumptions on the family of conditional risks
$\E[\loss(\theta; X, Y) \mid X]$ for all $\theta \in \Theta$ are
restrictive, as we study learning problems over a flexible class of machine
learning models $\theta \in \Theta$ (e.g., random forests, gradient boosted
decision trees, kernel methods, neural networks). In this work, we instead
consider a scalable nonparametric approach involving the variational
representation
\begin{equation}
  \label{eq:pop-var}
  \E[\hinge{\condrisk - \eta}]
  = \sup_{h: \mc{X} \to [0, 1]}
  \E_P [h(X) (\lossvar - \eta)].
\end{equation}
As the space $\{h : \mc{X} \to [0, 1]\}$ is too large to effectively estimate the
quantity~\eqref{eq:pop-var}, we consider approximations via easier-to-control
function spaces $\mc{H} \subset \{h : \mc{X} \to \R\}$ and study the problem
\begin{equation}
  \label{eq:variational-opt}
  \minimize_{\theta \in \Theta, \eta} \left\{ \frac{1}{\alpha_0} \sup_{h \in \mc{H}}
  \E_P [h(X) (\lossvar - \eta)] + \eta \right\}.
\end{equation}
By choosing $\mc{H}$ appropriately---e.g.\ as a reproducing kernel Hilbert
space~\citep{BerlinetAg04, CristianiniSh04} or a collection of bounded
H\"{o}lder continuous functions---we can develop analytically and
computationally tractable approaches to minimizing
Eq.~\eqref{eq:variational-opt} to approximate Eq.~\eqref{eq:grouploss}.

Since the variational approximation to the dual objective
% $\theta \mapsto \inf_{\eta} \left\{ \frac{1}{\alpha_0} \sup_{h \in \mc{H}}
%   \E_P [h(X) (\lossvar - \eta)] + \eta \right\}$
is a lower bound on the worst-case subpopulation risk $\risk(\theta)$, it does
not (in general) provide uniform control over subpopulations
$Q_0 \in \mixdist$. Motivated by empirical observations that confirm the
limitations of this approach, we propose and study a more ``robust''
formulation than the problem~\eqref{eq:grouploss} that provides a natural
upper bound on the worst-case subpopulation risk~$\risk(\theta)$. Our proposed
formulation variational form analogous to Eq.~\eqref{eq:variational-opt} and
is estimable. If we consider a broader class of distributional shifts, we
arrive at a more conservative formulation than the
problem~\eqref{eq:grouploss}.  Define the R\'{e}nyi
divergence-ball~\citep{ErvenHa14} of order $q$
\begin{equation*}
  \mixdistq \defeq \{Q : \dren{Q}{P_X} \le \Delta\}~~~\mbox{where}~~~
  \dren{P}{Q} \defeq \frac{1}{q - 1} \log \int \left(\frac{dP}{dQ}\right)^q
  dQ.
\end{equation*}
Then for $1/p + 1/q = 1$ and $p \in (1,\infty)$, Lemma~\ref{lemma:cvar}
and~\citet[Section 3.2]{DuchiNa21} show
\begin{equation*}
  R_{\kdual}(\theta) \defeq \sup_{Q \in \mixdistq} \mathop{\E}_{X \sim Q}
  \left[\condrisk\right] = \inf_\eta
  \left\{\exp(\Delta / \kdual) \mathop{\E}_{X \sim P_X}
    \left[\hinge{\condrisk - \eta}^\kdual \right]^{\kdualinv}
    + \eta \right\}.
\end{equation*}

Abstracting the particular choice of uncertainty in $P_X$, for
$\kdual \in [1, \infty]$, the dual reformulation~\citep{Shapiro17, DuchiNa21}
\begin{equation}
  R_{\kdual}(\theta) = \inf_{\eta \ge 0} \left\{ 
    \frac{1}{\alpha_0}
    \left(
      \E_{X\sim \pmarg}\left[\hinge{\condrisk - \eta}^\kdual\right] \right)^{\kdualinv}
    + \eta \right\}
  \label{eq:two-norm-bound}
\end{equation}
always upper bounds the worst-case subpopulation
performance~\eqref{eq:grouploss}. As we show in Section~\ref{section:bounds},
for Lipschitz conditional risks
$x \mapsto \E[\loss(\param; (X,Y)) \mid X = x]$, Eq.~\eqref{eq:two-norm-bound}
is equal to a variant of the problem~\eqref{eq:variational-opt} where we take
$\mc{H}$ to be a particular collection of H\"{o}lder-continuous functions
allowing estimation from data.  Because our robustness approach in this paper
is new, there is limited analysis---either empirical or theoretical---of
similar problems.  Consequently, we perform some initial empirical evaluation
on simulations to suggest the appropriate approximation spaces $\mc{H}$ in the
dual form~\eqref{eq:variational-opt} (see Section~\ref{section:spaces}). Our
empirical analysis shows that the upper bound~\eqref{eq:two-norm-bound}
provides good performance compared to other variational procedures based
on~\eqref{eq:variational-opt}, which informs our theoretical development and
more detailed empirical evaluation to follow.

%We provide in
%Section~\ref{section:confounding} a generalization of our procedure under
%confounding, showing that we can still provide upper bounds under bounded
%confounding (see Assumption~\ref{assumption:bdd-confounding}).  In the rest of
%the paper, we refer to this method as the marginal distributionally robust
%optimization (DRO) method for reasons explained later.

% , applied to the
% estimator~\eqref{eq:procedure}---
We develop an empirical surrogate to the risk~\eqref{eq:two-norm-bound} in
Section~\ref{section:bounds}. A key advantage of our finite-sample procedure
is that it does not depend on unrealistic parametric assumptions on the
conditional risk $\condrisk$. Our main theoretical
result---Theorem~\ref{theorem:main-thm}---shows that the model
$\what{\theta}^{\rm rob}_n \in \R^d$ minimizing this empirical surrogate
achieves
\begin{equation*}
  \sup_{\pworst \in \mixdist} \E_{X\sim \pworst}[
    \E[\loss(\what{\theta}^{\rm rob}_n; (X,Y)) \mid X] ] \le \inf_{\theta \in \Theta} R_{\kdual}(\theta) +
  O\left(n^{-\frac{\kdual-1}{d+1}}\right),
\end{equation*}
with high probability
% , where
% \begin{equation*}
%   \label{eq:vopt}
%   \vopt \defeq
%   \inf_{\theta \in \Theta, \eta \ge 0} \left\{
%     \frac{1}{\alpha_0} \left( \E_{X\sim \pmarg}\left[
%   \hinge{\condrisk - \eta}^\kdual\right] \right)^{\kdualinv} + \eta\right\}
% \end{equation*}
whenever $x \mapsto \E[\loss(\theta; (X, Y)) \mid X = x]$ is suitably
smooth. In a rough sense, then, we expect that $\kdual$ trades between
approximation error---via the gap between
$\inf_{\theta \in \Theta} R_{\kdual}(\theta)$ and
$\inf_{\theta \in \Theta} R(\theta)$---and estimation error.

While our convergence guarantee gives the nonparametric rate
$O(n^{-\frac{\kdual-1}{d+1}})$, we empirically observe that our procedure
achieves low worst-case losses even when the dimension $d$ is large.  We
conjecture that this follows because our empirical approximation to
the $L^{\kdual}$ norm bound~\eqref{eq:two-norm-bound} is an \emph{upper bound}
with error only $O(n^{-\frac{1}{4}})$, but a \emph{lower bound} at the
conservative rate $O(n^{-\frac{\kdual-1}{d+1}})$. Such results---which we
present at the end of Section~\ref{section:bounds}---seem to point to the
conservative nature of our convergence guarantee in practical scenarios.  In
our careful empirical evaluation on semantic similarity assessment and
recidivism prediction tasks (Section~\ref{section:results}), we observe that
our procedure learns models that perform uniformly well across unseen minority
subpopulations and difficult examples. Nevertheless, the pessimistic
dependence on the dimension is unavoidable under nonparametric assumptions on
the conditional risk $\E[\loss(\theta; X, Y) \mid X]$, as we show in
Section~\ref{section:hardness}.  In light of these fundamental hardness
results, identifying a realistic yet restricted class of conditional risks
that allow faster statistical convergence is an interesting topic of future
work.

\subsection{Related work}

Several important issues within statistics and machine learning closely
relate to our goals of uniform performance across subpopulations.  We briefly
touch on a few of these connections here and hope that further linking them
may yield alternative approaches and deeper insights.

%% The goal of uniform performance across subpopulations is closely related to
%% existing work on covariate shifts and distributional
%% robustness, and we provide a partial review of the literature.

% In contrast to existing work on covariate shift, we consider unknown and worst
% case covariate shifts that are constrained via the subpopulation proportion
% $\alpha_0$. On the other hand, we are distinct from the robust optimization
% literature in considering structured robustness on the marginal distribution
% of $X$, and carefully analyzing the estimator in the context of covariate
% shifts. Our worst-case formulation~\eqref{eq:grouploss} is also closely
% related to existing methods for learning fair and causally robust models,
% which we detail below.

\paragraph{Covariate shifts.}

A number of authors study the case where a target distribution of
interest is different from the data-generating distribution---known as
covariate shift or sample selection bias~\citep{Shimodaira00,
  Ben-DavidBlCrPe07, SugiyamaKrMu07}. Much of the work focuses on the domain
adaptation setting where the majority of the observations come from a source
population (and corresponding domain) $\pdist$.  These methods require (often
unsupervised) samples from an a priori \emph{fixed} target domain, and apply
importance weight methods to reweight the observations when training a model
for the target~\citep{StorkeySu06, BickelBrSc07, GrettonSmHuScBoSc09,
  HuangGrBoKaScSm07}.
% via density ratios or density matching in a kernel Hilbert space
For multiple domains, representation based methods can identify sufficient
statistics not affected by covariate shifts
\citep{HeinzeMe17,GongZhLiTaGlSc16}.

On the other hand, our worst-case formulation assumes no knowledge of the
latent group distribution $\pworst$ (unknown target) and controls performance
on the worst subpopulation of size larger than $\alpha_0$.  Kernel-based
adversarial losses~\citep{WenYuGr14,LiuZi14,LiuZi17} minimize the worst-case
loss over importance weighted distributions, where the importance weights lie
within a reproducing kernel Hilbert space. These methods are similar in that
they consider a worst-case loss, but these worst-case weights provide no
guarantees (even asymptotically) for latent subpopulations.

\paragraph{Distributionally robust optimization.} A large body of work on
distributionally robust optimization (DRO) methods~\citep{Ben-TalHeWaMeRe13,
  BertsimasGuKa18, LamZh15, MiyatoMaKoNaIs15, DuchiGlNa21, NamkoongDu17,
  EsfahaniKu18, Shafieezadeh-AbadehEsKu15, BlanchetKaMu19, SinhaNaDu18,
  LeeRa17, GaoKl16, BlanchetKaZhMu17, GaoChKl17, LamQi19, StaibJe19,
  KuhnEsNgSh19} solves a worst-case problem over the \emph{joint distribution
  on $(X, Y)$}. On the other hand, our \emph{marginal DRO}
formulation~\eqref{eq:grouploss} studies shifts in the marginal covariate
distribution $X \sim \pmarg$. Concretely, we can formulate an analogue of our
marginal formulation~\eqref{eq:grouploss}
\begin{equation}
  \label{eq:joint-dro}
  \sup_{\pworst \in \jointmixdist} \E_{(X, Y) \sim \pworst}[\lossvar],
\end{equation}
where $\jointmixdist$ is the set of joint distributions $\pworst$ over
$(X, Y)$ such that $P = \alpha \pworst + (1-\alpha)\pnuisance$ for some
$\alpha \ge \alpha_0$ and probability $\pnuisance$ on $\mc{X} \times \mc{Y}$.
The \emph{joint DRO} objective~\eqref{eq:joint-dro} upper bounds the
\emph{marginal} worst-case formulation~\eqref{eq:grouploss}, and is frequently
too conservative (see Section~\ref{section:dro}).  By providing a tighter
bound on the worst-case loss~\eqref{eq:grouploss} under mixture covariate
shifts, our proposed finite-sample procedure~\eqref{eq:procedure} achieves
better performance on unseen subpopulations (see Sections~\ref{section:bounds}
and~\ref{section:results}). For example, the joint DRO
bound~\eqref{eq:joint-dro} applied to zero-one loss for classification may
result in a degenerate non-robust estimator that upweights \emph{all}
misclassified examples~\citep{HuNiSaSu18}, but our marginal DRO formulation
mitigates these issues by using the underlying metric structure.

Similar to our formulation~\eqref{eq:grouploss}, distributionally robust
methods defined with appropriate Wasserstein distances---those associated with
cost functions that are infinity when values of $Y$ differ---also consider
distributional shifts in the marginal covariate distribution $X \sim \pmarg$.
Such formulations allow incorporating the geometry of $X$, and consider local
perturbations in the covariate vector (with respect to some metric on
$\mc{X}$). Our worst-case subpopulation formulation~\eqref{eq:grouploss}
departs from these methods by considering all large enough \emph{mixture
  components (subpopulations)} of $\pmarg$, giving strong fairness and
tail-performance guarantees for learning problems.

\paragraph{Fairness.} A growing literature recognizes the challenges of
fairness within statistical learning~\citep{DworkHaPiReZe12, HardtPrSr16,
  KilbertasRoPaHaJaSc17, KearnsNeRoWu18, HashimotoSrNaLi18, ChenKaMaSvUd19},
which motivates our approach as well. Among the many approaches to this
problem, researchers have proposed that models with similar behavior across
demographic subgroups are fair~\citep{DworkHaPiReZe12, KearnsNeRoWu18}.  The
closest approach to our work is the use of Lipschitz constraints as a way to
constrain the labels predicted by a model~\citep{DworkHaPiReZe12}.  Rather than
directly constraining the prediction space, we use the Lipschitz continuity of
the conditional risk to derive upper bounds on model performance.  The gap
between joint DRO and marginal DRO relates to
``gerrymandering''~\citep{KearnsNeRoWu18}: fair models can be unreasonably
pessimistic by guaranteeing good performance against minority subpopulations
with high \emph{observed loss}---which can be a result of random
noise---rather than high \emph{expected loss}~\citep{DworkHaPiReZe12,
  KearnsNeRoWu18, HebertKiReRo17}. Our marginal DRO approach mitigates such
gerrymandering behavior relative to the joint DRO
formulation~\eqref{eq:joint-dro}; see Section~\ref{section:bounds} for a more
detailed discussion.

\paragraph{Causal inference.} A
common goal in causal inference is to learn models that perform well under
interventions, and one formulation of
causality is as a type of invariance across environmental
changes~\citep{PetersBuMe16}. In this context, our formulation seeking models
$\theta$ with low loss across marginal distributions on $X$ is an analogue of
observational studies in causal inference.  B\"{u}hlmann, Meinshausen, and
colleagues have proposed a number of procedures similar in spirit to our
marginal DRO formulation~\eqref{eq:grouploss}, though the key difference in
their approaches is that they assume that underlying environmental changes or
groups are \emph{known}. Their maximin effect methods find linear models that
perform well over heterogeneous data relative to a fixed baseline with known
or constrained population structure~\citep{MeinshausenBu15,
  RothenhauslerMeBu16, BuhlmannMe16}, while anchor
regression~\citep{RothenhauslerBuMePe18} fits regression models that perform
well under small perturbations to feature values. \citet{HeinzeMe17} consider
worst-case covariate shifts, but assume a decomposition between causal and
nuisance variables, with replicate observations sharing identical causal
variables. \citet{PetersBuMe16} use heterogeneous environments to discover
putative causal relationships in data, identifying robust models and
suggesting causal links.  Our work, in contrast, studies models that are
robust to \emph{mixture covariate shifts}, a new type of restricted
intervention over all large enough subpopulations.

% To cite
%domain adapt
% https://openreview.net/forum?id=HyPpD0g0Z
% http://proceedings.mlr.press/v48/gong16.html
% http://proceedings.mlr.press/v28/zhang13d.pdf
% https://arxiv.org/abs/1707.06422
% https://arxiv.org/abs/1507.05333
% https://papers.nips.cc/paper/3019-mixture-regression-for-covariate-shift.pdf

%unsupervised
% http://proceedings.mlr.press/v32/wen14-supp.pdf 
% http://papers.nips.cc/paper/5458-robust-classification-under-sample-selection-bias.pdf
% https://cs.nyu.edu/~roweis/papers/invar-chapter.pdf
% http://jmlr.csail.mit.edu/papers/volume10/bickel09a/bickel09a.pdf
% https://arxiv.org/pdf/1712.10050.pdf
% other names for our setting - simple covariate shift, prior probability shift, sample selection bias, imbalanced data, domain shift, and source component shift

% \subsection{Connections to causal inference}

% % We note however that loss minimization under covariate shift does not require
% % a causal interpretation and can be approached completely
% % probabilistically.

% \hn{I'm not sure exactly what points we're trying to make in this section. I
%   may also have butchered your original intent by rewriting some parts. Please
%   take a look (and revert to previous version if you want) $+$ make more
%   punchy.}

% While this limits the scope of the work, estimation under minimax confounding
% is far too pessimistic, and many machine learning tasks consist of $Y$
% generated via human labeling of $X$. In this scenario, the labels are by
% definition exogenous, and our model is appropriate.

%%% Local Variables:
%%% mode: latex
%%% TeX-master: "main"
%%% End:

\section{Performance Under Mixture Covariate Shift}
\label{section:dro}

We begin by reformulating the worst-case loss over mixture covariate
shifts~\eqref{eq:grouploss} via its dual (Section~\ref{section:cvar}).  We
first consider a simpler setting in which we can collect replicate labels $Y$
for individual feature vectors $X$---essentially, the analogue of a randomized
study in causal inference problems---showing that in this case appropriate
sample averages converge quickly to the worst-case loss~\eqref{eq:grouploss}
(Section~\ref{section:cert}).  Although this procedure provides a natural gold
standard when $x \mapsto \E[\loss(\theta; (x, Y)) \mid X = x]$ is estimable,
it is impossible to implement when large sets of replicate labels are
unavailable. This motivates the empirical fitting procedure we propose in
Eq.~\eqref{eq:procedure} to come, which builds out of the tractable upper
bounds we present in Section~\ref{section:bounds}.

% As written, this loss~\eqref{eq:grouploss} is a minimax optimization problem,
% and directly optimizing the objective with respect to $q$ does not provide a
% certificate for the performance of the model at test time. Popular optimizers
% such as stochastic gradient descent cannot guarantee approximate optimality of
% its solutions, and thus it is critical to derive objectives which are
% meaningful beyond the strict optimum. To do this, we will derive the dual of
% the worst case loss~\eqref{eq:grouploss}, which is a minimization problem.

\subsection{Upper bounds for mixture covariate shift}
\label{section:cvar}

% There is a correspondence between worst-case latent mixtures and
% expected loss thresholded at a level $\eta$.

Taking the dual of the inner maximization problem over covariate
shifts~\eqref{eq:grouploss} gives the below result.
\begin{lemma}
  \label{lemma:cvar}
  If $\E[| \condrisk|] < \infty$, then
  \begin{align}
    \label{eq:cvar}
    \sup_{\pworst \in \mixdist}\mathop{\E}_{X\sim \pworst}[ \condrisk ]
    = \inf_{\eta \in \R} \left\{
    \frac{1}{\alpha_0} \mathop{\E}_{X\sim \pmarg}\left[
      \hinge{\condrisk - \eta}\right]
      + \eta \right\}.
  \end{align}
  If additionally $0 \le \condrisk \le \lbound$ w.p.\ 1, the infimizing
  $\eta$ lies in $[0, \lbound]$.
\end{lemma}
\noindent See Section~\ref{section:proof-of-cvar} for the proof. The dual
form~\eqref{eq:cvar} is the conditional-value-at-risk (CVaR) of the
conditional risk $\condrisk$; CVaR is a common measure of risk in the
portfolio and robust optimization literatures~\citep{RockafellarUr00,
  ShapiroDeRu09}, but there it applies to an \emph{unconditional} loss, making
it (as we discuss below) conservative for the problems we consider.
%% Therefore
%% the worst-case loss~\eqref{eq:grouploss} over mixture covariate shifts
%% can be computed whenever $\condrisk$ is known.

The joint DRO~\eqref{eq:joint-dro} problem is more conservative than its
marginal counterpart~\eqref{eq:cvar} where the adversary selects over
distributions with a fixed $P_{Y \mid X}$; the joint DRO dual objective
$\inf_\eta \{\frac{1}{\alpha_0} \E[\hinge{\lossvar - \eta}] + \eta\}$ is
greater than the marginal DRO~\eqref{eq:cvar} unless $Y$ is a deterministic
function of $X$.  In Section~\ref{section:bounds}, we provide an approximation
to the marginal DRO dual form~\eqref{eq:cvar}, and one of our contributions is
to show that our procedure has better theoretical and empirical performance
than conservative estimators using the joint DRO
objective~\eqref{eq:joint-dro}.  Furthermore, we expect the joint DRO problem
to exhibit particular sensitivity to outliers in $Y \mid X$ unlike its
marginal counterpart. Both joint and marginal DRO are sensitive to outliers in
$X$---addressing this is an important topic of future research.

\begin{figure*}
  \subcaptionbox{\scriptsize Data for a 1-dimensional regression problem
    (circle) and the worst case distribution $\pworst$ for joint and marginal
    DRO (triangle/square). \label{fig:ped1}}[0.31\linewidth]{
    \includegraphics[scale=0.47]{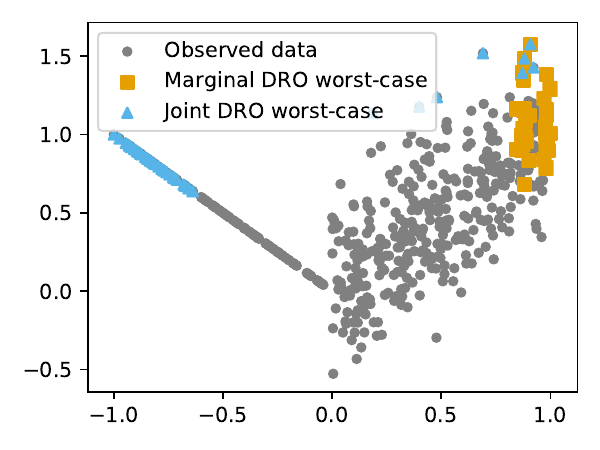}} \centering~
  \subcaptionbox{\scriptsize Best fit lines according to each loss. Only marginal DRO selects a line which fits both the $X>0$ and $X<0$ groups.    
    \label{fig:ped3}}[0.31\linewidth]{
    \includegraphics[scale=0.47]{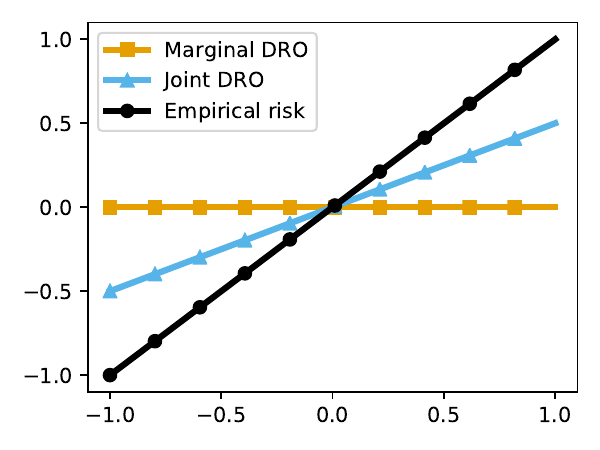}}~
  \subcaptionbox{\scriptsize Loss under different regression
    coefficients. Unlike Marginal DRO, ERM dramatically underestimates and
    joint DRO overestimates the worst case
    loss~\eqref{eq:grouploss}.\label{fig:ped2}}[0.31\linewidth]{
    \includegraphics[scale=0.47]{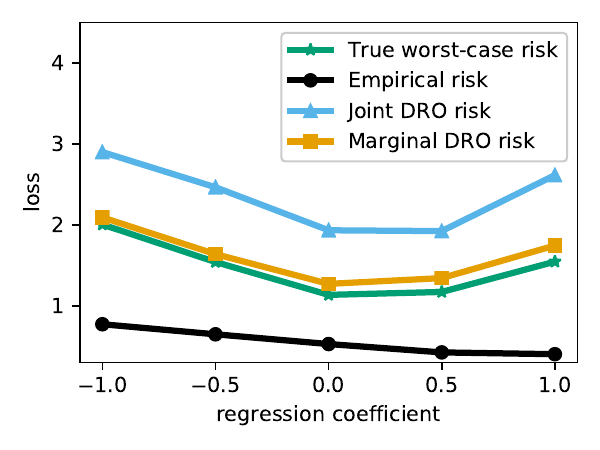}}
  \caption{Toy problem of $L_1$ regression through origin.}
\end{figure*}

To illustrate the advantages of the marginal distributionally robust approach,
consider a misspecified linear regression problem, where we predict
$\what{Y} = X \theta$ and use absolute the loss
$\loss(\theta; (x, y)) = |\theta x - y|$. Letting
$\varepsilon \sim \normal(0, 1)$, the following mixture model generates the
data
\begin{equation}
    Z \sim \text{Bernoulli}(0.15),
    ~~~X = (1-2Z) \cdot \text{Uniform}([0,1]),
    ~~~Y = |X| + \indic{X \ge 0} \cdot \varepsilon 
  \label{eqn:toy-problem}
\end{equation}
so that the subpopulation $Q_0(\cdot) \defeq P(\cdot | Z = 1)$ has minority
proportion $15\%$.  We plot observations from this model in
Figure~\ref{fig:ped1}, where 85\% of the points are on the right, and have
high noise.  The model with the best uniform performance is near $\theta =
0$, which incurs similar losses between left and right groups. In contrast,
the empirical risk minimizer ($\theta = 1$) incurs a high loss of $1$ on the
left group and $\sqrt{2/\pi}$ on the right one.

% TODO note - compute the exact value of losses on both sides, and find the exat minimizer for theta.

Empirical risk minimization tends to ignore the (left) minority group,
resulting in high loss on the minority group $X<0$
(Figure~\ref{fig:ped3}). The joint DRO solution~\eqref{eq:joint-dro} minimizes
losses over a worst-case distribution $\pworst$ consisting of examples that
receive high loss (blue triangles), which tend to be samples on $X>0$ due to
noise.  This results in a loose upper bound on the true worst-case risk as
seen in Figure~\ref{fig:ped2}.  Our proposed estimator selects a worst-case
distribution consisting of examples with high \emph{conditional risk}
$\condrisk$ (Figure~\ref{fig:ped1}, orange squares). This worst-case
distribution is not affected by the noise level, and results in a close
approximation to the true loss (Figure~\ref{fig:ped2}).

% If we ignore
% the Lipshitz constraint and attempt to find a plug-in estimator to our dual
% worst-case loss~\eqref{eq:cvar}, we obtain 

\subsection{Estimation via replicates}
\label{section:cert}

A natural approach to estimating the dual form~\eqref{eq:cvar} is a two
phase strategy, where we draw $X_1, \ldots, X_n \simiid P_X$ and then---for
each $X_i$ in the sample---draw a secondary sample of size $m$ i.i.d.\ from
the conditional $Y \mid X = X_i$. We then use these empirical samples to
estimate $\E[\loss(\theta; (X, Y)) \mid X = X_i]$. While it is not always
possible to collect replicate labels for a single $X$, \emph{human
  annotated} data---which is common in machine learning
applications~\citep{MarcusSaMa94, AsuncionNe07}---allows
replicate measurement, where we may ask multiple annotators to label the
same $X$.

We show this procedure can yield explicit finite sample bounds with error at
most $O(n^{-1/2} + m^{-1/2})$ for the population marginal robust
risk~\eqref{eq:grouploss} when the losses are bounded.
\begin{assumption}
  \label{assumption:bdd}
  For $\lbound < \infty$, we have $\loss(\param; (x, y)) \in [0,
  \lbound]$ for all $\theta \in \Theta, x \in \mc{X}, y \in \mc{Y}$.
\end{assumption}
\noindent Since we often want to show uniform concentration guarantees over
$\theta \in \Theta$, we make the following standard assumption to control
the size of the model class.
\begin{assumption}
  \label{assumption:lipschitz}
  $\theta \mapsto \loss(\theta; X, Y)$ is $K$-Lipschitz a.s., and
  $D \defeq \sup_{\theta, \theta' \in \Theta} \ltwo{\theta - \theta'} < \infty$.
\end{assumption}
\noindent The following estimate approximates the worst-case
loss~\eqref{eq:grouploss} for a fixed value of $\theta$.
\begin{proposition}
  \label{prop:repcert}
  Let Assumption~\ref{assumption:bdd} hold. There exists
  a universal constant $C$ such that for any
  fixed $\theta \in \Theta$, with
  probability at least $1-\delta$
  \begin{align*}
    \Bigg| \risk(\theta)
      - \inf_{\eta \in [0, \lbound]} \bigg\{ 
    \frac{1}{\alpha_0 n}  \sum_{i=1}^n
    \hingebigg{\frac{1}{m} \sum_{j=1}^m \loss(\theta; (X_i, Y_{i, j}))
      - \eta}
    + \eta \bigg\}
    \Bigg| \le C \frac{\lbound}{\alpha_0}
    \sqrt{\frac{1 + \log \frac{1}{\delta}}{\min\{m, n\}}}.
  \end{align*}
  If Assumption~\ref{assumption:lipschitz} also holds, then there exists
  another universal constant $C'$ such that
  $C'\frac{M + DK}{\alpha_0} \sqrt{\frac{1 + \log \frac{1}{\delta}}{\min\{m,
      n\}}}$ bounds the left hand side uniformly over $\theta \in \Theta$ with
  probability at least $1-\delta$.
\end{proposition}
\noindent See Section~\ref{section:proof-of-repcert} for the proof. The
estimator in Proposition~\ref{prop:repcert} approximates the worst-case
loss~\eqref{eq:grouploss} well for large enough $m$ and $n$.
However---similar to the challenges of making causal inferences
from observational data and estimating conditional treatment effects---it
is frequently challenging or impossible to collect replicates for individual
observations $X$, as each $X$ represents an unrepeatable unique measurement.
Consequently, the quantity in Proposition~\ref{prop:repcert} is a type
of gold standard, but achieving it can be practically challenging.

%% When $m = 1$, the above estimator
%% corresponds to the empirical approximation to the joint DRO
%% problem~\eqref{eq:joint-dro}, which can be prohibitively loose in
%% approximating the worst-case loss $\risk(\theta)$.

%%% Local Variables:
%%% mode: latex
%%% TeX-master: "main.tex"
%%% End:

\section{Variational Approximation to Worst-Case Loss}
\label{section:spaces}

The difficulty of collecting replicate data, coupled with the
conservativeness of the joint DRO objective~\eqref{eq:joint-dro} for
approximating the worst-case loss $\risk(\theta)$, impel us to study tighter
approximations that do not depend on replicates. Recalling
the variational representation~\eqref{eq:pop-var}, our goal is to minimize
\begin{equation*}
  \risk(\theta) =
  \inf_{\eta} \left\{ \frac{1}{\alpha_0} \sup_{h: \mc{X} \to [0, 1]}
  \E_P [h(X) (\lossvar - \eta)] + \eta \right\}.
\end{equation*}
As we note in the introduction, this quantity is challenging to work with, so
we restrict $h$ to subsets $\mc{H} \subset \{h : \mc{X} \to [0, 1]\}$.  The
advantage of this formulation and its related
relaxations~\eqref{eq:variational-opt} is that it replaces the dependence on
the conditional risk with an expectation over the joint distribution on
$(X, Y)$, which we may estimate using the empirical distribution, as we
describe in the next section.

Each choice of a collection of functions
$\mc{H} \subset \{h : \mc{X} \to [0, 1]\}$ to approximate the variational
form~\eqref{eq:pop-var} in the formulation~\eqref{eq:variational-opt} yields a
new optimization problem. The lack of a ``standard'' choice motivates us to
perform experiments to direct our development. In
Section~\ref{section:spaces-appendix}, we develop several candidate
approximations that are computationally feasible. \emph{A priori} it is
unclear whether different formulations should yield better performance; at
least at this point, our theoretical understanding provides similarly limited
guidance.  To this end, we perform a small simulation study in
Section~\ref{section:comparison} to direct our coming deeper theoretical and
empirical evaluation, discussing the benefits and drawbacks of various choices
of $\mc{H}$ through the example we introduce in Figure~\ref{fig:ped1}, Section
\ref{section:cvar}. For ease of exposition, we initially defer these
comparisons to Section~\ref{section:spaces-appendix} and focus on developing
the approximation method that exhibits the best empirical performance.

% \subsection{Example approximations and empirical variants}
% \label{section:procedures}

% Instead of the previous two choices of $\mc{H}$ that are strictly contained
% in the space of bounded functions, we now consider a variational procedure
% that controls an upper bound to the worst-case loss $\risk(\theta)$.

We consider the $L^p$ upper
bound~\eqref{eq:two-norm-bound} on $\risk(\theta)$ as---as
we shall see---it provides the best
empirical performance. Recall that a function $f: \mc{X} \to \R$ is
$(\alpha, c)$-H\"older continuous for $\alpha \in (0, 1]$ and $c> 0$ if
$|f(x) - f(x') | \le c \norm{x-x'}^\alpha$ for all $x, x' \in \mc{X}$.  We
consider the function class consisting of \emph{$L^p$ bounded H\"older
  functions}, which we motivate via an $L^\kdual$-norm
bound~\eqref{eq:two-norm-bound} on the dual objective~\eqref{eq:cvar}.
For any $\kdual \in (1, \infty)$ and $\kexp = \frac{\kdual}{\kdual-1}$ we have
\begin{align}
  & \E_{X\sim \pmarg}\left[\hinge{\condrisk - \eta}\right]
    \le \left( \E_{X\sim \pmarg}\left[\hinge{\condrisk - \eta}^\kdual \right]\right)^{\kdualinv}
    \nonumber \\
  & \qquad\qquad\qquad ~ = \sup_{h} \left\{ 
    \E\left[ h(X) (\lossvar - \eta) \right]
    \mid h: \mc{X} \to \R_+, ~ \E[h(X)^\kexp] \le 1
    \right\}.
    \label{eq:ltwo-var}
\end{align}
If $x \mapsto \E[\loss(\param; (X,Y)) \mid X = x]$ is H\"older continuous, then the function
\begin{equation}
  \label{eq:opt-hstar}
  h\opt(x) \defeq
  \frac{\hinge{\E[\loss(\param; (X,Y)) \mid X = x]-\eta}^{\kdual-1}}
  {\left( \E_{X\sim \pmarg}\left[\hinge{\condrisk - \eta}^\kdual\right]
    \right)^{\kexpinv}}
\end{equation}
attaining the supremum in the variational form~\eqref{eq:ltwo-var} is H\"older
continuous with constant dependent on the magnitude of the denominator.  As we
show shortly, carefully selecting the smoothness constant and $L^p$ norm
radius allows us to ensure $h\opt \in \mc{H}$ and to derive guarantees for the
resulting estimator. 

Minimizing the $L^p$ upper bound rather than the original variational
objective (alternatively, seeking higher-order robustness than the CVaR of
the conditional risk $\condrisk$ as in our discussion of the
quantity~\eqref{eq:two-norm-bound}) incurs approximation error. In
practice, our experience is that this gap has limited effect, and the
following lemma---whose proof we defer to
Section~\ref{section:proof-of-approx-error}---quantifies the approximation
error in inequality~\eqref{eq:ltwo-var}.
\begin{lemma}
  \label{lemma:approx-error}
  Let Assumption~\ref{assumption:bdd} hold and $Z(X)= \condrisk$. For $\eta \in [0, \zbound]$
  \begin{align*}
    & \left( \E_{X\sim \pmarg}\left[\hinge{Z(X) - \eta}^\kdual \right]\right)^{\kdualinv}
    \le \min \Bigg\{
    (\zbound - \eta)^{\kexpinv} \left( \E\hinge{Z(X) -\eta} \right)^{\kdualinv}, \\
      & \qquad \E\hinge{Z(X) -\eta} + \kdual^{\kdualinv} (\zbound - \eta)^{\kexpinv} 
      \left(\E \left|
      \hinge{Z(X) - \eta} - \E[\hinge{Z(X) - \eta}]
      \right|\right)^{\kdualinv}
    \Bigg\}.
  \end{align*}
\end{lemma}
%\noindent Note that the above two alternative bounds are complementary; as
%$\eta \uparrow \esssup \condrisk$, the first bound becomes loose, but the
%second bound becomes tighter.

We now formally show that the $L^p$ variational form provides a tractable
upper bound to the worst-case loss for Lipschitzian conditional risks.
%As the space of all measurable
%functions $h: \mc{X} \to \R_+$ in the variational form~\eqref{eq:ltwo-var}
%is too large, we consider a restriction to Lipschitzian risks.
\begin{assumption}
  \label{assumption:loss-lip}
  For $\theta \in \Theta$,
  the mappings $(x, y) \mapsto \loss(\param; (x,y))$ and
  $x \mapsto \E[\loss(\param; (x,Y)) \mid X=x]$ are $L$-Lipschitz.
\end{assumption}
\noindent
% Recall that a function $f: \mc{X} \to \R$ is $(\alpha, c)$-H\"older
% continuous for $\alpha \in (0, 1]$ and $c < \infty$ if
% $|f(x) - f(x') | \le c \norm{x-x'}^\alpha$ for all $x, x' \in \mc{X}$.
To ease notation let $\holderbound$ denote the space of H\"older continuous
functions
\begin{equation}
  \label{eq:lipspace}
  \holderbound \defeq \left\{ h: \mc{X} \to \R,~(\kdual-1, L^{\kdual-1})\mbox{-H\"older continuous} \right\}.
\end{equation}

If Assumption~\ref{assumption:loss-lip} holds and the
denominator in the expression~\eqref{eq:opt-hstar} has lower bound
$\epsilon > 0$, then $\epsilon h\opt \in \holderbound$, and we can approximate
the variational form~\eqref{eq:ltwo-var} by solving an analogous problem over
smooth functions. Otherwise, we can bound the
$L^\kdual$-norm~\eqref{eq:ltwo-var} by $\epsilon^{q-1}$, which is small for
small values of $\epsilon$. Hence, if we define a variational objective over smooth
functions $\holderbound$
\begin{equation}
  \label{eq:variational}
  \popobj(\theta, \eta) \defeq
  \sup_{h \in \holderbound} \left\{
    \E\left[ \frac{h(X)}{\epsilon} (\lossvar - \eta) \right]
    ~~\Bigg|~~ h \ge 0,~
    \left( \E[h(X)^{\kexp}] \right)^{\kexpinv} \le \epsilon
  \right\},
\end{equation}
we arrive at a tight approximation to the variational
form~\eqref{eq:ltwo-var}, which we prove in Section~\ref{section:proof-of-eps-lip-bound}.
\begin{lemma}
  \label{lemma:eps-lip-bound} 
  Let Assumptions~\ref{assumption:bdd},~\ref{assumption:loss-lip} hold and let
  $\kdual \in (1, 2]$.
  Then, for any $\theta \in \Theta$ and $\eta \in \R$,
  \begin{align*}
    & \left( \E_{X\sim \pmarg}\left[\hinge{\condrisk - \eta}^\kdual
      \right]\right)^{\kdualinv} 
      = \inf_{\epsilon \ge 0} \left\{
      \popobj(\theta, \eta) \vee \epsilon^{\kexp-1} \right\}
  \end{align*}
  and for any $\epsilon > 0$,
  $\left( \popobj(\theta, \eta) \vee \epsilon^{\kexp-1} \right)
  -\epsilon^{\kexp-1} \le \left( \E_{X\sim \pmarg}\left[\hinge{\condrisk -
        \eta}^\kdual \right]\right)^{\kdualinv}$.
\end{lemma}
Empirically, a variational approximation to the $L^{\kdual}$-norm
bound~\eqref{eq:two-norm-bound} based on the function
class~\eqref{eq:lipspace} outperforms other potential approximations
(Section~\ref{section:comparison}). We choose to  focus on it in the sequel.

%%% Local Variables:
%%% mode: latex
%%% TeX-master: "main"
%%% End:

\section{Tractable Risk Bounds for $L^p$ Variational Problem}
\label{section:bounds}

\newcommand{\sepobj}{Z(\theta, \eta; (X, Y))}

In this section, we develop an empirical approximation to the $L^p$ norm
bounded H\"older class, and formally develop and analyze a marginal DRO
estimator $\what{\theta}^{\rm rob}_n$. We derive this estimator by solving an
empirical approximation of the upper bound~\eqref{eq:variational} and provide
a number of generalization guarantees for this procedure. We complement these
results in Section~\ref{section:hardness} and quantify the fundamental
hardness of optimizing over subpopulations $\mixdist$ using finite samples.

%% Our main result of this section shows that our empirical approximation to
%% this variational form~\eqref{eq:ltwo-var} guarantees worst-case
%% loss~\eqref{eq:grouploss} no worse than
%% \begin{equation*}
%%   \vopt \defeq
%%   \inf_{\theta \in \Theta, \eta \in [0, \lbound]} \left\{
%%     \frac{1}{\alpha_0} \left( \E_{X\sim \pmarg}\left[
%%   \hinge{\condrisk - \eta}^\kdual\right] \right)^{\kdualinv} + \eta\right\},
%% \end{equation*}
%% up to an $O(n^{-\frac{p-1}{d+1}})$-approximation error. In what follows, we
%% derive a tractable upper bound on the population variational
%% form~\eqref{eq:ltwo-var} for smooth losses, and we show that its empirical
%% plug-in is a convex optimization problem. Then, we prove that our plug-in
%% estimator generalizes, allowing us to guarantee optimal performance as claimed.
%% Despite the conservative nonparametric convergence rate, we conclude this
%% section by showing that our empirical approximation is an upper bound on the
%% population variational form~\eqref{eq:ltwo-var} only up to an $O(n^{-1/4})$
%% error term.

\subsection{The empirical estimator}
\label{section:empirical-estimator}

Since the variational approximation $\popobj$ does not use the unknown
conditional risk $\condrisk$, its empirical plug-in is a natural finite-sample
estimator. Defining
\begin{equation}
  \label{eq:emplipspace}
  \empholderbound \defeq \left\{ h \in \R^n:
    ~h(X_i) - h(X_j) \le L^{\kdual-1} \norm{X_i - X_j}^{\kdual-1}
    ~~\mbox{for all}~~i,j \in [n] \right\},
\end{equation}
% \begin{align}
%   \label{eq:ub-primal-emp}
%   \inf_{\eta \in [0, \lbound]} \sup_{h \in \emplipspace} \left\{ 
%   \frac{1}{2 \alpha_0} 
%   \left( \E_{\emp}\left[ \frac{h(X)}{\epsilon} (\lossvar - \eta) \right]
%   \vee \epsilon \right)
%   + \eta~~\Bigg|~~
%   \E_{\emp}[h^2(X)] \le \epsilon^2
%   \right\}.
% \end{align}
we consider the estimator
\begin{align}
  \label{eq:ub-primal-emp}
  \empobjshort
  \defeq \sup_{h \in \empholderbound} \left\{
  \E_{\emp}\left[ \frac{h(X)}{\epsilon} (\lossvar - \eta) \right]
  ~\Bigg|~ h \ge 0,
  \left( \E_{\emp}[h^\kexp(X)] \right)^{\kexpinv} \le \epsilon
  \right\}.
\end{align}
The following lemma shows that the plug-in $\empobjshort$ is the infimum
of a convex objective.
\begin{lemma}
  \label{lemma:empirical-dual}
  For a sample $(X_1, Y_1), \ldots, (X_n, Y_n)$ and $B \in \R_+^{n \times n}$,
  define the
  empirical loss
  {\small
  \begin{equation}
    % \begin{split}
     \hspace{-10pt} \empobj \defeq 
       \bigg( \frac{\kdual-1}{n} \sum_{i=1}^n
      \hingeBig{ \loss(\param; (X_i,Y_i))
        - \frac{1}{n} \sum_{j=1}^n (B_{ij} - B_{ji}) -\eta}^\kdual
      \bigg)^{\kdualinv} 
        + \frac{L^{\kdual-1}}{\epsilon n^2}
      \sum_{i,j = 1}^n \norm{X_i - X_j}^{\kdual-1} B_{ij}.
    % \end{split}
    \label{eq:empobj}
  \end{equation}
  }%
  Then
  $\empobjshort = \inf_{B \ge 0} \empobj$ for all $\epsilon > 0$.
\end{lemma}
\noindent See Section~\ref{section:proof-of-empirical-dual} for proof. We can
interpret dual variables $B_{ij}$ as a transport plan for transferring the
loss from example $i$ to $j$ in exchange for a distance dependent cost
represented by the last term in the preceding display.  The objective
$\empobj$ thus consists of transport costs and any losses larger than $\eta$
after smoothing according to the transport plan $B$.

Noting that $\empobj$ is jointly convex in $(\eta, B)$---and jointly convex in
$(\theta, \eta, B)$ if the loss $\theta \mapsto \lossvar$ is convex---we
consider the empirical minimizer
\begin{equation}
  \label{eq:procedure}
  \empparam \in \argmin_{\theta \in \Theta}
  \inf_{\eta \in [0, \lbound], B \in \R_+^{n\times n}} \left\{ 
    \frac{1}{\alpha_0 }
    \left( \empobj \vee \epsilon^{\kexp-1}\right) + \eta
  \right\}
\end{equation}
as an approximation to the worst-case mixture covariate shift
problem~\eqref{eq:grouploss}. We note that $\empparam$ interpolates between
the marginal and joint DRO solution; as $L \to \infty$, $B \to 0$ in the
infimum over $\empparam$ and $\empobjshort \to (\frac{p-1}{n} \sum_{i=1}^n
\hinges{\loss(\theta; (X_i, Y_i)) - \eta}^p)^{1/p}$, an existing
empirical approximation to the joint DRO problem~\citep{DuchiNa21}.

% More precisely, for bounded and $L$-smooth losses (see
% Assumption~\ref{assumption:loss-lip} for a precise statement), and a fixed
% value of $\epsilon > 0$, $\kdual \in (1,2]$, and
% $\kexp \defeq \kdual/(\kdual-1)$
% \begin{equation}
%   \label{eq:procedure}
%   \empparam \defeq \argmin_{\theta \in \Theta}
%   \inf_{\eta \in [0, \lbound], B \in \R^{n \times n}_+}
%   \left\{
%     \frac{1}{\alpha_0} \left( \empobj
%       \vee \epsilon^{\kexp-1} \right)
%     + \eta
%   \right\},
% \end{equation}
% where $\empobj$ is defined by the expression \jd{Aside: this kind of writing is passive and redundant}
% \begin{align}
%   \label{eq:empobj}
%   \empobj \defeq \frac{(\kdual-1)^\kdualinv}{\sqrt{n}}\norm{\hinge{\ell_n(\param;(X,Y))
%   - \eta - (B-B^\top)\onevec /n}}_\kdual +
%   \frac{L^{\kdual-1}}{\epsilon n^2} \tr(DB).
% %    \left( \frac{1}{n} \sum_{i=1}^n
% %    \hinge{ \loss(\theta; (X_i, Y_i)) - \eta
% %    - \frac{1}{n} \sum_{j=1}^n (B_{ij} - B_{ji}) }^2 
% %    \right)^{\half} + \frac{L}{\epsilon n^2}
% %    \sum_{i,j = 1}^n \norm{X_i - X_j} B_{ij}.
% \end{align}
% $\ell_n(\param;(X,Y))\in\mathbb{R}^n$ is the vector of losses over samples,
% and $D\in \mathbb{R}^{n\times n}$ is the pairwise distance matrix with entries
% $D_{ij} = \norm{X_i - X_j}^{\kdual-1}$. 

\subsection{Generalization and uniform convergence}

We now turn to uniform convergence guarantees based on concentration of
Wasserstein distances, which show that the empirical minimizer $\empparam$
in expression~\eqref{eq:procedure} is an approximately optimal solution to
the population bound~\eqref{eq:two-norm-bound}. First, we prove that the
empirical plug-in~\eqref{eq:ub-primal-emp} converges to its population
counterpart at the rate $O(n^{-\frac{p-1}{d+1}})$. For $\alpha \in
\openleft{0}{1}$, define the Wasserstein distance $W_\alpha(Q_1, Q_2)$
between two probability distributions $Q_1, Q_2$ on a metric space $\mc{Z}$
by
\begin{equation*}
  W_{\alpha}(Q_1, Q_2) \defeq
  \sup \left\{
  \left| \E_{Q_1}[h] - \E_{Q_2}[h] \right|
  \mid 
  h: \mc{Z} \to \R,~(\alpha, 1)\mbox{-H\"older continuous}
  \right\}.
\end{equation*}
The following result---whose proof we defer to
Section~\ref{section:proof-of-plug-in-rates}---shows that the empirical
plug-in~\eqref{eq:ub-primal-emp} is at most $W_{p-1}(P, \emp)$-away from its
population version.
\begin{lemma}
  \label{lemma:plug-in-rates}
  Let Assumptions~\ref{assumption:bdd},~\ref{assumption:loss-lip} hold, and
  $\diam(\mc{X}) + \diam(\mc{Y}) \le R$.  For
  $p \in \openleft{1}{2}, q = p / (p-1)$,
  \begin{align*}
    \sup_{\theta \in \Theta, \eta \in [0, \lbound]}
    \left| \epsilon \vee \empobjshort - \epsilon \vee \popobj(\theta, \eta)
    \right| \le B_{\epsilon} W_{\kdual-1}(\emp, P)
  \end{align*}
  for
  \begin{equation}
    \label{eq:eps-const}
    % \mbox{\noindent where}~~~~~~
    B_{\epsilon} \defeq
    \epsilon^{-\kexp} 2^{\kexp-1} R\lbound L^2
    + \epsilon^{-1} 2^{\kexp-1} L\left( 2\lbound + (\kexp-1)LR \right)
    + \epsilon^{\kexp-2} (\kexp-1) 2^{\kexp-2} L
    + LR.
  \end{equation}
\end{lemma}

Our final bound follows from the fact that the Wasserstein distance between
empirical and population distributions converges at rate
$n^{-(p-1)/(d+1)}$. (See Section~\ref{section:proof-of-main-thm} for proof.)
In the next subsection, we show that the exponential dependence on the
dimension is unavoidable even under more restrictive assumptions on the
conditional risk $\E[\loss(\theta; X, Y \mid X]$.
\begin{theorem}
  \label{theorem:main-thm}
  Let Assumptions~\ref{assumption:bdd},~\ref{assumption:loss-lip} hold,
  $\kdual \in (1, 2]$, $\diam(\mc{X}) + \diam( \mc{Y}) \le R$, and
  $\frac{d+1}{2} > p-1$. For constants $c_1, c_2 > 0$ depending on
  $\lbound, d, \kdual$, with probability  at least
  $1-c_1 \exp\left( -c_2n (t ^{\frac{d+1}{\kdual-1}} \wedge t^2) \right)$
  \begin{align}
    % \begin{split}
      \sup_{\pworst(x) \in \mixdist}
      \E_{X\sim \pworst}[ & \E[\loss(\empparam; (X,Y)) \mid X] ]
       \le \inf_{\eta \in [0, \lbound]} \left\{ 
      \frac{1}{\alpha_0}
      \left( \E\left[\hinge{\E[\loss(\empparam; (X,Y)) \mid X] - \eta}^\kdual
        \right] \right)^{\kdualinv}
      + \eta \right\} \nonumber \\
      &  \le \inf_{\theta \in \Theta, \eta \in [0, \lbound]} \left\{ 
      \frac{1}{\alpha_0}
      \left( \E\left[\hinge{\E[\loss(\theta; (X,Y))\mid X] - \eta}^\kdual\right] \right)^{\kdualinv}
      + \eta \right\}
      + \frac{\epsilon^{\kexp-1}}{\alpha_0}
      + \frac{2B_{\epsilon}t}{\alpha_0}.
      \label{eq:plug-in-bound} 
    % \end{split}
  \end{align}
\end{theorem}

Our concentration bounds exhibit tradeoffs for the worst case
loss~\eqref{eq:grouploss} under mixture covariate shifts; in
Theorem~\ref{theorem:main-thm}, the power $\kdual$ trades between
approximation and estimation error. As $\kdual \downarrow 1$, the value
$\inf_{\theta \in \Theta} R_p(\theta)$ defined by the infimum of the
expression~\eqref{eq:two-norm-bound} over $\theta \in \Theta$ approaches the
optimal value
$\inf_{\theta \in \Theta} \sup_{\pworst \in \mixdist}\EX{X\sim \pworst}[
\condrisk]$ so that approximation error goes down, but estimation becomes more
difficult.

\paragraph{Upper bounds at faster rates} Theorem~\ref{theorem:main-thm} shows
the empirical estimator $\empparam$ is approximately optimal with respect to
the $L^{\kdual}$-bound~\eqref{eq:two-norm-bound}, but with a conservative
$O(n^{-\frac{p-1}{d+1}})$-rate of convergence. On the other hand, we can still
show that $\empobjshort$ provides an \emph{upper bound} to the worst-case loss
under mixture covariate shifts~\eqref{eq:grouploss} at the faster rate
$O(n^{-\frac{1}{4}})$. This provides a conservative estimate on the
performance under the worst-case subpopulation. See
Section~\ref{section:proof-of-fast-rate-ub} for the proof.
\begin{proposition}
  \label{prop:fast-rate-ub}
  Let Assumptions~\ref{assumption:bdd} and~\ref{assumption:loss-lip} hold. There
  exist numerical constants $c_1, c_2 < \infty$ such that the following
  holds.
  Let $\theta \in \Theta$, $\epsilon > 0$,
  and $\kdual \in \openleft{1}{2}$. Then
  with probability at least $1-2\gamma$, uniformly over $\eta \in [0, \zbound]$
  \begin{align*}
    \E[\hinge{\condrisk - \eta}^\kdual]^{\kdualinv}
    \le \max\left\{\epsilon^{\kexp-1}, (1+\tau_n )^{\kexp-1}
    \empupperbd
    + \frac{c_1\zbound^2}{\epsilon^{\kexp-1}}
    \sqrt{\frac{1}{n}\log \frac{1}{\gamma}}
    \right\}
  \end{align*}
  where
  $\tau_n \defeq c_2\zbound^2\epsilon^{-\kexp} \sqrt{\frac{1}{n}
    \log\frac{1}{\gamma}}$ and
  $L_n(\gamma) \defeq L(1+\tau_n(\gamma, \epsilon))^{-\kexpinv}$.  If
  Assumption~\ref{assumption:lipschitz} further holds, the same bound with
  $M^2 + M^{p-1} KD$ in place of $M^2$ holds uniformly over
  $\theta \in \Theta$.
\end{proposition}
%\noindent See Section~\ref{section:proof-of-fast-rate-ub} for the proof.

% letting $\theta = \empparam$ in Proposition~\ref{prop:fast-rate-ub} and
% recalling the dual form~\eqref{eq:cvar}
% \begin{align*}
%   & \sup_{\pworst \in \mixdist}\EX{X\sim \pworst}[\E[\loss(\empparam; (X,Y)) \mid X]] \\
%   & \le \inf_{\eta \in [0, \zbound]} \left\{
%     \left( (1+\tau_n(\gamma, \epsilon))^{\kexpinv}~
%       \what{g}_{\epsilon, L_n(\gamma), \kdual, n} (\theta, \eta) \vee \epsilon \right)
%     + \eta \right\}
%   + \frac{c_2\zbound^2}{\epsilon^{\kexp-1}} \sqrt{\frac{1}{n} \log \frac{1}{\gamma}}
% \end{align*}
% with probability at least $1-2\gamma$.

%%% Local Variables:
%%% mode: latex
%%% TeX-master: "main"
%%% End:

\section{Fundamental hardness of marginal DRO}
\label{section:hardness}

So far in our development, we only required flexible nonparametric assumptions
on the conditional risk $\E[\loss(\theta; X, Y) \mid X = x]$ for all
$\theta \in \Theta$. We view this as a practically important aspect of our
approach; a learning procedure should not depend on unrealistic modeling
assumptions. In this section, we show that the pessimistic scaling with the
problem dimension we saw in the previous section is unavoidable when
considering a nonparametric class of conditional risks. Optimization of both
the original worst-case subpopulation risk~\eqref{eq:grouploss} and the
$L^p$-norm the upper bound are governed by similar pessimistic dependence on
the dimension.

We study the fundamental hardness of optimizing the worst-case subpopulation
risk
$\risk(\theta; P) = \sup_{\pworst \in \mixdist(P)}\EX{X\sim \pworst}[
\E_P[\loss(\theta; X, Y) \mid X]]$, where we now make explicit the dependence
on the data-generating distribution $P$ in the notation.  We show that the
fundamental statistical difficulty of solving marginal DRO problems follow a
standard nonparametric rate when only requiring the conditional risk
$x \mapsto \E_P[\loss(\theta; X, Y) \mid X = x]$ to be a H\"{o}lder-smooth
function. Recall that the H\"{o}lder class $\holderball{\holdersmooth}$ of
$\holdersmooth$-smooth functions for
$\holdersmooth_1 = \ceil{\holdersmooth} - 1$ and
$\holdersmooth_2 = \holdersmooth - \holdersmooth_1$ is
\begin{equation}
  \label{eqn:holder-smooth}
  \holderball{\holdersmooth}
  \defeq \left\{ \mu(\cdot) \in C^{\holdersmooth_1}(\mc{X}):
    \sup_{\tiny
      \begin{array}{c}x\in \mc{X} \\
        \sum_{k=1}^d \gamma^k< \holdersmooth_1 \end{array}}|D^\gamma \mu(x)| \le 1,
    ~\sup_{\tiny \begin{array}{c} x \neq x' \in \mc{X} \\
                   \sum_{k=1}^d \gamma^k=\holdersmooth_1
         \end{array}} \frac{|D^\gamma \mu(x) - D^\gamma \mu(x')|}{\norm{x - x'}^{\holdersmooth_2}} \le 1
     \right\},
\end{equation}
where $C^{\holdersmooth_1}(\mc{X})$ denotes the space of
$\holdersmooth_1$-times continuously differentiable functions on $\mc{X}$, and
$D^{\gamma} = \frac{\partial^{\gamma}} {\partial^{\gamma^1} \ldots
  \partial^{\gamma^d}}$, for any $d$-tuple of nonnegative integers
$\gamma = (\gamma^1, \ldots, \gamma^d)$. Let $\mathfrak{P}_{\holdersmooth}$ be
the set of data-generating distributions with H\"{o}lder smooth conditional
risk uniformly over $\theta \in \Theta$
\begin{equation*}
  \mathfrak{P}_{\holdersmooth}
  \defeq  \left\{ P: \E_P[\loss(\theta; X, Y) \mid X = \cdot]
  \in \holderball{\holdersmooth}~~ \mbox{for~all}~ \theta \in \Theta,
  ~ |Y| \le 1~P\mbox{-a.s.}\right\}.
\end{equation*}

We study the finite sample minimax risk for a sample of size $n$
\begin{equation}
  \label{eqn:minimax}
  \mc{M}_n \defeq \inf_{\what{\theta}}
  \sup_{P \in \mathfrak{P}_{\holdersmooth}}
  \E_P\left[ \risk(\what{\theta}; P) - \inf_{\theta \in \Theta} \risk(\theta; P) \right]
\end{equation}
where the outer infimum is over all measurable functions of the data
$\{X_i, Y_i\}_{i=1}^n$. In the definition~\eqref{eqn:minimax}, the inner
supremum is not to be confused with the worst-case over subpopulations in
$\mixdist$ defining our distributionally robust formulation.
With this, in Appendix~\ref{section:proof-lower-bound} we
prove the following.
\begin{theorem}
  \label{theorem:lower-bound}
  Let $\mc{X} = [0, 1]^d$, $\Theta = [0, 1]$, $\loss(\theta; X, Y) = \theta
  \cdot Y$.  There are constants $N, c > 0$ depending on $(d, \alpha_0,
  \holdersmooth)$, such that for all $n \ge N$, $\mc{M}_n \ge c
  n^{-\frac{2\holdersmooth}{2\holdersmooth + d'}}$ where $d' = d$ for odd
  $d$ and $d-1$ for even $d$.
\end{theorem}
\noindent
Our minimax lower bound shows that the exponential sample complexity
in the dimension $d$ is unavoidable in the nonparametric minimax
sense~\eqref{eqn:minimax}, so that while the bounds
Theorem~\ref{theorem:main-thm} guarantees may not be completely sharp,
the worst-case exponential dependence on dimension $d$ is real.
As is typical in nonparametric estimation,
we recover parametric rates as $\holdersmooth \to \infty$. More
carefully identifying the (problem-dependent) constants
$c, N$ remains a goal of future work.

A similar argument shows it is equally difficult to optimize the $L_p$-upper
bound~\eqref{eq:two-norm-bound} on the worst-case subpopulation risk.
We again study the finite sample minimax risk for a sample of size $n$
\begin{equation*}
  \mc{M}_{n, p} \defeq \inf_{\what{\theta}}
  \sup_{P \in \mathfrak{P}_{\holdersmooth}}
  \E_P\left[ \risk_{\kdual}(\what{\theta}; P) - \inf_{\theta \in \Theta} \risk_{\kdual}(\theta; P) \right]
\end{equation*}
where the outer infimum is over all measurable functions of the data
$\{X_i, Y_i\}_{i=1}^n$.
In Appendix~\ref{sec:proof-lower-bound-higher-order}, we prove
the following result via a trivial adaptation of the proof
of Theorem~\ref{theorem:lower-bound}.
\begin{corollary}
  \label{cor:lower-bound-higher-order}
  Let the conditions of Theorem~\ref{theorem:lower-bound} hold.
  There are constants $N, c > 0$
  depending on $(d, \alpha_0, \holdersmooth)$
  such that for $n \ge N$,
  $\mc{M}_{n, p} \ge c
  n^{-\frac{2\holdersmooth}{2\holdersmooth + d'}}$ where $d' = d$ for odd $d$
  and $d-1$ for even $d$.
\end{corollary}

%%% Local Variables:
%%% mode: latex
%%% TeX-master: "main"
%%% End:

\section{Experiments}
\label{section:results}

We now present empirical investigations of the procedure~\eqref{eq:procedure},
focusing on two main aspects of our results.  First, our theoretical results
exhibit nonparametric rates of convergence, so it is important to understand
whether these upper bounds on convergence rates govern empirical performance
and the extent to which the procedure is effective.  Second, on examples with
high conditional risk $\condrisk$, we expect our procedure to improve
performance on minority groups and hard subpopulations when compared against
joint DRO and empirical risk minimization (ERM). The code for all experiments
can be found in \url{https://github.com/hsnamkoong/marginal-dro}.

To investigate both of these issues, we begin by studying simulated data
(Section~\ref{sec:simul}) so that we can evaluate true convergence precisely.
We see that in moderately high dimensions, our procedure outperforms both ERM
and joint DRO on worst-off subpopulations; we perform a parallel simulation
study in Section~\ref{sec:conf-sim} for the confounded case (minimizing
$\confempobjshort$ of Lemma~\ref{lemma:conf-empirical-dual}).  After our
simulation study, we continue to assess the efficacy of our procedure on real
data, using our method to predict semantic similarity
(Section~\ref{sec:semsim}), wine quality (Section~\ref{sec:wine}) and crime
recidivism (Section~\ref{sec:compas}). In all of these experiments, our
results are consistent with our expectation that our
procedure~\eqref{eq:procedure} typically improves performance over unseen
subpopulations.

Hyperparameter choice is important in our procedures. We must choose a
Lipschitz constant $L$, worst-case group size $\alpha_0$, risk level
$\epsilon$, and moment parameter $\kdual$.  In our experiments, we see that
cross-validation is attractive and effective.  We treat the value $L/\epsilon$
as a single hyperparameter to estimate via a hold-out set or, in effort to
demonstrate sensitivity to the parameter, plot results across a range of
$L/\epsilon$.  As the objective~\eqref{eq:procedure} is convex standard
methods apply; we use (sub)gradient descent to optimize the problem parameters
over $\theta, \eta, B$.  In each experiment, we compare our marginal DRO
method against two baselines: empirical risk minimization (ERM) and joint
DRO~\eqref{eq:joint-dro}. ERM minimizes the empirical risk
$\minimize_{\theta \in \Theta} \E_{\emp}[\loss(\theta; (X, Y))]$, and provides
very weak guarantees on subpopulation performance. The joint DRO
formulation~\eqref{eq:joint-dro} is the only existing method that provides an
upper bound to the worst-case risk. We evaluate empirical plug-ins of
the dual
formulation ($p \ge 1$)
%% \begin{equation*}
%%   \inf_{\eta \in \R} \left\{ 
%%     \frac{1}{\alpha_0} \E\hinge{\lossvar - \eta}
%%     + \eta \right\},
%% \end{equation*}
%% as well as its $L^p$-norm counterpart for any $p \ge 1$
\begin{equation}
  \label{eq:joint-dro-dual}
  \inf_{\eta \in \R} \left\{ 
    \frac{1}{\alpha_0} \E[\hinge{\lossvar - \eta}^p]^{1/p}
    + \eta \right\},
\end{equation}
which is the joint DRO counterpart of our marginal DRO
procedure~\eqref{eq:procedure} for the same value of $p$. Joint DRO
formulations over other uncertainty sets (e.g.\ Wasserstein
balls~\citep{KuhnEsNgSh19}) do not provide guarantees on subpopulation
performance as they protect against different distributional shifts, including
adversarial attacks~\citep{SinhaNaDu18}.

\subsection{Simulation study: the unconfounded case}
\label{sec:simul}

\begin{figure*}
  \centering
  \includegraphics[width=\textwidth]{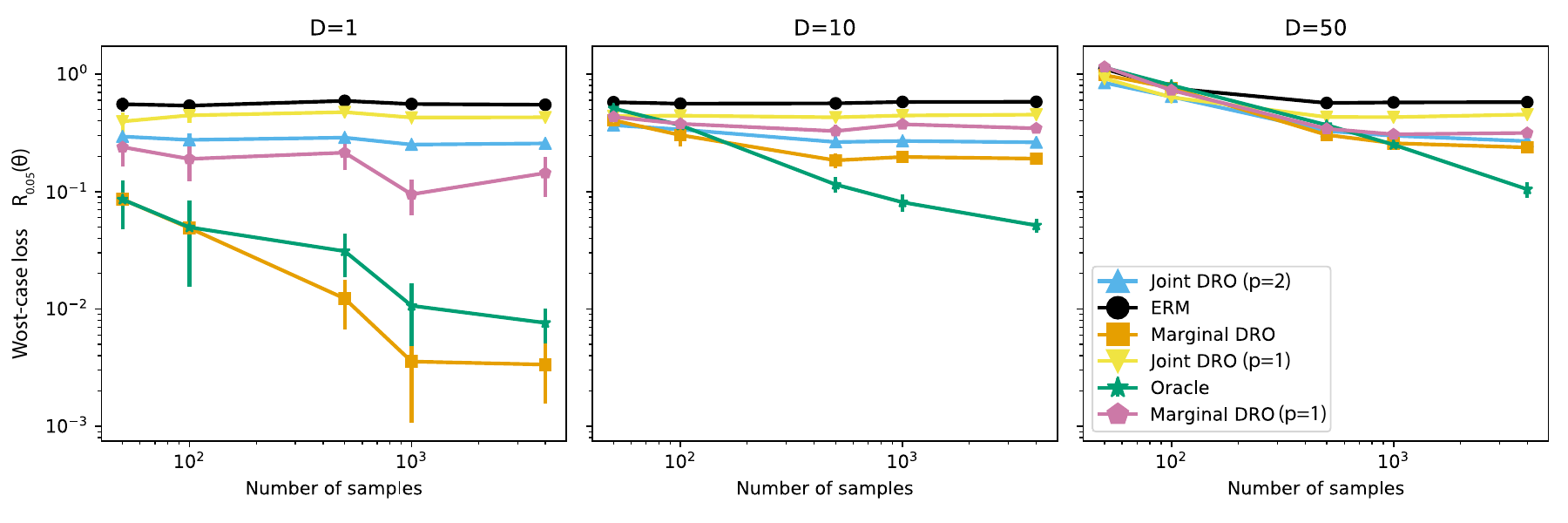}
  \caption{Dimension and sample size dependence of robust loss surrogates. The two marginal DRO methods correspond to different choices in the variational approximation ($L_p$ H\"older and Bounded H\"older).}
  \label{fig:dimdep}
    \ifdefined\useorstyle \vspace{-10pt} \fi
\end{figure*}

Our first simulation study focuses on the unconfounded
procedure~\eqref{eq:procedure}, where the data follows the
distribution~\eqref{eq:simdist}, and the known ground truth allows us to
carefully measure the effects of the problem parameters $(n,d,\alpha)$ and
sensitivity to the smoothness assumption $L$. We focus on the regression
example from Sec.~\ref{section:comparison} with loss $\loss(\theta; x, y) =
|\theta^\top x - y|$, so the procedure~\eqref{eq:procedure} is an empirical approximation
to minimizing the worst-case objective $\sup_{\pworst \in
  \mixdist}\EX{X\sim\pworst} \left[\E[|\theta^\top X - Y| \mid X ]\right]$.

The simulation distribution~\eqref{eq:simdist} captures several aspects of
loss minimization in the presence of heterogeneous subpopulations. The
subpopulations $X_1 \ge 0$ and $X_1 < 0$ constitute a majority and minority
group, and minimizing the risk of the majority group comes at the expense of
risk for the minority group. The two subpopulations also define an oracle
model that minimizes the maximum loss over the two groups. As the uniform
distribution exhibits the slowest convergence of empirical distributions for
Wasserstein distance~\citep{FournierGu15}---and as Wasserstein convergence
underpins our $n^{-1/d}$ rates in Lemma~\ref{lemma:plug-in-rates}---we use the
uniform distribution over covariates $X$.  We train all DRO models with
worst-case group size $\alpha_0=0.3$ and choose the estimated Lipschitz
parameter $L/\epsilon$ by cross-validation on a replicate-based estimate of
the worst-case loss~\eqref{eq:simval} (below) using a held-out set of 1000
examples and 100 repeated measurements of $Y$. We do not
regularize as $d \ll n$.

\paragraph{Effect of the $p$-norm bound}
We evaluate the difference in model quality as a function of $p$, which
controls the tightness of the $p$-norm upper bound.  Our convergence
guarantees in Theorem~\ref{theorem:main-thm} are looser for $p$ near $1$,
though such values achieve smaller asymptotic bias to the true
sub-population risk, while values of $p$ near 2 suggest a more favorable
sample size dependence in the theorem.

In Figure~\ref{fig:dimdep}, we plot the results of experiments for each
suggested procedure, where the horizontal axes index sample
size and the vertical axes an empirical approximation to the worst-case loss
\begin{equation}
  \label{eq:simval}
  R_{\alpha_0}(\theta) \defeq
  \sup_{\pworst \in \mc{P}_{\alpha_0,X}}
  \EX{X\sim\pworst}\E\left[|\theta^\top X - Y| \mid X\right]
\end{equation}
over the worst 5\% of the population (test-time $\alpha_0=0.05$);
the plots index dimensions $d = 1, 10, 50$.
. We evaluate
a worst-case error smaller than the true mixture proportion to measure
our procedure's robustness \emph{within} the minority subgroup.
The plots suggest
that the choice $p=2$ (Marginal DRO) outperforms $p=1$ (Linf
Marginal DRO), and performance generally seems to degrade as $p\downarrow 1$.
%Conveniently, when $p = 2$ the optimization
%problem~\eqref{eq:procedure} is a second-order cone problem (SOCP), which
%standard mathematical programming solvers support.
We consequently focus on the $p = 2$ case for the remainder of this section.

\paragraph{Sample size and dimension dependence}
We use the same experiment to also examine the pessimistic $O(n^{-1/d})$
convergence rate of our estimator~\eqref{eq:procedure}; this is substantially
worse than that for ERM and joint DRO~\eqref{eq:joint-dro}, both of which have
convergence rates scaling at worst as $1 / \sqrt{n}$~\citep{DuchiNa21}.
In low dimensions ($d=1$ to $d=10$)
convergence to the optimal function value---which we can compute
exactly---is relatively fast, and marginal DRO becomes substantially better
with as few as 500 samples (Figure~\ref{fig:dimdep}). In higher dimensions
($d=50$), marginal DRO convergence is slower, but it is only worse than the
joint DRO solution when $n=d=100$. At large sample sizes $n>1000$,
marginal DRO begins to strictly outperform the two baselines as measured
by the worst-case 5\% loss~\eqref{eq:simval}.

\begin{figure}
  \centering
  \subcaptionbox{Loss for various worst group sizes.\label{fig:pvsrho}}[0.49\textwidth]{  \includegraphics[width=0.4\textwidth]{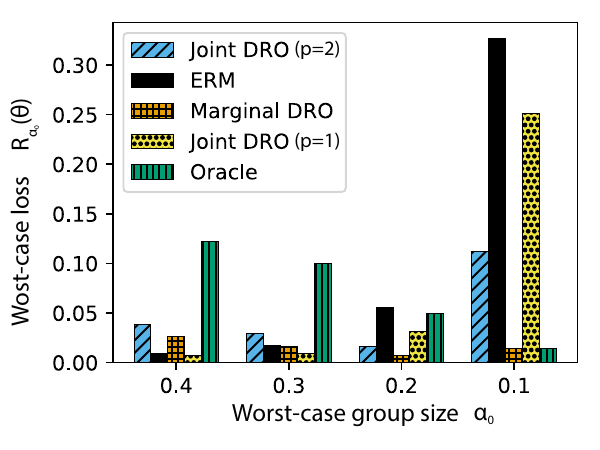}}
  \subcaptionbox{Marginal DRO losses across $L/\epsilon$}[0.49\textwidth]{ \includegraphics[width=0.4\textwidth]{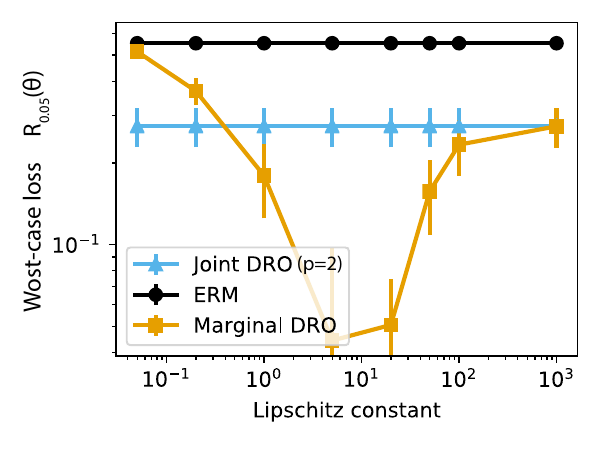}
    }
    \caption{Sensitivity of marginal DRO losses to test-time worst-case group
      size (left) and Lipschitz constant estimate (right).}
      \ifdefined\useorstyle \vspace{-10pt} \fi
    \label{fig:lip-sensitivity}
  \end{figure}

  Additional extended results in the supplement demonstrate that these results are robust to changes in the type of loss ($L_1$ vs $L_2$) Section~\ref{section:losschoice} and can be obtained with only a factor of 2 computational overhead Section~\ref{section:compoverhead}.

\paragraph{Sensitivity to robustness level}
We rarely know the precise minority proportion $\alpha_{\rm true}$, so that in
practice one usually provides a postulated lower bound $\alpha_0$; we
investigate sensitivity to its specification. We fix the data generating
distribution $\alpha_{\rm true}=0.15$ and train DRO models with
$\alpha_0=0.3$, while evaluating them using varying test-time worst case group
size $\alpha_0$ in Eq.~\ref{eq:simval}. We show the results of varying the
test-time worst-case group size in Figure~\ref{fig:pvsrho}. Marginal DRO
obtains a loss within $1.2$ times the oracle model regardless of the test-time
worst-case group size, while both ERM and joint DRO incur substantially higher
losses on the tails.

\paragraph{Sensitivity to Lipschitz constant}
Finally, the empirical bound~\eqref{eq:plug-in-bound} requires an estimate
of the Lipschitz constant of the conditional risk.
We vary the estimate $L/\epsilon$ in Figure~\ref{fig:lip-sensitivity}) for $\alpha=0.3$, $d=2$, and $n=1000$,
showing that the marginal robustness formulation has some
sensitivity to the parameter, though there is a range of
several orders of magnitude through which it outperforms
the joint DRO procedures. The behavior that it exhibits is expected,
however: the choice $L = 0$ reduces the marginal DRO procedure to
ERM in the bound~\eqref{eq:plug-in-bound}, while
the choice $L = \infty$ results in the joint DRO approach. In higher dimensions, marginal DRO will increasingly behave like joint DRO, leading to a smaller range of Lipschitz constants where marginal DRO performs well.

%% We show that small errors in the Lipschitz constant estimate do not result
%% in degenerate parameter estimates. Overestimating $L$ results in
%% interpolation of our estimator towards joint DRO, while underestimating
%% interpolates towards ERM (Figure~\ref{fig:lip-sensitivity}). This behavior
%% is consistent with the bound~\eqref{eq:plug-in-bound}, where $L=\infty$
%% reduces to joint DRO, and $L=0$ reduces to ERM.

\subsection{Semantic similarity prediction}
\label{sec:semsim}

We now present the first of our real-world evaluations of the marginal DRO
procedure~\eqref{eq:procedure}, focusing on a setting where we have multiple
measured outcome labels $Y$ for each covariate $X$, so it is possible to
accurately estimate the worst-case loss over covariate shifts.  We consider
the WS353 lexical semantic similarity prediction dataset
\citep{AgirreAlHaKrPaSo09} where the features are pairs of words, and labels
are a set of $13$ human annotations rating the word similarity on a $0$--$10$
scale. In this task, our goal is to use noisy human annotations of word
similarities to learn a robust model that accurately predicts word
similarities over a large set of word pairs.

 %(We could allow more general models, but we wish only to measure similarity, and we wish to keep the dimension of the resulting models reasonably small.)

We represent each word pair as the difference $(x_1 - x_2)$ of the word
vectors $x_1$ and $x_2$ associated to each word in
GloVe~\citep{PenningtonSoMa14} and cast this as a standard metric learning
task of predicting a scalar similarity $Y$ with a word-pair vector $X$ via the
quadratic model $x \mapsto x^\top \theta_1 x + \theta_2$,
$\theta_1 \in \R^{d \times d}$, $\theta_2 \in \R$. We use the absolute
deviation loss $|y - x^\top \theta_1 x + \theta_2|$. The training set consists
of $1989$ individual annotations of word similarities (ignoring any replicate
structure), and we fit the marginal DRO model with $p=2$, joint DRO
models~\eqref{eq:joint-dro-dual} with $p = 1, 2$, and an ERM model. All
methods use the same ridge regularizer tuned for the \emph{ERM model}. We
train all DRO procedures using $\alpha_0=0.3$ and tune the Lipschitz
constant via a held out set using the empirical estimate to the worst-case loss
based on replicate annotations (as in Proposition~\ref{prop:repcert}).

To evaluate each model $\theta = (\theta_1, \theta_2)$, we take an empirical
approximation to the worst-case loss over the word pairs with respect to the
\emph{averaged} human annotation
\begin{equation}
  \label{eq:semeval}
  R_{\alpha_0}(\theta) \defeq
  \sup_{\pworst \in \mc{P}_{\alpha_0,X}}
  \EX{X\sim\pworst}\left[|X^\top\theta_1 X + \theta_2 - \E\left[Y \mid X\right]|\right].
\end{equation}
This is a worst-case version of the standard word similarity
evaluation~\citep{PenningtonSoMa14}, where we also use averaged replicate
human annotations as the ground truth $\E[Y \mid X]$ in our evaluations, and
consider test-time worst-case group sizes $\alpha_0$ ranging from 0.01 to 1.0.

\begin{figure}[h]
  \centering
  \includegraphics[scale=0.7]{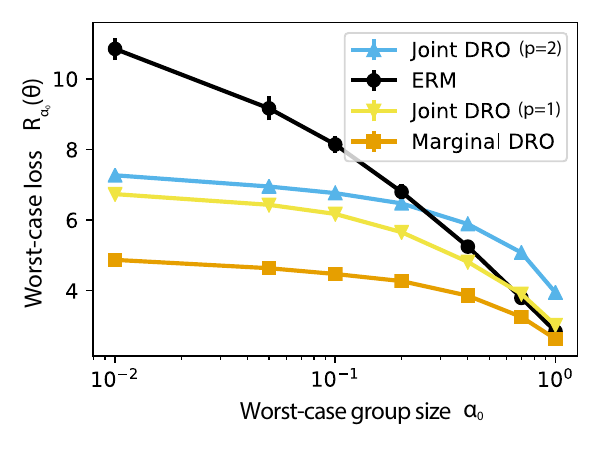}
  \caption{Semantic similarity prediction task, with worst-case prediction
    error $R_{\alpha_0}(\theta)$ (Eq.~\ref{eq:semeval}) over subgroups (y-axis)
    evaluated over varying test time worst-case group sizes $\alpha_0$
    (x-axis).}
  \label{fig:wvec}
  \ifdefined\useorstyle \vspace{-10pt} \fi
\end{figure}

All methods achieve low average error over the entire dataset, but ERM, joint
DRO and marginal DRO exhibit disparate behaviors for small subgroups. ERM
incurs large errors at $\sim 5\%$ of the test set, resulting in near
random prediction. Applying the joint DRO estimator reduces error by nearly
half and marginal DRO reduces this even further (Figure~\ref{fig:wvec}).

%\begin{figure}
%  \centering
%    \includegraphics[scale=0.8]{figs/compass_heat.pdf}
%\caption{Recidivism prediction task as a covariate shift problem. Models trained on the full dataset are evaluated on subgroups.}
%\label{fig:compas}
%\end{figure}

\subsection{Distribution shifts in wine quality prediction}
\label{sec:wine}
Next, we show that marginal DRO ($p=2$) can yield improvements outside
of the worst-case subgroup assumptions we have studied thus far. The
UCI wine dataset~\citep{CortezCeAlMaRe09} is a regression task with 4898
examples and 12 features, where each example is a wine with measured
chemical properties and the label $y \in \R$ is a subjective
quality assessment; the data naturally splits into subgroups of white and
red wines. We consider a distribution shift problem where the regression
model is trained on red wines but tested on (subsets of) white wines. Unlike
the earlier examples, the test set here does not correspond to
subpopulations of the training distribution, and the chemical
features of red wine are likely distinct from those for white
wines, violating naive covariate shift assumptions.

We minimize the absolute deviation loss $\loss(\theta; x, y) = |\theta^\top
x - y|$ for linear predictions, tuning baseline parameters (e.g.\ ridge
regularization) on a held-out set that is i.i.d.\ with the training
distribution.  Our training distribution is 1500 samples of red wines and
the test distribution is all white wines.
We evaluate models via their
loss over worst-case subgroups of the white wines, though in
distinction from earlier experiments, we
have no replicate labels. Thus we measure the joint DRO
loss~\eqref{eq:joint-dro}, i.e.\
\begin{equation*}
  R_{\alpha_0, \mbox{\scriptsize joint}}(\theta) \defeq \sup_{\pworst \in
    \mc{P}_{\alpha_0,(X_{w}, Y_{w})}} \EX{X_w,Y_w
  \sim\pworst}\left[|\theta^\top X_w - Y_w|\right],
\end{equation*}
so that the worst-case loss $R_{\alpha_0, \mbox{\scriptsize joint}}(\theta)$
measures the subgroup losses under the
white wine distribution. % $\mc{P}_{\alpha_0,(X_{w}, Y_{w})}$.

Figure \ref{fig:wine} shows the worst-case loss $R_{\alpha_0,
  \mbox{\scriptsize joint}}(\theta)$ over the test set as a function of
$\alpha_{0}$. Here, Marginal DRO with Lipschitz constants $L/\epsilon$
varying over $0.1$ to $300$ and $\alpha_0=0.3$ provides improvements over
both joint DRO and ERM baselines across the entire range of test-time
$\alpha_0$. Marginal DRO improves losses both for the pure distribution
shift from red to white wines (test-time group
size $\alpha=1.0$) as well as for the more pessimistic groups with small
test-time
proportion $\alpha_0$.
Distribution shift from red to white wines appears difficult to capture for the
pessimistic joint DRO methods.

\begin{figure*}
  \centering
  % \subcaptionbox{$L=0.1$}[0.49\textwidth]{ 
  \includegraphics[width=0.4\textwidth]{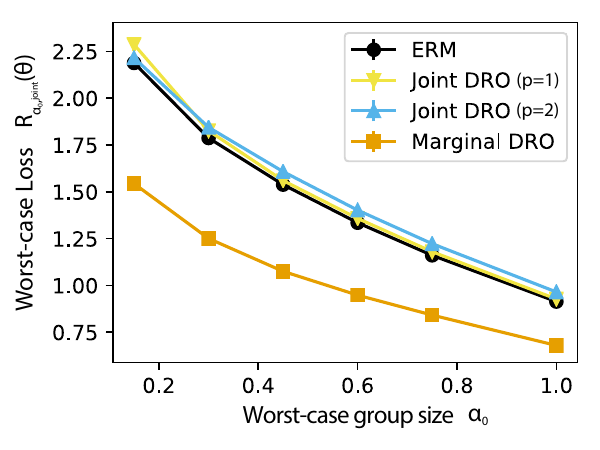}
%}
  % \subcaptionbox{$L=300$}[0.49\textwidth]{
  \includegraphics[width=0.4\textwidth]{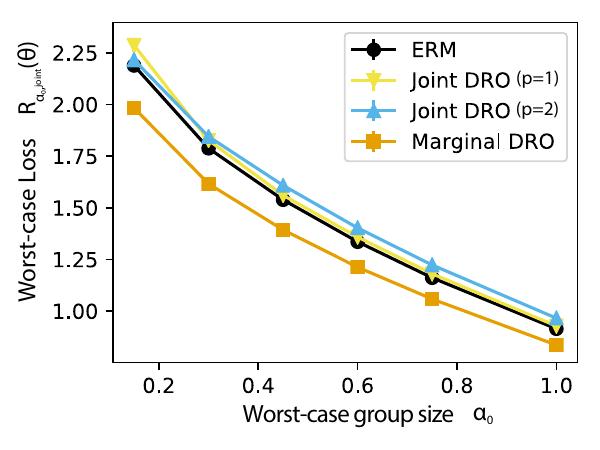}
%}
  \caption{Marginal DRO improves worst-case loss
    $R_{\alpha_0, \mbox{\scriptsize joint}}(\theta)$ for the wine quality
    prediction task under a real world red to white wine distribution
    shift. The gain holds on a wide range of Lipschitz constants from
    $L/\epsilon=0.1$ (left) to $300$ (right).}
  \label{fig:wine}
  \ifdefined\useorstyle \vspace{-10pt} \fi
\end{figure*}

\subsection{Recidivism prediction}
\label{sec:compas}

Finally, we show that marginal DRO $(p=2)$ can control the loss over a minority
group on a recidivism prediction task.  The COMPAS recidivism dataset
\citep{Chouldechova17a} is a classification task where examples are
individual convicts, features consist of binary demographic labels (such as
African American or not) and description of their crimes, and the label is
whether they commit another crime after release (recidivism). We use the fairML toolkit
version of this dataset \citep{Adebayo16}. Classification algorithms for
recidivism have systematically discriminated against minority groups, and
this dataset illustrates such
discrimination~\citep{BarocasSe16}.  We consider this dataset from the
perspective of achieving uniform performance across various groups. There
are $10$ binary variables in data, each indicating a potential split of the
data into minority and majority group (e.g.\ young vs.\ not young, or
Black vs.\ non-Black), of which $7$ have enough ($n >
10$) observations in each split to make reasonable error estimates.  We
train a model over the full population (using all the features), and for
each of the $7$ demographic indicator variables and evaluate
the held-out 0-1 loss over both the associated majority group and
minority group. % We then report the maximum of these two losses as the worst case loss for the demographic split.

Our goal is to ensure that the classification accuracy remains high
\emph{without} explicitly splitting the data on particular demographic labels
(though we include them in our models as they have predictive power). We use
the binary logistic loss with linear models and a $70/30$
train/test split. We set $\alpha_0=0.4$ for all DRO methods (approximately matching demographic statistics in the United States, with 60\% white and 40\% other races), and apply
ridge regularization to all models with regularization parameter
tuned for the \emph{ERM model}.
% $100$ to $0.001$.

\begin{table}
\resizebox{\textwidth}{!}{
\begin{tabular}{c | c c c c c c c}
Method & Old & Young & Black & Hispanic & Other race & Female & Misdemeanor\\
\toprule
ERM & 37.7 $\pm$ 0.8 & 44.6 $\pm$ 1.0 & 37.7 $\pm$ 0.8 & 37.5 $\pm$ 0.9 & 37.9 $\pm$ 1.1 & 37.5 $\pm$ 0.9 & 37.6 $\pm$ 0.8\\
\midrule
Joint ($p=2$) - ERM & 10.0 $\pm$ 2.0 & 4.0 $\pm$ 1.7 & 9.7 $\pm$ 1.7 & 9.6 $\pm$ 1.8 & 10.9 $\pm$ 2.3 & 9.2 $\pm$ 1.8 & 9.0 $\pm$ 1.6\\
Joint ($p=1$) - ERM & 5.8 $\pm$ 1.7 & 1.2 $\pm$ 1.7 & 5.2 $\pm$ 1.5 & 7.4 $\pm$ 1.9 & 6.6 $\pm$ 2.1 & 5.9 $\pm$ 1.7 & 5.1 $\pm$ 1.5\\
Marginal L=0.01 - ERM & -0.7 $\pm$ 0.7 & 1.4 $\pm$ 1.1 & -1.0 $\pm$ 0.7 & -1.7 $\pm$ 0.9 & -2.4 $\pm$ 1.1 & -1.5 $\pm$ 0.8 & -1.5 $\pm$ 0.7\\
Marginal L=0.001 - ERM & -1.1 $\pm$ 0.7 & 0.4 $\pm$ 1.1 & -1.4 $\pm$ 0.7 & -2.1 $\pm$ 0.9 & -2.6 $\pm$ 1.1 & -1.9 $\pm$ 0.8 & -1.8 $\pm$ 0.7\\
\bottomrule

\end{tabular}
}
\caption{Worst-case error of recidivism prediction models across demographic
  subgroups. The ERM row shows baseline worst-case error; subsequent
  rows show error differences from baseline (negatives indicate
  lower error). }
\label{tab:compas}
\end{table}

Table~\ref{tab:compas} presents the 0-1 loss on the seven demographic splits
over 100 random train/test splits. For each attribute (table column) we
split the test set into examples for which the attribute is true and false
and report the average 0-1 loss on the worst of the two groups.
The first row gives
the average
worst-case error and associated 95\% standard error for the ERM
model. For the DRO based models (remaining rows), we report the
average differences with respect to the baseline ERM model and
standard error intervals.  Unlike our earlier regression tasks, joint
DRO (both $L_2$ and $L_1$) performs worse than ERM on almost all demographic
splits. On the other hand, we find that marginal DRO with the appropriate
smoothness constant $L \in \{10^{-2}, 10^{-3}\}$ reduces classification
errors between $1$--$2\%$ on the worst-case group across various demographics,
with the largest error reduction of 3\% occurring in the young vs.\ old
demographic split.

% \thc{do we need a conclusion / discussion?}

%%% Local Variables:
%%% mode: latex
%%% TeX-master: "main"
%%% End:

\ifdefined\useorstyle
\setlength{\bibsep}{0em}
\else
\setlength{\bibsep}{.65em}
\fi
\bibliographystyle{abbrvnat}
\bibliography{bib,misc}

\ifdefined\useorstyle

\subsection*{Author Bios}

\begin{itemize}
\item John Duchi is an associate professor of Statistics and Electrical
  Engineering and (by courtesy) Computer Science at Stanford University. His
  work spans statistical learning, optimization, information theory, and
  computation.
\item Tatsunori Hashimoto is an assistant professor in the Computer Science
  department at Stanford university. His research uses tools from statistics
  to make machine learning systems more robust and reliable---especially in
  challenging tasks involving natural language.
\item Hongseok Namkoong is an assistant professor in the Decision, Risk, and
  Operations division at Columbia Business School and a member of the Columbia
  Data Science Institute. His research interests lie at the interface of
  machine learning, operations research, and causal inference, with a
  particular emphasis on developing reliable learning methods for
  decision-making problems.
\end{itemize}

%% Here starts the e-companion (EC)
%%%%%%%%%%%%%%%%%%%%%%%%%%%%%%%%%%%%%%%%%%%%%%%%%%%%%%%%%%
\ECSwitch

%\ECDisclaimer
%%%%%%%%%%%%%%%%%%%%%%%%%%%%%%%%%%%%%%%%%%%%%%%%%%%%%%%%%%

%%% Main head for the e-companion
\ECHead{Appendix}\

\section{Alternative variational approximations}
\label{section:spaces-appendix}

Recalling the variational representation~\eqref{eq:pop-var}, we wish to minimize
the variational approximation
\begin{equation*}
  \inf_{\eta} \left\{ \frac{1}{\alpha_0} \sup_{h \in \mc{H}}
    \E_P [h(X) (\lossvar - \eta)] + \eta \right\}.
\end{equation*}
For each choice of $\mc{H}$ we propose below, we consider an empirical
approximation $\what{\mc{H}}$, the subset of $\mc{H}$ restricted to mapping
$\{X_1, \ldots, X_n\} \to \R$ instead of $\mc{X} \to \R$, solving the
empirical alternative
\begin{align}
  \label{eq:emp-variational}
  \minimize_{\theta \in \Theta, \eta} \left\{ \frac{1}{\alpha_0}
  \sup_{h \in \what{\mc{H}}}
  \E_{\emp} [h(X) (\lossvar - \eta)] + \eta \right\}.
\end{align}
We design our proposals so the dual of the inner
supremum~\eqref{eq:emp-variational} is computable.  When the conditional risk
$x \mapsto \E[\loss(\param; (X,Y)) \mid X = x]$ is smooth, we can provide
generalization bounds for our procedures. We omit detailed development for our
first two procedures---which we believe are natural proposals, justifying a
bit of discussion---as neither is as effective as the last procedure in our
empirical evaluations, which controls the $L^p$ upper
bound~\eqref{eq:two-norm-bound} on $\risk(\theta)$.

\subsection{Example approximations and empirical variants}
\label{section:procedures}

\paragraph{Reproducing Hilbert kernel spaces (RKHS)} Let
$K : \mc{X} \times \mc{X} \to \R_+$ be a reproducing
kernel~\citep{BerlinetAg04, Aronszajn50} generating the reproducing kernel
Hilbert space $\mc{H}_K$ with associated norm $\norm{\cdot}_K$. For any
$R \in \R_+$, we can define a norm ball
\begin{equation*}
  \mc{H}_{K, R} \defeq \left\{h \in \mc{H}_K: \norm{h}_K \le R, h \in [0,1] \right\}
\end{equation*}
and consider the variational approximation~\eqref{eq:variational-opt} with
$\mc{H} = \mc{H}_{K, R}$.  To approximate the population variational problem
$\sup_{h \in \mc{H}_{K, R}} \E[h(X) (\lossvar - \eta)]$, we consider a
restriction of the same kernel $K$ to the sample space
$\{ X_1, \ldots, X_n\}$. Let
$K_n = \{ K(X_i, X_j)\}_{1\le i, j \le n}$ be the Gram matrix
evaluated on samples $X_1, \ldots, X_n$, and define the empirical
approximation
\begin{equation*}
  \what{\mc{H}}_{K, R} \defeq
  \left\{ h \in [0, 1]^n: h = K_n \xi~\mbox{ for some}~\xi \in \R^n~\mbox{such that}~\frac{1}{n^2} \xi^\top K_n \xi \le R
  \right\}.
\end{equation*}
(Recall that if $h(x) = \sum_{i = 1}^n K(x, X_i) \xi_i$, then
$\norm{h}_K^2 = \frac{1}{n^2} \xi^\top K_n \xi$.)  To compute the
empirical problem~\eqref{eq:emp-variational} with $\what{\mc{H}} =
\what{\mc{H}}_{k, R}$, we take the dual of the inner supremum. Simplifying
the dual form---whose derivation is a standard exercise in convex
optimization---we get
\begin{equation}\label{eq:rkhs-opt}
  \minimize_{\theta \in \Theta, \eta \in \R, \beta \in \R^{n}}
  \left\{ \frac{1}{\alpha_0 n} \sum_{i=1}^n
    \hinge{ \loss(\param; (X_i,Y_i))
      -  \eta + \beta_i}
    + \frac{1}{n} \sqrt{R^{-1} \beta^\top K_n \beta}
  \right\}.  
\end{equation}
For convex losses $\lossvar$, this is a convex optimization
problem in $(\theta, \beta, \eta)$.

\paragraph{H\"older continuous functions (bounded H\"older)}

Instead
of the space of bounded functions, we restrict attention to H\"older
continuous functions
\begin{equation}
  % \mc{H}_{c, \alpha} \defeq \left\{
  % h: \mc{X} \to [0, 1] \mid h~\mbox{is}~(\alpha, c)\mbox{-H\"older continuous}
  % \right\},
  \holderbound \defeq \left\{ h: \mc{X} \to [0,1] \mid
    h~\mbox{is}~~(\kdual-1, L^{\kdual-1})\mbox{-H\"older continuous} \right\},
  \label{eqn:holder-continuous-fns}
\end{equation}
where the particular scaling with respect to $\kdual \in (1, 2]$ and $L > 0$
is for notational convenience in Section~\ref{section:bounds}. The empirical plug-in of
$\holderbound$ is
\begin{equation*}
  \what{\mc{H}}_{L, p} \defeq \left\{ h: \{ X_1, \ldots, X_n \} \to [0, 1]
    \mid h~\mbox{is}~(\kdual-1, L^{\kdual-1})\mbox{-H\"older continuous} \right\},
\end{equation*}
the empirical plug-in of the variational problem~\eqref{eq:variational-opt}
with $\mc{H} = \holderbound$ is given by the
procedure~\eqref{eq:emp-variational} with
$\what{\mc{H}} =\what{\mc{H}}_{L, p}$. Taking the dual of the inner supremum
problem, we have the following equivalent dual formulation of the empirical
variational problem
\begin{equation}\label{eq:lipriskbd}
  \minimize_{\theta \in \Theta, \eta, B \in \R^{n \times n}_+}
  \bigg\{ \frac{1}{\alpha_0 n} \sum_{i=1}^n
    \hingeBig{ \loss(\param; (X_i,Y_i))
      - \frac{1}{n} \sum_{j=1}^n (B_{ij} - B_{ji}) -\eta}
    + \frac{L^{\kdual-1}}{n^2}
    \sum_{i,j = 1}^n \norm{X_i - X_j}^{\kdual-1} B_{ij}
  \bigg\}.
\end{equation}
For convex losses $\theta \mapsto \lossvar$, this is again a convex
optimization problem in $(\theta, B, \eta)$, and is always smaller than the
empirical joint DRO formulation~\eqref{eq:joint-dro}.

By definition, the population H\"older continuous variational approximation
provides the lower bound on the worst-case loss
\begin{equation*}
  \risk_{L, \kdual}(\param) \defeq \inf_\eta \sup_{h \in \holderbound}
  \left\{ \frac{1}{\alpha_0} \E_P [h(X) (\lossvar - \eta)] + \eta \right\}
  \leq \risk(\theta).
\end{equation*}
As a consequence, $\risk_{L,\kdual}$ cannot upper bound the
subpopulation loss  $\E_{X\sim \pworst} [\condrisk]$ uniformly over
$\pworst \in \mixdist$.
Nonetheless, for any
subpopulation $\pworst \in \mixdist$ with Lipschitz density
$\frac{d\pworst}{dP}: \mc{X} \to \R_+$, then $\risk_{L,2}(\param)$ does
provide a valid upper bound on $\E_{X\sim \pworst} [\condrisk]$:
\begin{lemma}\label{lemma:lipriskbd}
  Let $\pworst \in \mixdist$ be any distribution with
  $L$-Lipschitz density
  $\frac{d\pworst}{dP}: \mc{X} \to \R_+$. Then
  \begin{equation*}
    \E_{X\sim \pworst} [\condrisk] \leq \risk_{L,2}(\param) \leq \risk(\theta).
  \end{equation*}
\end{lemma}
See Section~\ref{section:proof-of-lip-risk-bound} for a proof. For example,
if $P$ and $\pworst \in \mixdist$ both have Lipschitz log densities, then
$\frac{d\pworst}{dP}$ is also Lipschitz, as the following example shows.

\begin{example}
  Let $P$ be absolutely continuous with respect to some $\sigma$-finite
  measure $\mu$, denote $p(X) \defeq \frac{dP}{d\mu}(X)$ and
  $q_0(X) \defeq \frac{d\pworst}{d\mu}(X)$ and assume $\log p$ and $\log q_0$
  are $L$-Lipschitz.  Let $h(x) \defeq \frac{q_0(x)}{p(x)}$, and consider any
  fixed $x,x'\in \mathcal{X}$. If we assume that without loss of generality that
  $h(x)$ > $h(x')$, then
  \begin{align*}
    \frac{\left|h(x)- h(x')\right|}{\|x-x'\|}
%    & = \frac{h(x)\left(1-\exp(\log q_0(x') - \log q_0(x)
%      + \log p(x) - \log p(x'))\right)}{\|x-x'\|} \\
    & = \frac{h(x)}{\norm{x - x'}}
    \left(1 - \exp\Big(\log \frac{p(x)}{q_0(x)}
    - \log \frac{p(x')}{q_0(x')}\Big)\right)
    \leq \frac{L}{\alpha_0},
  \end{align*}
  where the final inequality follows because $h(x) = q_0(x) / p(x) \leq
  1/\alpha_0$ and $\exp(x) \geq 1+x$.  Consequently, then $x \mapsto
  q_0(x)/p(x)$ is $(L/\alpha_0)$-Lipschitz.
\end{example}

\subsection{Empirical Comparison of Variational Procedures}
\label{section:comparison}

% \emph{A priori} it is unclear whether one class or another in our choices
% in Section~\ref{section:procedures} should yield better performance;
% at least at this point, our theoretical understanding provides similarly
% limited guidance. To that end, we perform a small simulation study to direct
% our coming deeper theoretical and empirical evaluation,
% discussing the benefits and drawbacks of various choices of $\mc{H}$
% through the example we introduce in Figure~\ref{fig:ped1}, Section
% \ref{section:cvar}.
%The three estimators we consider are the functions defined by a Gaussian RKHS (Laplacian and Matern kernels did not qualitatively affect these results), the bounded H\"older-continuous functions, and the $L^p$ norm upper bound.

We consider an elaborated version of the data
mechanism~\eqref{eqn:toy-problem} to incorporate higher dimensionality,
with the data generating distribution
\begin{equation}
  \begin{split}
    Z \sim \bernoulli(0.15),~~
    & X_1 = (1-2Z) \cdot \uniform([0,1]),
    ~~X_2, \ldots, X_d \simiid \uniform([-1,1])\\
    &  Y = |X_1| + \indic{X_1 \ge 0} \cdot \varepsilon,
    ~~~ \varepsilon \sim \normal(0, 1).
  \end{split}
  \label{eq:simdist}
\end{equation}
\noindent
Our goal is to predict $Y$ via $\what{Y} = \theta^\top X$, and we use the
absolute loss $\loss(\theta; (x, y)) = |y - \theta^\top x|$.  We provide
details of the experimental setup, such as the estimators and optimizers,
in Section \ref{section:results}.  In brief, we perform a grid search over
all hyperparameters (Lipschitz estimates and kernel scales) for each method
over 4 orders of magnitude; for the RKHS-based estimators, we test Gaussian,
Laplacian, and Matern kernels, none of which have qualitative differences
from one another, so we present results only for the Gaussian kernel $K(x,
x') = \exp(-\frac{1}{2 \sigma^2} \ltwo{x - x'}^2)$.

\paragraph{Dimension dependence}
We first investigate the dimension dependence of the estimators, increasing
$d$ in the model~\eqref{eq:simdist} from $d = 2$ to $50$ with a fixed sample
size $N = 5000$.

\begin{figure}[h]
  \centering
  \includegraphics[scale=0.63]{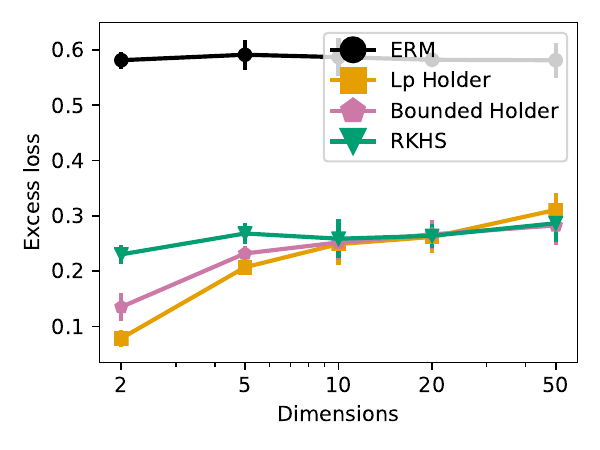}
  \caption{Variational estimates based on RKHS are less affected by the
    dimensionality of the problems, but perform worse than the H\"older
    continuous function approaches overall.}
  \label{fig:comp1}
\end{figure}

Under model~\eqref{eq:simdist}, we consider marginal distributionally robust
objective~\eqref{eq:grouploss} and evaluate the excess risk $\risk(\theta) -
\inf_\theta \risk(\theta)$ for the choice $\alpha_0 = .15$, the hardest 15\%
of the data.  As $d$ grows, we expect estimation over H\"older continuous
functions to become more difficult, and for the RKHS-based estimator to
outperform the others.  Figure~\ref{fig:comp1} bears out this intuition
(plotting the excess risk): high dimensionality induces less degradation in
the RKHS approach than the others. Yet the absolute performance of the
H\"older-based methods is better, which is unsurprising, as we are
approximating a discontinuous indicator function.

\paragraph{Sample size dependence}
We also consider the sample dependence of the estimators, fitting models
using losses with robustness level set to $\alpha_0 = .15$, then evaluating
their excess risk $\risk(\what{\theta}) - \inf_\theta \risk(\theta)$ using
$\alpha_0 \in \{.05, .15\}$, so that we can see the effects of
misspecification, as it is unlikely in practice that we know the precise
minority population size against which to evaluate. Unlike the
$L^p$-H\"older class (Eq.~\eqref{eq:ltwo-var}), the bounded $c$-H\"older
continuous function class~\eqref{eqn:holder-continuous-fns} can approximate
the population optimum of the original variational
problem~\eqref{eq:pop-var} as $n\to\infty$ and $c\to\infty$. Because of
this, we expect that as the sample size grows, and $c$ is set optimally, the
bounded H\"older class will perform well.

\begin{figure}[h]
  \subcaptionbox{$\alpha_0=0.15$ for both train and test.\label{fig:comp2}}[0.48\linewidth]{ \includegraphics[scale=0.63]{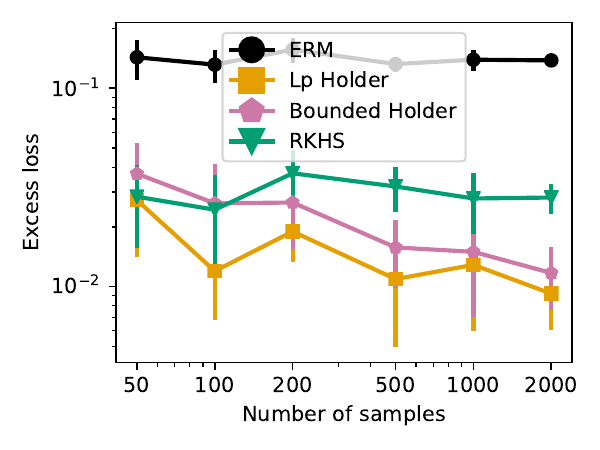}}
  \centering~
  \subcaptionbox{$\alpha_0=0.15$ for training, $\alpha_0=0.05$ for testing.
    \label{fig:comp3}}[0.48\linewidth]{
    \includegraphics[scale=0.63]{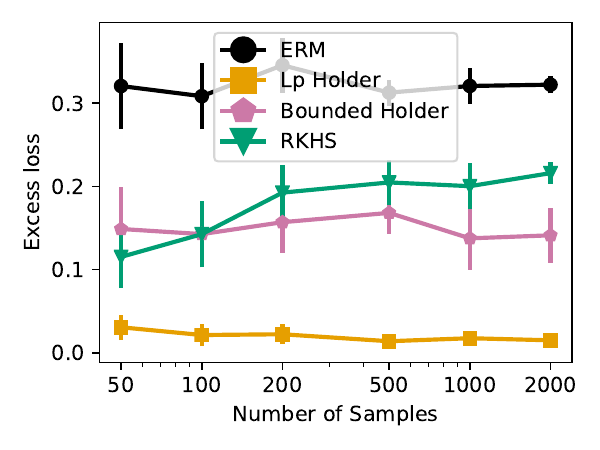}}
  \caption{Performance of the function classes in low dimensional high
    sample size settings with well-specified robustness level (left) and
    misspecified robustness level (right).}
  \ifdefined\useorstyle
  \vspace{-20pt}
  \fi
\end{figure}
On example~\eqref{eqn:toy-problem} with $d = 1$, we observe in
Figure~\ref{fig:comp2} that both H\"older continuous class estimators with
constants set via a hold-out set perform well as $n$ grows, achieving
negligible excess error. In contrast, the RKHS approach incurs high loss even
with large sample sizes. Although both H\"older class approaches perform
similarly when the robustness level $\alpha_0$ is set properly, we find that
the $L^p$-H\"older class is \emph{substantially} better when the test time
robustness level changes. The $L^p$-H\"older based estimator is the only one
which provides reasonable estimators when training with $\alpha_0=0.15$ and
testing with $\alpha_0=0.05$ (Figure \ref{fig:comp3}). Motivated by these
practical benefits, we study finite sample properties of the $L^p$-bound
estimator in this paper.

\paragraph{Real dataset}

Finally, we expand the scope of empirical evaluations by studying the wine
quality estimation experiment (Section~\ref{sec:wine}). We observe that our
proposed marginal DRO approach continues to be more accurate compared to
alternative variational approximations.

\begin{figure}
  \includegraphics[width=0.4\textwidth]{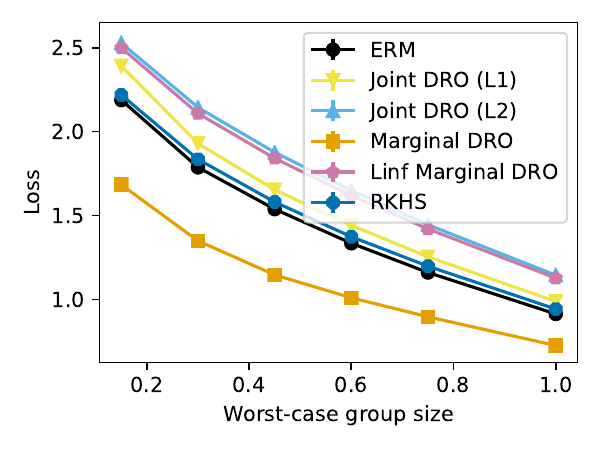}
  \centering
  \caption{Comparison of various smoothness assumptions on the conditional risk on the wine quality prediction dataset.}
  \label{fig:wine}
\end{figure}

%%% Local Variables:
%%% mode: latex
%%% TeX-master: "main"
%%% End:

\section{Risk bounds under confounding}
\label{section:confounding}

\newcommand{\confsepobj}{Z(\theta, \eta; (X, C, Y))}

% Outline:
% Start with - currently we assume pcond is fixed, but this might not be true
% concrete example - if you have confounded prediction, then this is not gonna be true.
% goal - find a model that performs uniformly well over potential confounders given the effect of confounders is small.
% we will show a generalization of the empirical estimator which provides upper bound to the loss under confounding (DEFINE) 

Our assumption that $\pcond$ is fixed for each of our marginal populations
over $X$ is analogous to the frequent assumptions in causal inference that
there are no unmeasured confounders~\citep{ImbensRu15}. When this is
true---for example, in machine learning tasks where the label $Y$ is a human
annotation of the covariate $X$---minimizing worst-case loss over covariate
shifts is natural, but the assumption may fail in other real-world problems. For
example, in predicting crime recidivism $Y$ based on the type $X$ of crime
committed and race $Z$ of the individual, unobserved confounders $C$ (e.g.\
income, location, education) likely vary with race.  Consequently, we provide
a parallel to our earlier development that provides a sensitivity analysis to
unmeasured (hidden) confounding.

Let us formalize. Let $C \in \mc{C}$ be a random variable, and in analogy
to~\eqref{eq:grouploss} we define
\begin{equation}
  \label{eq:grouploss-confounding}
  \confmixdist \defeq
  \left\{\pworst :
  \exists~ \alpha \ge \alpha_0
  ~\mbox{and~measure}
  ~\pnuisance~\mbox{on}~(\mc{X} \times \mc{C})
  ~ \mbox{s.t.} ~P_{(X, C)} = \alpha \pworst + (1-\alpha)\pnuisance
  \right\}.
\end{equation}
Our goal is then to minimize the worst-case loss under mixture covariate
shifts
\begin{equation}
  \minimize_{\param\in\paramdomain}
  \sup_{\pworst \in \confmixdist}\EX{(X, C)\sim \pworst}[ \E[\loss(\param; (X,Y))
      \mid X, C]].
  \label{eqn:worst-case-confounded-risk}
\end{equation}
Since the confounding variable $C$ is unobserved, we extend our robustness
approach assuming a bounded effect of confounding, and derive conservative
upper bounds on the worst-case loss.

%% we consider confounders
%% that have bounded effect on the conditional risk and derive a conservative
%% upper bound on the worst-case loss
%% $\sup_{\pworst \in \confmixdist}\EX{(X, C)\sim \pworst}[ \E[\loss(\param; (X,Y)) |
%% X, C]]$.
%% % and the assumption of a fixed $\pcond$ is equivalent to assuming that there
%% % are no unmeasured confounders~\cite[Section
%% % 7.4.5]{Pearl09}.  % $\confthresh$ (Assumption~\ref{assumption:bdd-confounding}).
%% As a result, unlike the unconfounded case where we showed two-sided
%% concentration results to guarantee asymptotic consistency, our confounded
%% estimator can only provide an upper bound that becomes tight in the
%% unconfounded case.

We make the following boundedness definition and
assumption on the effects of $C$.
\begin{definition}
  \label{definition:bdd-confounding}
  The triple $(X, Y, C)$ is at most
  \emph{$\confthresh$-confounded for the loss $\loss$} if
  \begin{equation*}
    \norm{\condrisk - \confcondrisk}_{L^{\infty}(P)} \le \confthresh.
  \end{equation*}
\end{definition}
\begin{assumption}
  \label{assumption:bdd-confounding}
  The triple $(X,Y,C)$ is at most $\confthresh$-confounded for the loss $\loss$.
\end{assumption}
\noindent
Paralleling earlier developments, we derive a variational bound on the
worst-case confounded risk~\eqref{eqn:worst-case-confounded-risk}. If $C = Y$,
our worst-case formulation approaches the joint DRO problem as
$\delta \to \infty$.

% \noindent For any given degree of confounding $\confthresh$, our estimator
% interpolates with our unconfounded estimator~\eqref{eq:procedure}, and a
% fully confounded estimator.

\paragraph{Confounded variational problem}
Under confounding,
a development completely parallel to Lemma~\ref{lemma:cvar}
and H\"{o}lder's inequality yields
the dual
\begin{align*}
  % \lefteqn{
  & \sup_{\pworst \in \confmixdist}
    \EX{(X, C)\sim \pworst}[ \E[\loss(\param; (X,Y)) \mid X, C]] \\ % } \\
  %% & = \inf_{\eta \in \R} \left\{
  %%   \frac{1}{\alpha_0}
  %%   \E_{(X, C) \sim P_{(X, C)}}\left[\hinge{
  %%   \E[\loss(\param; (X,Y)) | X, C] - \eta}\right]
  %%   + \eta \right\} \\
   & \le \inf_{\eta \in \R} \left\{
    \frac{1}{\alpha_0}
    \left(\E_{(X, C) \sim P_{(X, C)}}\left[\hinge{
        \E[\loss(\param; (X,Y)) \mid X, C] - \eta}^{\kdual}
      \right]\right)^{\frac{1}{\kdual}}
    + \eta \right\}
\end{align*}
for all $\kdual \ge 1$.  Taking the variational form of the $L^\kdual$-norm
for $p \in \openleft{1}{2}$ yields
\begin{equation}
  \label{eq:confoundvar}
  \begin{split}
    & \left( \E_{\pdist_{X,C}} \left[ \hinge{\confcondrisk-\eta}^\kdual
      \right]\right)^{\kdualinv} \\
    & = \sup_{h} \left\{ 
    \E\left[ h(X,C) (\confcondrisk - \eta) \right]
    ~\Big|~ h\ge 0, \E[h^\kexp(X,C)] \le 1
    \right\}.
  \end{split}
\end{equation}

Instead of the somewhat challenging variational problem over $h$, we
reparameterize problem~\eqref{eq:confoundvar} as $h(X) + f(X,C)$, where $h$ is
smooth and $f$ is a bounded residual term, which---by taking the worst case over
bounded $f$---allows us to provide an upper bound on the worst-case
problem~\eqref{eq:grouploss-confounding}.  Let $\holderbound$ be the space of
H\"older functions~\eqref{eq:lipspace} and $\bddspace$ be the space of bounded
functions
\begin{equation*}
  \bddspace \defeq \left\{
  f: \mc{X} \times \mc{C} \to \R~\mbox{measurable},
  ~\norm{f(X, C)}_{L^{\infty}(P)} \le \confthresh^{\kdual-1}
  \right\}.
\end{equation*}
Then defining the analogue of the unconfounded
variational objective~\eqref{eq:variational}
\begin{align*}
  % \label{eq:conf-variational}
  % \lefteqn{\confpopobjshort \defeq} \\
  % &
      \confpopobjshort \defeq
  \sup_{h + f \ge 0} \left\{
  \E\left[\frac{h(X) + f(X, C)}{\epsilon} (\lossvar - \eta)\right]
  \mid
  h \in \holderbound, f \in \bddspace,
  \norm{h + f}_{L^\kexp(P)} \le \epsilon
  \right\}, % \nonumber
\end{align*}
the risk $\confpopobj$ is $\epsilon$-close to the variational
objective~\eqref{eq:confoundvar}.  See Appendix~\ref{section:proof-conf} for
proof.
\begin{lemma}\label{lemma:conf}
  Let Assumptions~\ref{assumption:bdd}, \ref{assumption:loss-lip}, and
  \ref{assumption:bdd-confounding} hold. Then, for any $\theta \in \Theta$
  and $\eta \in \R$, we have
  \begin{align}
    \label{eq:resbound}
    \left( \E_{\pdist_{X,C}} \left[ \hinge{\confcondrisk-\eta}^\kdual
      \right]\right)^{\kdualinv}
    = \inf_{\epsilon \ge 0}
    \left\{ \confpopobjshort \vee \epsilon^{\kexp-1} \right\}
 \end{align}
 and for any $\epsilon > 0$,
 $ (\confpopobjshort \vee \epsilon^{\kexp-1}) - \epsilon^{\kexp-1}
   \le \left( \E_{\pdist_{X,C}} \left[ \hinge{\confcondrisk-\eta}^\kdual
   \right]\right)^{\kdualinv}.$
\end{lemma}

\paragraph{Confounded estimator}
By replacing $\holderbound$ with the empirical
version $\empholderbound$ (the set of H\"{o}lder functions on the
empirical distribution) and $\bddspace$ with
the empirical counterpart
$\mc{F}_{\confthresh,p,n} \defeq \{f \in \R^n
\mid \max_{i \le n} |f(X_i, C_i)| \le \confthresh^{p-1}\}$,
we get the obvious empirical
plug-in $\confempobjshort$ of the
population quantity $\confpopobjshort$. % ~\eqref{eq:conf-variational}.
In this case,
a duality argument provides the
following analogue of Lemma~\ref{lemma:empirical-dual},
which follows because the class $\mc{F}_{\confthresh,p,n}$ simply
corresponds to an $\linf{\cdot}$ constraint on a vector in $\R^n$.

\begin{lemma}
  \label{lemma:conf-empirical-dual}
  For any $\epsilon > 0$ and $(X_1,Y_1), \ldots, (X_n,Y_n)$, we have
  \begin{align*}
    \confempobjshort
    & = \inf_{ B \in \R^{n \times n}_+} \Bigg\{
      \bigg( \frac{\kdual-1}{n} \sum_{i=1}^n
      \hingeBig{ \loss(\param; (X_i,Y_i))
      - \frac{1}{n} \sum_{j=1}^n (B_{ij} - B_{ji}) -\eta}^\kdual
      \bigg)^{\kdualinv} \\
    & \hspace{80pt}
      + \frac{L^{\kdual-1}}{\epsilon n^2} \sum_{i,j = 1}^n \norm{X_i - X_j} B_{ij}
      + \frac{2 \confthresh^{\kdual-1}}{\epsilon n^2} \sum_{i, j = 1}^n |B_{ij}|
      \Bigg\}.
  \end{align*}
\end{lemma}
\noindent See Appendix~\ref{section:proof-conf-empirical-dual} for the
proof.  The lemma is satisfying in that it smoothly interpolates, based on
the degree of confounding $\confthresh$, between marginal distributionally robust
optimization (when $\confthresh = 0$, as in Lemma~\ref{lemma:empirical-dual}) and
the fully robust joint DRO setting as $\confthresh \uparrow \infty$, which
results in the choice $B = 0$.

\paragraph{Upper bound on confounded objective}
In analogy with Proposition~\ref{prop:fast-rate-ub}, the
empirical plug-in $\confempobjshort$ is an upper bound on the
population objective under confounding. Although our estimator only
provides an upper bound, it provides practical procedures for controlling the
worst-case loss~\eqref{eq:grouploss-confounding} when
Assumption~\ref{assumption:bdd-confounding} holds,
as we observe in the next section.
The next proposition, whose proof we provide
in Section~\ref{section:fast-conf}, shows the upper bound.

\begin{proposition}
  \label{prop:fast-rate-ub-conf}
  Let Assumptions~\ref{assumption:bdd}, \ref{assumption:loss-lip}, and
  \ref{assumption:bdd-confounding} hold.
  There exist universal constants $c_1, c_2 < \infty$ such that
  the following holds.
  Let $ \theta \in \Theta$,
  $\epsilon > 0$, and $\kdual \in \openleft{1}{2}$. Then with probability at
  least $1-2\gamma$, uniformly in $\eta \in [0, \zbound]$
  \begin{align*}
    & \left( \E_{P_{X,C}}\hinge{\confcondrisk - \eta}^\kdual
      \right)^{\kdualinv}  \\
    & \le \max\left\{\epsilon^{\kexp-1},~ (1+\tau_n(\gamma, \epsilon))^{\kexpinv}~
    \confempupperbd
    + \frac{c_1\zbound^2}{\epsilon^{\kexp-1}} \sqrt{\frac{\log\frac{1}{\gamma}}{n}}
    \right\}
  \end{align*}
  where $\tau_n(\gamma, \epsilon) \defeq
  \frac{c_2\zbound^2}{\epsilon^{\kexp-1}} \sqrt{\frac{1}{n}
    \log\frac{1}{\gamma}}$, $\confthresh_n(\gamma) \defeq \confthresh
  (1+\tau_n(\gamma, \epsilon))^{-\kexpinv}$, and $L_n(\gamma) \defeq L
  (1+\tau_n(\gamma, \epsilon))^{-\kexpinv}$.
\end{proposition}

\subsection{Simulation study: the confounded case}
\label{sec:conf-sim}

To complement our results in the unconfounded case, we extend our simulation
experiment by adding unmeasured confounders, investigating the risk upper
bounds of Lemma~\ref{lemma:conf-empirical-dual} and
Proposition~\ref{prop:fast-rate-ub-conf}.  We generate data nearly
identically to model~\eqref{eq:simdist}, introducing a
confounder $C$:
\begin{align*}
  Z \sim \bernoulli(0.15),~~
  & X_1 = (1-2Z) \cdot \uniform([0,1]),
  ~~X_2, \ldots, X_d \simiid \uniform([0,1]) \\
  &
  Y = |X_1| + \indic{X_1 \ge 0} \cdot C,
  ~~
  C \sim \uniform(\{-1,0.5,0,0.5,1\}).
\end{align*}
To evaluate a putative parameter $\theta$, we approximate the worst-case
risk 
\begin{equation}
  \label{eq:conftestval}
  R_{\alpha_0}(\theta, c) \defeq
  \sup_{\pworst \in \mc{P}_{\alpha_0,X}}
  \EX{X\sim\pworst}\E\left[|\theta^\top X - Y| \mid X, C=c\right].
\end{equation}
via the plug-in replicate estimate in Proposition~\ref{prop:repcert}, using a
sample of size $n = 2000$, $m = 10$ replicates, and test-time worst-case group
size $\alpha_0=0.05$.  The gap in the conditional risk from
$\confthresh$-confounding
(i.e.\ $\condrisk - \confcondrisk$) in
Definition~\ref{definition:bdd-confounding} varies monotonically with
$|c|$. To construct estimates $\what{\theta}$, we minimize the dual
representation of the empirical confounded risk $\confempobjshort$ in
Lemma~\ref{lemma:conf-empirical-dual}, varying the postulated level
$\confthresh$ of confounding while fixing the training time worst-case group
size $\alpha_0=0.1$.  We present results in Fig.~\ref{fig:confbound}, which
compares the worst-case loss~\eqref{eq:conftestval} as $c$ varies (vertical axis) for different
methods to select $\what{\theta}$ (horizontal axis). We compare marginal
DRO~\eqref{eq:procedure} (which assumes no confounding $\delta = 0$), the
procedure minimizing empirical confounded risk $\confempobjshort$ as we vary
$\confthresh$, and the joint DRO procedure~\eqref{eq:joint-dro-dual} (full
confounding) with $p = 1, 2$, and empirical risk minimization (ERM). 
The figure shows the (roughly) expected result that the confounding-aware risks
achieve lower worst-case loss~\eqref{eq:conftestval} as the postulated
confounding level $\confthresh$ increases with the actual amount of
confounding, with joint DRO and ERM achieving worse performance.

\begin{figure}[h]
  \centering
  \includegraphics[width=0.9\textwidth]{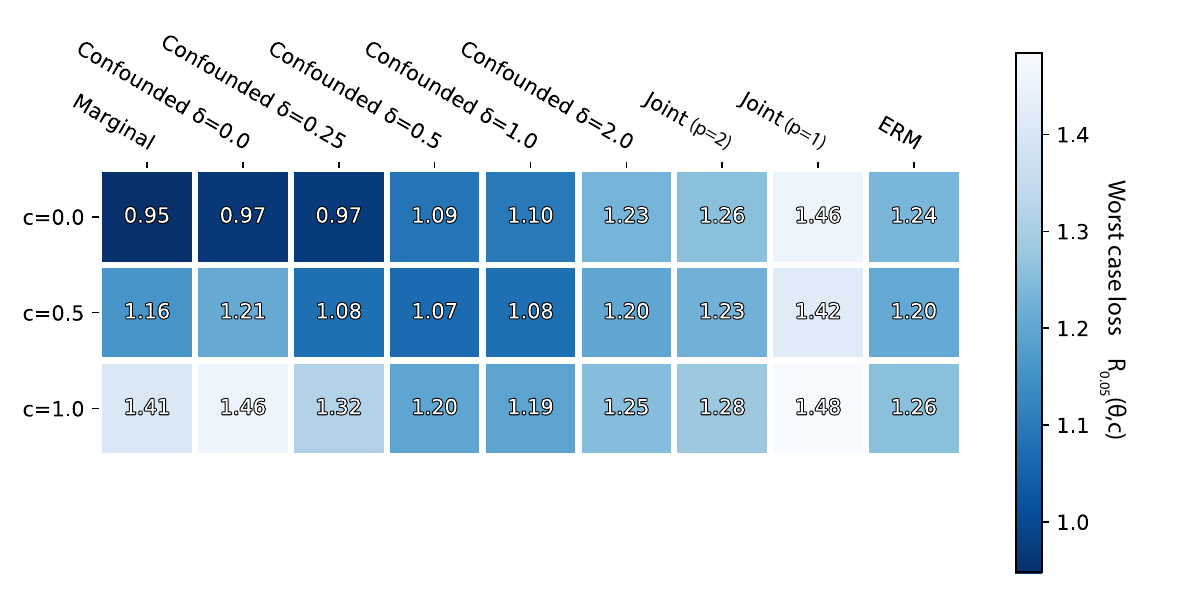}
  \caption{Worst-case losses (Eq.~\ref{eq:conftestval}) incurred by each model
    (column) when varying the influence of the unobserved confounder $c$
    (rows). Adjusting marginal DRO to account for the level of confounding
    (Lemma \ref{lemma:conf-empirical-dual}) by varying $\delta$ improves
    worst-case loss.}
  \label{fig:confbound}
  \ifdefined\useorstyle \vspace{-20pt} \fi
\end{figure}

%%% Local Variables:
%%% mode: latex
%%% TeX-master: "main"
%%% End:

\section{Additional validation experiments}
\label{section:addexpt}

In this section, we provide additional experiments quantifying the
computational overhead associated with marginal DRO, and analyze the
robustness of previous results with respect to the choice of the loss
function. Finally, we compare different smoothness assumptions on the
conditional risk on real-world data.

\subsection{Computational overhead}
\label{section:compoverhead}

To analyze the computational overhead of marginal DRO, we benchmark both the
number of gradient oracle calls (defined as the number of times we take the
gradient of the objective with respect to $\theta$ or $B$) and overall
runtime. We focus on the simulation experiments in Section~\ref{sec:simul}
with $d=10$ and $n=100$. The results in Figure~\ref{fig:runtime} show that
marginal DRO incurs a 2-3x overhead, but the overall runtime costs of all
algorithms are still low.  The gradient call comparisons show that the
increased runtimes are primarily due to the increased cost of individual
gradient calls (as each gradient computation for marginal DRO involves
gradients over both the model parameters $\theta$ as well as the transport
matrix $B$) rather then increases in the number of overall gradient steps.

\begin{figure}
  \includegraphics[width=0.4\textwidth]{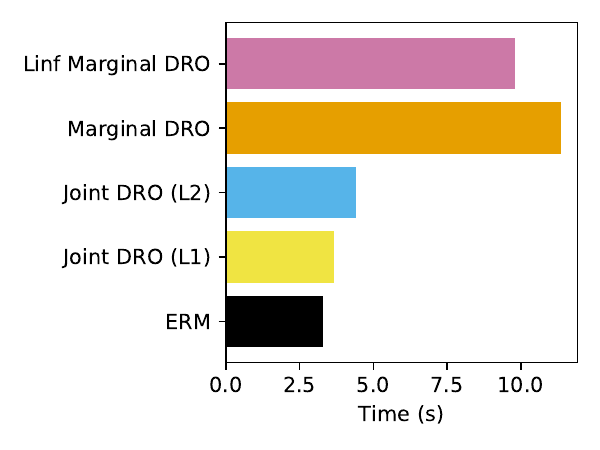}
  \includegraphics[width=0.4\textwidth]{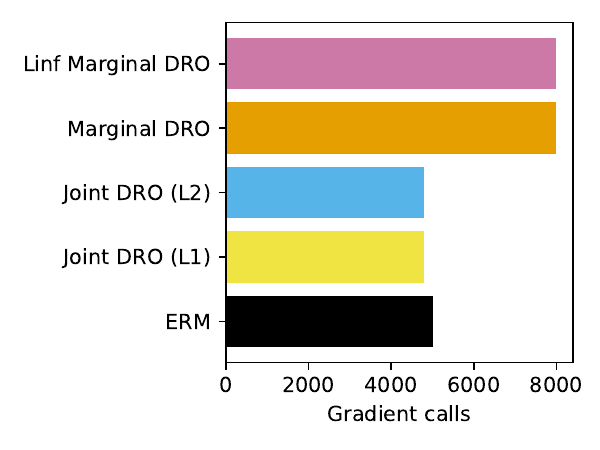}  \centering
  \caption{Runtime costs of marginal DRO and baselines}
  \label{fig:runtime}
\end{figure}

All DRO methods including joint and marginal DRO use bisection search to
optimize the dual parameter $\eta$, but this additional cost does not
substantially affect the number of gradient calls or runtime, as we re-use
initializations for gradient descent across different steps of the bisection
search. The added number of gradient calls to marginal DRO methods arise from
the fact that we must tune one additional hyperparameter $(L/\epsilon)$ which
involves additional grid search on top of the bisection search.

There is an added $2 \times$ gradient call overhead for the marginal DRO variants, which must additionally perform hyperparameter search over $L/\epsilon$ on a held out set. Even with re-using initializations, this results in a higher number of gradient oracle calls. Finally, there is an additional runtime overhead for both marginal DRO variants in terms of runtime due to the larger number of parameters.

\subsection{Effect of loss function choice}
\label{section:losschoice}

We also quantify the impact of the loss function by re-running the large scale
simulation in Figure~\ref{fig:dimdep} using the squared loss rather than the
absolute deviation loss, keeping the same experimental settings such as
validation methods and hyperparameters. We find in Figure~\ref{fig:ltwo} that
the results in the simulation remain qualitatively same as before. The main
difference is that $L_\infty$ marginal DRO performs slightly better, but none
of the ERM or joint DRO baselines perform well even in this setting. Together
with the classification results discussed in previous sections, these squared
loss results suggest that the performance gains of marginal DRO are not
limited to minimizing the absolute deviation loss.

\begin{figure}
  \includegraphics[width=0.8\textwidth]{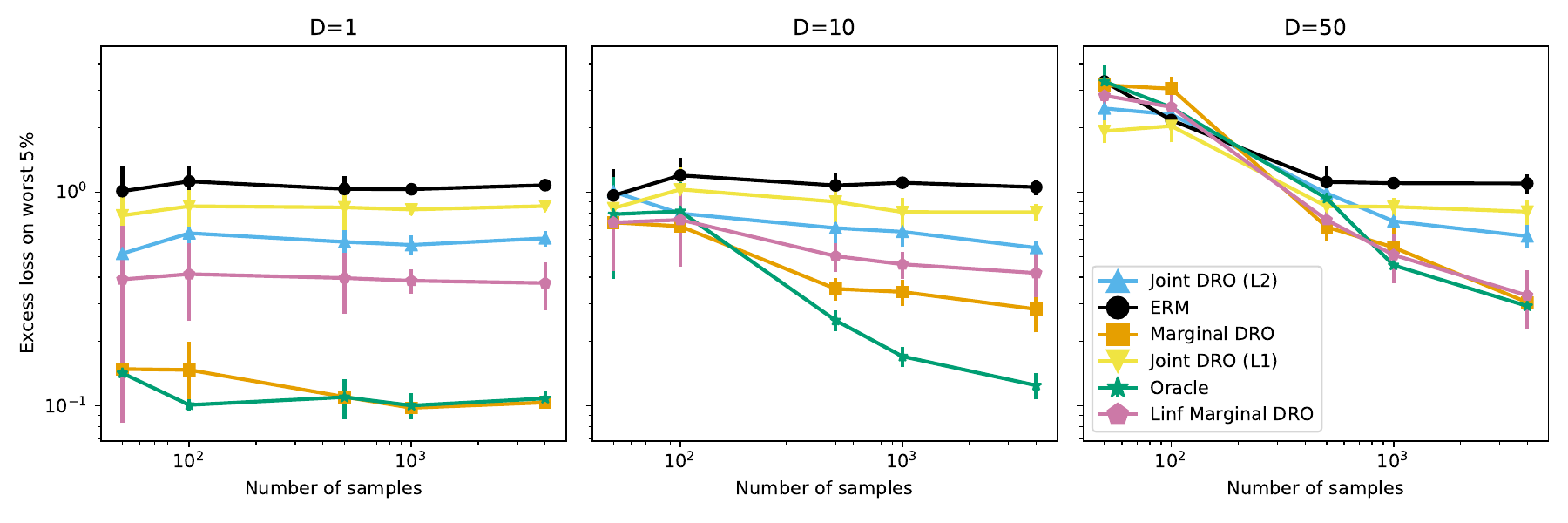}
  \centering
  \caption{Simulation experiments with the squared loss rather than absolute
    deviation loss}
  \label{fig:ltwo}
\end{figure}

%%% Local Variables:
%%% mode: latex
%%% TeX-master: t
%%% End:

% -*- mode: latex -*- %

\newcommand{\helaff}{\rho_{\textup{hel}}}
\newcommand{\opnorm}[1]{\norm{#1}_{\textup{op}}}

\section{Proofs}

% \subsection{Rough proof for h\"older continuity}
% \label{section:holder}

% We begin by showing that for any $\kdual < 2$,
% \[ \| \hinge{f(x)}^{\kdual-1} - \hinge{f(x')}^{\kdual-1}\| \leq \|\hinge{f(x)} - \hinge{f(x')}\|^{\kdual-1}.\]

% Without loss of generality let $\hinge{f(x)} \geq \hinge{f(x')}$ and define $\Delta := \hinge{f(x)} - \hinge{f(x')}$, now by subadditivity $(\hinge{f(x)}^{\kdual-1}    \leq \hinge{\hinge{f(x)}-\Delta}^{\kdual-1} + \hinge{\Delta}^{\kdual-1})$,
% \[ \| \hinge{f(x)}^{\kdual-1} - \hinge{f(x')}^{\kdual-1}\|  = \hinge{f(x)}^{\kdual-1} - \hinge{f(x)+\Delta}^{\kdual-1} \leq \hinge{\Delta}^{\kdual-1} = \|\hinge{f(x)} - \hinge{f(x')}\|^{\kdual-1}.\]

% If $\hinge{f(x)}$ is $L$-Lipschitz, this further implies that
% \[ \| \hinge{f(x)}^{\kdual-1} - \hinge{f(x')}^{\kdual-1}\|  \leq L^{\kdual-1}\|x-x'\|^{\kdual-1}\]

\subsection{Proof of Lemma~\ref{lemma:cvar}}
\label{section:proof-of-cvar}

% We first demonstrate an exact correspondence between worst-case distributions
% $\pworst$ and $L_\infty$ balls on the density ratio. For this proof we will
% let $p$ and $q$ be the density functions corresponding to the distributions
% $P$ and $Q$.
% \begin{lemma}
%   \label{lemma:boundrat}
%   For a fixed $p$, and $\alpha_0>0$,
%   \[\left\{p(y|x)q_0(x): ~q_0 \in \mixdist\right\} =
%     \left\{p(x,y)L(x) :
%       0 \leq L(x) \leq 1/\alpha_0, ~ \int L(x)dp = 1\right\}.\]
% \end{lemma}
% \begin{proof}
%   Let $q(x,y) \in \left\{p(y|x)\pworst(x): \quad q_0 \in \mixdist\right\}$.
%   Consider that the density ratio with $p(x,y)$ is bounded by 
%   $L(x) = \frac{q(x,y)}{p(x,y)} = \frac{q_0(x)}{p(x)} \leq 1/\alpha_0$. This implies  
%   $\left\{p(y|x)q_0(x): \quad q_0 \in \mixdist\right\}\subseteq
%   \left\{p(x,y)L(x) : 0 \leq L(x) \leq 1/\alpha_0, \quad \int L(x)dp =
%     1\right\}$.
    
%   In the other direction, let $q(x,y)\in \left\{p(x,y)L(x): 0 \leq L(x) \leq 1/\alpha_0, \quad \int L(x)dp = 1\right\}$. We will now explicitly
%   construct the worst case mixture using $L(x) \defeq \frac{\int q(x,y)dy}{\int p(x,y)dy}$ as
%   \begin{align*}
%     q_0(x) &= p(x)L(x)\\
%     q_1(x) &= \frac{(1-\alpha_0 L(x))p(x)}{1-\alpha_0}.
%   \end{align*}
%   The construction of $q$ implies $q_0$ is a distribution over $\mathcal{X}$, and similarly for $q_1$ due to $L(x)\leq 1/\alpha_0$. Finally, we verify that 
%   \[p(x) = \alpha_0 q_0(x) + (1-\alpha_0) q_1(x),\]
%   which implies $q_0 \in \mixdist$ which implies the equivalence of the two distribution balls.
% \end{proof}

We begin by deriving a likelihood ratio reformulation, where we use
$W \defeq \condrisk$ to ease notation
\begin{equation}
  \label{eqn:like-reform}
  \sup_{\pworst \in \mixdist} \E_{X\sim \pworst} [W]
  = \sup_{L}
  \left\{ \E_{\pdist}[L W] \Bigg|
    L: \Omega \to [0, 1/\alpha_0],~\mbox{measurable},~\E_p[L] = 1
  \right\}.
\end{equation}
To see that inequality ``$\le$'' holds, let $\pworst \in \mixdist$ be a
probability over $\mc{X}$. Note that $\pworst$ induces a distribution over
$(\Omega, \sigma(X))$, which we denote by the same notation for simplicity.
Since $\pmarg = \alpha_0 \pworst + (1-\alpha_0) \pnuisance$ for some
probability $\pnuisance$ and $\alpha_0 \in (0, 1)$, we have
$\pworst \ll \pmarg$. Letting $L \defeq \frac{d\pworst}{d\pmarg}$, it belongs
to the constraint set in the right hand side and we conclude $\le$ holds. To
see the reverse inequality ``$\ge$'', for any likelihood ratio
$L: \Omega \to [0, 1/\alpha_0]$, let $\pworst \defeq \pmarg L$ so that
$\pworst(A) \defeq \E_{\pmarg}[\indic{A} L]$ for all $A \in \sigma(X)$.
Noting
$\pnuisance \defeq \frac{1}{1-\alpha_0}\pmarg - \frac{\alpha_0}{1-\alpha_0}
\pworst$ defines a probability measure and
$\pmarg = \alpha_0 \pworst + (1-\alpha_0) \pnuisance$, we conclude that
inequality $\ge$ holds.

Next, the following lemma gives a variational form for conditional
value-at-risk, which corresponds to the worst-case loss~\eqref{eq:grouploss}
under mixture covariate shifts.
\begin{lemma}[{\citep[Example 6.19]{ShapiroDeRu09}}]
  \label{lemma:cvar-var}
  For any random variable $W: \mc{X} \to \R$ with $\E|W| < \infty$,
  \begin{align*}
    & \sup_{L}
      \left\{ \E_{\pdist}[L W] \Bigg|
      L: \Omega \to [0, \frac{1}{\alpha_0}],~\mbox{measurable},~\E_p[L] = 1
      \right\} \\
    & = \inf_{\eta \in \R} \left\{ 
      \frac{1}{\alpha_0} \E_{X\sim \pmarg}\left[\hinge{W - \eta}\right]
      + \eta \right\}.
  \end{align*}
\end{lemma}
\noindent From the reformulation~\eqref{eqn:like-reform} and
Lemma~\ref{lemma:cvar-var}, we obtain the first result.

We now show that when $W \in [0, \zbound]$,
\begin{align}
  \label{eq:cvar-bdd}
  \inf_{\eta \in \R} \left\{ 
  \frac{1}{\alpha_0} \E[\hinge{W - \eta}]
  + \eta \right\}
  = \inf_{\eta \in [0, \lbound]} \left\{ 
  \frac{1}{\alpha_0} \E[\hinge{W - \eta}]
  + \eta \right\}.
\end{align}
Noting that
$\eta \mapsto g(\eta) \defeq \frac{1}{\alpha_0} \E[\hinge{W - \eta}] + \eta$ is
strictly increasing on $\openright{\lbound}{\infty}$ since $g(\eta) = \eta$ for
$\eta \in \openright{\lbound}{\infty}$, we may assume w.l.o.g.\ that
$\eta \le \lbound$.
Further, for $\eta \le 0$, we have
\begin{align*}
  g(\eta)
  = \frac{1}{\alpha_0} \E[W] + \left(\frac{1}{\alpha_0} - 1\right) |\eta|
  \ge \frac{1}{\alpha_0} \E[W] = g(0).
\end{align*}
We conclude that the equality~\eqref{eq:cvar-bdd} holds.

% \holds{Proof of Lemma~\ref{lemma:bdd-eta}}

% Before we begin upper bounding the dual form~\eqref{eq:cvar} in ernest, we
% first note that we can restrict the range of the dual variable $\eta$ for
% bounded losses. See Appendix~\ref{section:proof-of-bdd-eta} for a proof of the
% below lemma.
% \begin{lemma}
%   \label{lemma:bdd-eta}
%   For any random variable $W \in [0, \lbound]$ and $\alpha \in (0, 1)$,
%   \begin{align}
%     \label{eq:cvar-bdd}
%     \inf_{\eta \in \R} \left\{ 
%     \frac{1}{2\alpha} \E[\hinge{W - \eta}]
%     + \eta \right\}
%     = \inf_{\eta \in [0, \lbound]} \left\{ 
%     \frac{1}{2\alpha} \E[\hinge{W - \eta}]
%     + \eta \right\}.
%   \end{align}
% \end{lemma}
% \label{section:proof-of-bdd-eta}

% Noting that
% $\eta \mapsto g(\eta) \defeq \frac{1}{2\alpha} \E[\hinge{W - \eta}] + \eta$ is
% strictly increasing on $[\lbound, \infty)$ since $g(\eta) = \eta$ for
% $\eta \in [\lbound, \infty)$, we have
% \begin{align*}
%   \inf_{\eta \in \R} \left\{ 
%   \frac{1}{2\alpha} \E[\hinge{W - \eta}]
%   + \eta \right\} 
%   = \inf_{\eta \le \lbound} \left\{ 
%   \frac{1}{2\alpha} \E[\hinge{W - \eta}]
%   + \eta \right\}.
% \end{align*}
% Further, for $\eta \le 0$, we have
% \begin{align*}
%   g(\eta)
%   = \frac{1}{2\alpha} \E[W] + \left(\frac{1}{2\alpha} - 1\right) |\eta|
%   \ge \frac{1}{2\alpha} \E[W] = g(0).
% \end{align*}
% We conclude that the equality~\eqref{eq:cvar-bdd} holds.

\subsection{Proof of Proposition~\ref{prop:repcert}}
\label{section:proof-of-repcert}

Using the dual of Lemma~\ref{lemma:cvar}, we first show that with probability
at least $1-\gamma$,
\begin{align}
  & \sup_{\eta \in [0, \lbound]} \Bigg|
  \E_{X\sim \pmarg}\left[\hinge{\condrisk - \eta}\right]
  -\frac{1}{n}  \sum_{i=1}^n
  \hingebigg{\frac{1}{m} \sum_{j=1}^m \loss(\theta; (X_i, Y_{i, j}))- \eta}
  \Bigg| \nonumber \\
  & \le C \lbound
  \sqrt{\frac{1 + \log \frac{1}{\gamma}}{\min\{m,n\}}}.
    \label{eqn:dual-concentration}
\end{align}
As the above gives a uniform approximation to the dual objective
$\frac{1}{\alpha_0} \E[\hinge{\condrisk - \eta}] + \eta$, the proposition will
then follow.

To show the result~\eqref{eqn:dual-concentration}, we begin by noting that
\begin{align}
  & \sup_{\eta \in [0, \lbound]} \Bigg|
  \E_{X\sim \pmarg}\left[\hinge{\condrisk - \eta}\right]
  - \frac{1}{n}  \sum_{i=1}^n
  \hingebigg{\frac{1}{m} \sum_{j=1}^m \loss(\theta; (X_i, Y_{i, j}))- \eta}
  \Bigg| \nonumber \\
  & \le
    \sup_{\eta \in [0, \lbound]} \Bigg|
    \E_{X\sim \pmarg}\left[\hinge{\condrisk - \eta}\right]
    - \frac{1}{n}  \sum_{i=1}^n
    \hinge{\E[\loss(\theta; (X_i, Y)) \mid X = X_i] - \eta}
    \Bigg| \nonumber \\
  & \qquad
    + \sup_{\eta \in [0, \lbound]}
    \frac{1}{n}  \sum_{i=1}^n
    \bigg|\hinge{\E[\loss(\theta; (X_i, Y)) \mid X = X_i] - \eta}
    -\hingebigg{\frac{1}{m} \sum_{j=1}^m \loss(\theta; (X_i, Y_{i, j}))- \eta}
    \bigg|.
    \label{eqn:unif-concentration}
\end{align}

To bound the first term in the bound~\eqref{eqn:unif-concentration}, note
that since $\eta \mapsto \hinge{Z - \eta}$ is $1$-Lipschitz, a standard
symmetrization and Rademacher contraction argument~\cite{BoucheronLuMa13,
  BartlettMe02} yields
\begin{equation*}
  \sup_{\eta \in [0, \lbound]} \Bigg|
  \E\left[\hinge{\condrisk - \eta}\right]
    - \frac{1}{n}  \sum_{i=1}^n
  \hinge{\E[\loss(\theta; (X_i, Y)) \mid X = X_i] - \eta}
  \Bigg|
  \le  C \sqrt{\frac{\lbound^2}{n} (1 + t)}
\end{equation*}
with probability at least $1- e^{-t}$. To bound the second term in the
bound~\eqref{eqn:unif-concentration}, we first note that
\begin{align*}
  \lefteqn{\sup_{\eta \in [0, \lbound]} 
    \left| \hinge{\E[\loss(\theta; (X, Y)) \mid X = X_i] - \eta}
    -\hingebigg{\frac{1}{m} \sum_{j=1}^m \loss(\theta; (X_i, Y_{i, j}))- \eta}
    \right|} \\
  & \le 
    \left| \E[\loss(\theta; (X_i, Y_{i, j})) \mid X = X_i] -
    \frac{1}{m} \sum_{j=1}^m \loss(\theta; (X_i, Y_{i, j}))
    \right|
\end{align*}
since $|\hinge{x-\eta} - \hinge{x'-\eta}|\le |x-x'|$. The preceding quantity
has bound $\lbound$, and using that its expectation is at most $\lbound /
\sqrt{m}$ the bounded differences inequality implies the uniform
concentration result~\eqref{eqn:dual-concentration}.

The second result follows from a nearly identical argument by noting that we
still have the Lipschitz relation
\begin{equation*}
  | \hinge{\E[\loss(\theta; X, Y) \mid X]  - \eta}
  -   \hinge{\E[\loss(\theta'; X, Y) \mid X]  - \eta'}|
  \le K \ltwo{\theta - \theta'} + |\eta - \eta'|.
\end{equation*}

\subsection{Proof of Lemma~\ref{lemma:lipriskbd}}
\label{section:proof-of-lip-risk-bound}
  Let $\mc{L}_{\alpha_0} \defeq \{ h : \mc{X} \to \R_+ \mid \E_P[h(X)] =
  \alpha_0\}$.  
  Since $\frac{d\pworst}{dP}$ is a likelihood ratio and
  $\E[d\pworst / dP] = 1$, we have the upper bound
  \begin{equation*}
    \E_{X\sim \pworst}[\condrisk] =
    \E_{P}\left[\frac{d\pworst(X)}{dP(X)}\lossvar \right]
    \leq \sup_{h \in \mc{H}_{L,2}\cap\mc{L}_{\alpha_0}}
    \E_P\left[ \frac{h(X)}{\alpha_0}\lossvar\right].
  \end{equation*}
  Then we use the sequence of inequalities, starting from our
  dual representation on $\risk_{L,2}$, that
  \begin{align*}
    \risk_{L,2}(\param)
    & = \inf_\eta \sup_{h \in \mc{H}_{L,2}} 
    \frac{1}{\alpha_0} \E_P\left[h(X)(\lossvar - \eta)\right] + \eta \\
    & \ge \sup_{h \in \mc{H}_{L,2}} \inf_\eta
    \frac{1}{\alpha_0} \E_P\left[h(X) (\lossvar - \eta) \right] + \eta \\
    & \ge \sup_{h \in \mc{H}_{L,2} \cap \mc{L}_{\alpha_0}}
    \frac{1}{\alpha_0} \E_P\left[h(X) (\lossvar - \eta)\right] + \eta
    = \sup_{h \in \mc{H}_{L,2} \cap \mc{L}_{\alpha_0}}
    \E_P[h(X) \lossvar].
  \end{align*}
  This gives the result.

\subsection{Proof of Lemma~\ref{lemma:approx-error}}
\label{section:proof-of-approx-error}

Since $Z, \eta \in [0, \zbound]$, we have
\begin{equation*}
  \E_{X\sim \pmarg}\left[\hinge{Z(X) - \eta}^\kdual \right]
  \le (\zbound - \eta)^{\kdual-1} \E_{X\sim \pmarg}\left[\hinge{Z(X) - \eta} \right]
\end{equation*}
which gives the first bound. To get the second bound, note that for a
$L$-Lipschitz function $f$, we have $\E[f(X)] \le f(\E[X]) + L\E|X - \E[X]|$.
Since $f(x) = x^p$ is $\kdual(\zbound-\eta)^{\kdual-1}$-Lipschitz on
$[0, \zbound-\eta]$, we get
\begin{align*}
  \E_{X\sim \pmarg}\left[\hinge{Z(X) - \eta}^\kdual \right]
  \le \left(\E_{X\sim \pmarg}\left[\hinge{Z(X) - \eta} \right]\right)^\kdual
  + \kdual (\zbound-\eta)^{\kdual-1} \E \left|
      \hinge{Z(X) - \eta} - \E[\hinge{Z(X) - \eta}]
      \right|.
\end{align*}
Taking $1/\kdual$-power on both sides, we obtain the second bound.

\subsection{Proof of Lemma~\ref{lemma:eps-lip-bound}}
\label{section:proof-of-eps-lip-bound}

First, we argue that
\begin{align}
  & \sup_{h} \left\{ 
    \E\left[ h(X) (\lossvar - \eta) \right]
    ~~\Big|~~ h: \mc{X} \to \R_+, ~\E[h^\kexp(X)] \le 1
    \right\} \nonumber \\
  & \le \sup_{h \in \holderbound} \left\{
    \E\left[ \frac{h(X)}{\epsilon} (\lossvar - \eta) \right]
    \vee \epsilon^{\kexp-1}~~\Bigg|~~h\ge 0,
    ~\left( \E[h^\kexp(X)] \right)^{\kexpinv} \le \epsilon
    \right\} \label{eq:eps-bound}
\end{align}
and for any $\epsilon > 0$. We consider an arbitrary but fixed $\theta$ and
$\eta$.

Suppose that
$\epsilon^{\kexp-1} \ge \left(\E_{X\sim \pmarg}\left[\hinge{\condrisk - \eta}^\kdual\right]\right)^{\kdualinv}$, then
\begin{align*}
  \epsilon^{\kexp-1}
  & \ge \left(\E_{X\sim \pmarg}\left[\hinge{\condrisk - \eta}^\kdual\right]\right)^{\kdualinv} \\
  & = \sup_{h} \left\{ 
      \E\left[ h(X) (\condrisk - \eta) \right]
      ~~\Big|~~ h: \mc{X} \to \R~\mbox{measurable},~h \ge 0,~\E[h^{\kexp}(X)] \le 1
    \right\} \\
  & \ge \sup_{h \in \holderbound}
    \left\{ 
    \E\left[ \frac{h(X)}{\epsilon} (\condrisk - \eta) \right]:
    ~h \ge 0,~
    \left( \E[h^\kexp(X)] \right)^{\kexpinv} \le \epsilon
    % ~~\Big|~~ h: \mc{X} \to \R_+,~\mbox{measurable}, \E[h^2(X)] \le 1
    \right\},
\end{align*}
and we have the upper bound. On the other hand, assume
$\epsilon^{\kexp-1} \leq \left(\E_{X\sim \pmarg}\left[\hinge{\condrisk -
      \eta}^\kdual\right]\right)^{\kdualinv}$. The inner supremum in
Eq.~\eqref{eq:ltwo-var} is attained at $h\opt$ defined in
expression~\eqref{eq:opt-hstar}, and
% \begin{equation*}
%   h\opt(x) \defeq
%   \frac{\hinge{ \E[\loss(\param, (x,Y)) \mid X=x] - \eta}^{1/(\kexp-1)}}{
%   \left(\E_{X\sim \pmarg}\left[\hinge{\condrisk - \eta}^\kdual\right]\right)^{\kexpinv}}.
% \end{equation*}
from Assumption~\ref{assumption:loss-lip}, for any $x, x' \in \mc{X}$,
\begin{align*}
  |h\opt(x) - h\opt(x')| & \le \frac{1}{\epsilon}
  \left| \hinge{\E[\loss(\param; (X,Y)) \mid X=x]-\eta}^{\kdual-1}
    - \hinge{\E[\loss(\param; (X,Y)) \mid X=x']-\eta}^{\kdual-1}
  \right| \\
  & \le \frac{1}{\epsilon}
  \left| \hinge{\E[\loss(\param; (X,Y)) \mid X=x]-\eta}
    - \hinge{\E[\loss(\param; (X,Y)) \mid X=x']-\eta}
  \right|^{\kdual-1}  \\
  & \le \frac{1}{\epsilon}
  \left|\E[\loss(\param; (X,Y)) \mid X=x]
    - \E[\loss(\param; (X,Y)) \mid X=x']
  \right|^{\kdual-1} \\
  & \le \frac{L^{\kdual-1}}{\epsilon} \norm{x - x'}^{\kdual-1},
\end{align*}
where we used
$\epsilon \le \left(\E_{X\sim \pmarg}\left[\hinge{\condrisk -
      \eta}^\kdual\right]\right)^{\kexpinv}$ in the first inequality.  Thus,
we conclude that $\epsilon h\opt$ is in $\holderbound$, and obtain the
equality
\begin{align*}
  & \left(\E_{X\sim \pmarg}\left[\hinge{\condrisk - \eta}^\kdual\right]\right)^{\kdualinv} \\
  & =  \sup_{h} \left\{
  \E_{X\sim \pmarg} \left[ (\condrisk - \eta) h(X) \right]
  \Bigg| h: \mc{X} \to \R ~\mbox{measurable}, ~h \ge 0,~ \E[h^\kexp(X)] \le 1,
  ~\mbox{and}~\epsilon h \in \holderbound
    \right\} \\
  & = \sup_{h \in \holderbound} \left\{
    \E\left[ \frac{h(X)}{\epsilon} (\lossvar - \eta) \right]
    ~~\Bigg|~~h \ge 0,~
      \left( \E[h^\kexp(X)] \right)^{\kexpinv} \le \epsilon
      \right\}
\end{align*}
where we did a change of variables $h$ to $h / \epsilon$ in the last
equality. This yields the bound~\eqref{eq:eps-bound}.

Now, for
$\epsilon = \left(\E_{X\sim \pmarg}\left[\hinge{\condrisk -
      \eta}^\kdual\right]\right)^{\kexpinv}$, the bound~\eqref{eq:eps-bound}
is actually an equality. This proves the first claim. To show the second claim,
it remains to show that
\begin{align*}
  & \sup_{h \in \holderbound} \left\{
    \E\left[ \frac{h(X)}{\epsilon} (\lossvar - \eta) \right]
    \vee \epsilon^{\kexp-1}~~\Bigg|~~ h \ge 0,~
    \left( \E[h^\kexp(X)] \right)^{\kexpinv} \le \epsilon
    \right\} - \epsilon^{\kexp-1}\\
  & \le \sup_{h} \left\{ 
    \E\left[ h(X) (\lossvar - \eta) \right]
    ~~\Big|~~ h: \mc{X} \to \R,~\mbox{measurable},~h \ge 0,
    ~\E[h^{\kexp}(X)] \le 1
    \right\}.
\end{align*}
If
$\epsilon^{\kexp-1} \ge \left(\E_{X\sim \pmarg}\left[\hinge{\condrisk -
      \eta}^\kdual\right]\right)^{\kdualinv}$, then the left hand side is less
than or equal to $0$ by the same logic above. If
$\epsilon^{\kexp-1} \le \left(\E_{X\sim \pmarg}\left[\hinge{\condrisk -
      \eta}^\kdual\right]\right)^{\kdualinv}$, then we have
\begin{align*}
  & \sup_{h \in \holderbound} \left\{
    \E\left[ \frac{h(X)}{\epsilon} (\lossvar - \eta) \right]
    \vee \epsilon^{\kexp-1} ~~\Bigg|~~h \ge 0,~
    \left( \E[h^\kexp(X)] \right)^{\kexpinv} \le \epsilon \right\} \\
  & = \sup_{h} \left\{ 
    \E\left[ h(X) (\lossvar - \eta) \right]
    ~~\Big|~~ h: \mc{X} \to \R,~\mbox{measurable}, ~h \ge 0,~\E[h^\kexp(X)] \le 1
    \right\},
\end{align*}
so the result follows.

\subsection{Proof of Lemma~\ref{lemma:empirical-dual}}
\label{section:proof-of-empirical-dual}

We take the dual of the following optimization problem
\newcommand{\lossitem}{\ell(\param;(X_i,Y_i))}
\begin{align*}
  & \maximize_{h \in \R^n}
    ~~\frac{1}{n} \sum_{i=1}^n \frac{h_i}{\epsilon} (\lossitem - \eta) \\
  & \subjectto ~~h_i \ge 0~~\mbox{for all}~~i \in [n], 
    ~~\frac{1}{n} \sum_{i=1}^n h_i^\kexp \le \epsilon^\kexp,\\
  & \hspace{45pt}
    ~~h_i - h_j \le L^{\kdual-1} \norm{X_i - X_j}^{\kdual-1}~~\mbox{for all}~~i,j \in [n]
\end{align*}
where $h_i \defeq h(X_i)$. To ease notation, we do a change of variables
$h_i \gets \frac{h_i}{\epsilon}$
\begin{align}
  \label{eq:primal}
  & \maximize_{h \in \R^n}
    ~~\frac{1}{n} \sum_{i=1}^n h_i (\lossitem - \eta) \\
  & \subjectto ~~h_i \ge 0
    ~~\mbox{for all}~~i \in [n], 
    ~~\frac{1}{n} \sum_{i=1}^n h_i^\kexp \le 1, \nonumber \\
  & \hspace{45pt}
    ~~h_i - h_j
    \le \frac{L^{\kdual-1}}{\epsilon} \norm{X_i - X_j}^{\kdual-1}~~\mbox{for all}~~i,j \in [n].
    \nonumber 
\end{align}
For $\gamma \in \R^n_+$, $\lambda \ge 0$, $B \in \R^{n \times n}_+$, the
associated Lagrangian is given by
\begin{align*}
  \mc{L}(h, \gamma, \lambda, B)
  & \defeq
    \frac{1}{n} \sum_{i=1}^n h_i (\lossitem-\eta)
    + \gamma^\top h
    + \frac{\lambda}{\kexp} \left( 1- \frac{1}{n} \sum_{i=1}^n h_i^\kexp\right) \\
  & \qquad  + \frac{1}{n^2}
    \left( \frac{L^{\kdual-1}}{\epsilon} \tr(B^\top D) - h^\top (B\onevec - B^\top \onevec)
    \right)
\end{align*}
where $D \in \R^{n \times n}$ is a matrix with entries
$D_{ij} = \norm{X_i - X_j}^{\kdual-1}$.  From strong duality, we have that the primal
optimal value~\eqref{eq:primal} is equal to
$\inf_{\gamma \in \R^n_+, \lambda \ge 0, B \in \R^{n \times n}_+} \sup_{h}
\mc{L}(h, \gamma, \lambda, B)$.

Since $h \mapsto \mc{L}(h, \gamma, \lambda, B)$ is a quadratic, a
bit of algebra shows that
\begin{align*}
  \sup_{h} \mc{L}(h, \gamma, \lambda, B)
  = \frac{\lambda}{\kexp} + \frac{L^{\kdual-1}}{\epsilon n^2} \tr (B^\top D)
  + \frac{1}{\kexp \lambda^{\kdual-1} n} \sum_{i=1}^n
  \left( \lossitem - \eta - \frac{1}{n} (B\onevec - B^\top \onevec)_i
  + \gamma_i \right)^\kdual.
\end{align*}
From complementary slackness,
\begin{align*}
  \inf_{\gamma \in \R^n_+}
    \sup_{h} \mc{L}(h, \gamma, \lambda, B)
  = \frac{\lambda}{\kexp} + \frac{L^{\kdual-1}}{\epsilon n^2} \tr (B^\top D)
  + \frac{1}{\kexp \lambda^{\kdual-1} n} \sum_{i=1}^n
  \hinge{\lossitem - \eta - \frac{1}{n} (B\onevec - B^\top \onevec)_i}^\kdual.
\end{align*}
Finally, taking infimum with respect to $\lambda \ge 0$, we obtain
\begin{align*}
  \inf_{\lambda \ge 0, \gamma \in \R^n_+}
  \sup_{h} \mc{L}(h, \gamma, \lambda, B)
  = \frac{L^{\kdual-1}}{\epsilon n^2} \tr (B^\top D)
    + \left( \frac{\kdual-1}{n} \sum_{i=1}^n
  \hinge{\lossitem - \eta
  - \frac{1}{n} (B\onevec - B^\top \onevec)_i}^\kdual \right)^{\kdualinv}.
\end{align*}
Unpacking the matrix notation, we obtain the result.

\subsection{Proof of Lemma~\ref{lemma:plug-in-rates}}
\label{section:proof-of-plug-in-rates}

From the extension theorem for H\"older continuous functions~\cite[Theorem
  1]{Minty70}, any ($\kdual-1$, $L^{\kdual-1}$)-H\"older continuous function
$h: \{X_1, \ldots, X_n\} \to \R$ extends to a ($\kdual-1$,
$L^{\kdual-1}$)-H\"older continuous $\bar{h}: \R^d \to \R$ with
$\mbox{range}(\bar{h}) \subseteq \mbox{range}(h)$ so that $h = \bar{h}$ on
$\{X_1, \ldots, X_n\}$.
% ) given by
% \begin{equation*}
%   \bar{h}(x) = \inf_{x' \in \{X_1, \ldots, X_n\}} \left\{h(x')
%     + L \norm{x - x'}\right\}
% \end{equation*}
% where $\bar{h}$ is $L$-Lipschtz.
Since $h \ge 0$ implies $\bar{h} \ge 0$, we have
\begin{align*}
  \empobjshort
  %% & = \sup_{h \in \empholderbound} \left\{
  %%   \E_{\emp} \left[ \frac{h(X)}{\epsilon} (\lossvar - \eta) \right] :
  %%   ~h \ge 0,~
  %%   \left( \E_{\emp}[h^{\kexp}(X)] \right)^{\kexpinv} \le \epsilon \right\} \\
  & = \sup_{h \in \holderbound} \left\{
    \E_{\emp} \left[ \frac{h(X)}{\epsilon} (\lossvar - \eta) \right] \mid
    ~h \ge 0,~
    \left( \E_{\emp}[h^{\kexp}(X)] \right)^{\kexpinv} \le \epsilon \right\}.
\end{align*}
To ease notation, for $c \in [0, \infty]$ define the function
$\cpopobj = R_{p,c\epsilon, L}$
%% \begin{align*}
%%   \cpopobj(\theta, \eta) \defeq
%%   \sup_{h \in \holderbound} \left\{
%%   \E\left[ \frac{h(X)}{\epsilon} (\lossvar - \eta) \right]
%%   ~~\Bigg|~~ h \ge 0,~
%%   \left( \E[h^\kexp(X)] \right)^{\kexpinv} \le c \epsilon
%%   \right\}
%% \end{align*}
so that $\popobj = \cpopobjone$. First, we establish the following
claim, which relates
$\popobj$ and $\cpopobj$.
\begin{claim}
  \label{claim:gc-bound}
  \begin{align*}
    \popobj(\theta, \eta)
    & \le \left( \frac{\epsilon}{c} \right)^{\kexp-1} \vee
    \left\{ \cpopobj(\theta, \eta) + (1-c)
      \left(\E\left[\hinge{\condrisk - \eta}^\kdual\right]\right)^{\kdualinv}
    \right\}
      ~~\mbox{if}~~ c < 1 \\
    \cpopobj(\theta, \eta)
    & \le c^{\kexp-1} \epsilon^{\kexp-1} \vee
    \left\{ \popobj(\theta, \eta) + (c-1)
      \left(\E\left[\hinge{\condrisk - \eta}^\kdual\right]\right)^{\kdualinv}
    \right\}      ~~\mbox{if}~~ c > 1.
  \end{align*}
\end{claim}
\begin{proof-of-claim}
  We only prove the bound when $c<1$ as the proof is similar when $c>1$.  In
  the case that
  \begin{equation*}
    \left( \frac{\epsilon}{c} \right)^{\kexp-1} \le
    \left(\E_{X\sim \pmarg}\left[
        \hinge{\condrisk - \eta}^{\kdual}\right]\right)^{\kexpinv},
  \end{equation*}
  the constraint sets that define $\popobj$ and $\cpopobj$ contain the
  maximizers $h\opt$ and $c h\opt$ (for $h\opt$ defined in
  expression~\eqref{eq:opt-hstar}), respectively. Hence,
  \begin{align*}
    & \popobj(\theta, \eta) = \left(\E_{X\sim \pmarg}\left[\hinge{\condrisk -
          \eta}^{\kdual}\right]\right)^{\kdualinv}
    ~\mbox{and}~\\
    & \cpopobj(\theta, \eta) = c \left(\E_{X\sim \pmarg}\left[\hinge{\condrisk -
          \eta}^\kdual\right]\right)^{\kdualinv}
  \end{align*}
  and the desired bound holds. Otherwise,
  $\popobj(\theta, \eta) \le (\epsilon/c)^{\kexp - 1}$.
  %% \begin{align*}
  %%   \epsilon^{\kexp-1} \vee \popobj(\theta, \eta)
  %%   \le \epsilon^{\kexp-1} \vee \left(\E_{X\sim \pmarg}\left[\hinge{\condrisk -
  %%         \eta}^\kdual\right]\right)^{\kdualinv}
  %%   \le \left( \frac{\epsilon}{c} \right)^{\kexp - 1}.
  %% \end{align*}
\end{proof-of-claim}

Using the two bounds in Claim~\ref{claim:gc-bound}, we now bound $\popobj$
by its empirical counterpart.
To obtain an upper bound on $\popobj$, let us first take
$c_1 \defeq (1-\what{\delta}_n)^{\kexpinv}$, where
\begin{equation*}
  \what{\delta}_n \defeq
  \frac{\kexp}{2} \wedge
  \kexp \epsilon^{-\kexp}  L^{\kdual-1}
  \left((LR)^{\kdual-1} + \epsilon\right)^{\kexp-1}
  W_{\kdual-1}(\emp, P).
\end{equation*}
Noting that $(1-\delta)^{-\kexpinv} \le 1 + \frac{4 \delta}{\kexp}$ for
$\delta \in (0, \half]$, and
$1 - (1-\delta)^{\kexpinv} \le \frac{2}{\kexp} \delta$, the first bound in
Claim~\ref{claim:gc-bound} yields for $\eta \ge 0$ that
\begin{equation}
  \label{eq:begin-first-bound}
  \epsilon^{\kexp-1} \vee \popobj(\theta, \eta)
  \le \epsilon^{\kexp-1} \vee \conepopobj(\theta, \eta)
  + 2^{\kexp-1} \epsilon + \frac{2\lbound}{\kexp}\what{\delta}_n.
\end{equation}
To bound $\conepopobj(\theta, \eta)$ by $\empobjshort$, we first note
\begin{equation}
  \label{eq:constr-swap}
  \conepopobj(\theta, \eta)
  \le \sup_{h \in \holderbound} \left\{
    \E\left[ \frac{h(X)}{\epsilon} (\lossvar - \eta) \right]
    ~~\Bigg|~~ h \ge 0,~
    \E_{\emp}[h^\kexp(X)] \le  \epsilon^\kexp
    \right\}.
\end{equation}
Indeed, for $h \in \holderbound$ satisfying
$\E_Q[h(X)^{\kexp}] \le \epsilon^{\kexp}$ for some probability measure $Q$,
$h^\kexp: \mc{X} \to \R$ is bounded by
$((LR)^{\kdual-1} + \epsilon)^{\kexp-1}$.  Hence, we have for all
$x, x' \in \mc{X}$
\begin{equation*}
  |h^\kexp(x) - h^\kexp(x')|
  \le \kexp \max\left\{h(x), h(x') \right\}^{\kexp-1}
  |h(x) - h(x')|
  \le \kexp L^{\kdual-1} \left((LR)^{\kdual-1} + \epsilon\right)^{\kexp-1}
  \norm{x-x'}^{\kdual-1}.
\end{equation*}
From the definition of the Wasserstein distance $W_{\kdual-1}$,
\begin{equation*}
  \sup_{h \in \holderbound}
  \left|
    \E_{\emp}[h^\kexp(X)] - \E[h^\kexp(X)]
  \right|
  \le \kexp L^{\kdual-1} \left((LR)^{\kdual-1} + \epsilon\right)^{\kexp-1}
  W_{\kdual-1}(\emp, P),
\end{equation*}
which implies that for any $h \in \holderbound$ satisfying
$\E[h^\kexp(X)] \le c_1^\kexp\epsilon^\kexp$
\begin{equation*}
  \E_{\emp}[h^\kexp(X)] \le \E[h^\kexp(X)] + \kexp L^{\kdual-1}
  \left((LR)^{\kdual-1} + \epsilon\right)^{\kexp-1}
  W_{\kdual-1}(\emp, P) \le \epsilon^\kexp.
\end{equation*}

To further bound the expression~\eqref{eq:constr-swap}, we check that for any
$\theta \in \Theta$ and $\eta \in [0, \lbound]$, the map
$(x, y) \mapsto \frac{h(x)}{\epsilon} (\loss(\theta; (x, y)) - \eta)$ is
H\"{o}lder continuous. By Assumption~\ref{assumption:loss-lip}, we observe
\begin{align*}
  & \left| \frac{h(x)}{\epsilon} (\loss(\theta; (x, y)) - \eta)
    - \frac{h(x')}{\epsilon} (\loss(\theta; (x', y')) - \eta)
    \right| \\
  & \le \frac{h(x)}{\epsilon}
    \left| \loss(\theta; (x, y)) - \loss(\theta; (x', y'))\right|
    + \left| \loss(\theta; (x', y')) - \eta \right|
    \frac{\left| h(x) - h(x') \right|}{\epsilon} \\
  & \le \frac{(LR)^{\kdual-1} + \epsilon }{\epsilon} L
    \norm{(x,y) - (x', y')}
    + \frac{ \lbound L^{\kdual-1} }{\epsilon} \norm{x - x'}^{\kdual-1} \\
  & = \frac{(LR)^{\kdual-1} + \epsilon}{\epsilon} LR
    \frac{\norm{(x,y) - (x', y')}}{R}
    + \frac{ \lbound L^{\kdual-1} }{\epsilon} \norm{x - x'}^{\kdual-1} \\
  & \le  \frac{(LR)^{\kdual-1} + \epsilon}{\epsilon} LR
    \frac{\norm{(x,y) - (x', y')}^{\kdual-1}}{R^{\kdual-1}}
    + \frac{ \lbound L^{\kdual-1} }{\epsilon} \norm{x - x'}^{\kdual-1} \\
  & \le \frac{1}{\epsilon}
    \left\{
    LR^{2-\kdual} ((LR)^{\kdual-1} + \epsilon) + \lbound L^{\kdual-1}
    \right\} \norm{(x, y) - (x', y')}^{\kdual-1}
\end{align*}
for all $(x, y), (x', y') \in \mc{X} \times \mc{Y}$. Using the definition of
the Wasserstein distance to bound right hand side of~\eqref{eq:constr-swap},
\begin{equation*}
  \conepopobj(\theta, \eta)
  \le \empobjshort +
  \frac{1}{\epsilon}
  \left\{ LR^{2-\kdual} ((LR)^{\kdual-1} + \epsilon) + \lbound L^{\kdual-1}
  \right\} W_{\kdual-1}(\emp, P).
\end{equation*}
Plugging in the preceding display in the bound~\eqref{eq:begin-first-bound}, we get
\begin{align*}
  \epsilon^{\kexp-1} \vee \popobj(\theta, \eta)
  & \le \epsilon^{\kexp-1} \vee \empobjshort
  + 2^{\kexp-1} \epsilon \\
  & \qquad + \zbound \left(
  1 \wedge
  2 \epsilon^{-\kexp} L^{\kdual-1}
  \left((LR)^{\kdual-1} + \epsilon\right)^{\kexp-1}
  W_{\kdual-1}(\emp, P)
  \right) \\
  & \qquad + \frac{1}{\epsilon}
  \left\{ LR^{2-\kdual} ((LR)^{\kdual-1} + \epsilon) + \lbound L^{\kdual-1}
  \right\}   W_{\kdual-1}(\emp, P).
\end{align*}

To obtain the lower bound on the empirical risk
Lemma~\ref{lemma:plug-in-rates} claims, let $c_2 \defeq (1+
\what{\delta}'_n)^{\kexpinv}$ where
\begin{equation*}
  \what{\delta}_n' = \kexp \epsilon^{-\kexp} L^{\kdual-1} \left((LR)^{\kdual-1}
    + \epsilon\right)^{\kexp-1} W_{\kdual-1}(\emp, P).
\end{equation*}
From the second bound in Claim~\ref{claim:gc-bound},
\begin{align*}
  \epsilon^{\kexp-1} \vee \ctwopopobj(\theta, \eta)
  \le \epsilon^{\kexp-1} \vee \popobj(\theta, \eta)
  +  \left( \frac{\epsilon^{\kexp-1}}{\kdual}
  + \frac{\lbound}{\kexp} \right) \what{\delta}_n'
\end{align*}
holds, and from a similar argument as before, we have
\begin{align*}
  \ctwopopobj(\theta, \eta)
  & \ge \sup_{h \in \holderbound} \left\{
    \E\left[ \frac{h(X)}{\epsilon} (\lossvar - \eta) \right]
    ~~\Bigg|~~ h \ge 0,~
    \E_{\emp}[h^\kexp(X)] \le  \epsilon^\kexp
    \right\} \\
  & \ge \empobjshort -  \frac{1}{\epsilon}
    \left\{ LR^{2-\kdual} ((LR)^{\kdual-1} + \epsilon) + \lbound L^{\kdual-1}
    \right\} W_{\kdual-1}(\emp, P).
\end{align*}
We conclude that
\begin{align*}
  \epsilon^{\kexp-1} \vee \empobjshort
  & \le \epsilon^{\kexp-1} \vee \popobj(\theta, \eta) \\
  & \qquad + 
    \left( (\kexp-1) L^{\kdual-1} \epsilon^{-1}
      + \lbound \epsilon^{-\kexp} L^{\kdual-1} \right)
     \left((LR)^{\kdual-1}
    + \epsilon\right)^{\kexp-1}W_{\kdual-1}(\emp, P) \\
  & \qquad
    + \frac{1}{\epsilon}
    \left( LR^{2-\kdual} ((LR)^{\kdual-1} + \epsilon) + \lbound L^{\kdual-1}
    \right) W_{\kdual-1}(\emp, P).
\end{align*}

\subsection{Proof of Theorem~\ref{theorem:main-thm}}
\label{section:proof-of-main-thm}

We use the following concentration result for the Wasserstein distance between
an empirical distribution and its population counterpart. We abuse notation
and denote by $c_1$ and $c_2$ constants that may change from line to line.
\begin{lemma}[\citet{FournierGu15}, Theorem 2]
  \label{lemma:wass-conv}
  Let $\kdual \in (1, 2]$ and $p-1 < \frac{d+1}{2}$.  Then for any $t > 0$,
  \begin{align*}
    \P\left( W_{p-1}(P, \emp)  \ge t \right) \le
      c_1\exp\left( - c_2 n (t ^{\frac{d+1}{\kdual-1}} \wedge t^2) \right)
  \end{align*}
  where $c_1$ and $c_2$ are positive constants that depend on
  $\lbound, d, \kdual$.
\end{lemma}
\noindent See~\citet{FournierGu15} and~\citet{Lei18} for general
concentration results.

% Recall the definition of the Orlicz $\psi_1$-norm
% \begin{equation*}
%   \norm{W}_{\psi_1} \defeq \inf \left\{ c > 0: \E\left[\exp\left(
%         \left| \frac{W}{c}\right|
%       \right)\right] \right\}.
% \end{equation*}
% \begin{lemma}[{\citep[Theorem 3.1 and Corollary 5.2]{Lei18}}]
%   \label{lemma:wass-conv}
%   Let $X$ and $Y$ be subexponential random variables so that
%   $\norm{X}_{\psi_1} <\infty$ and $\norm{Y}_{\psi_1} < \infty$. For $d > 2$,
%   for a constant $c_1 > 0$ that only depends on
%   $\norm{X}_{\psi_1}, \norm{Y}_{\psi_1}$ and a univeral constant $c_2 > 0$, 
%   we have
%   \begin{equation*}
%     \P(W_1(\emp, P) \ge t)
%     \le \exp\left( - c_1 n \hinge{t - c_2 n^{-1/d}}^2 \right).
%   \end{equation*}
% \end{lemma}
% For heavier tailed random variables, see~\citep{FournierGu15}.

% \begin{lemma}[{\citep[Theorem 2]{FournierGu15}}]
%   \label{lemma:wass-conv}
%   Assume that $\E\norm{X}^q + \E\norm{Y}^q < \infty$ for some $q > 2$.  Then,
%   for all $n \ge 1$, $t \in (0, \infty)$ and $\delta \in (0, q)$
%   \begin{equation*}
%     \P\left( W_1(P, \emp)  \ge t \right)
%     \le C_2 \left( a(N, t) \indic{t \le 1} + N (Nx)^{-(q-\delta)} \right)
%   \end{equation*}
%   where $C_1$ and $C_2$ are positive constants that depend on
%   $(\delta, q, d)$ and
%   \begin{align*}
%     a(N, t) \defeq
%     \begin{cases}
%       \exp\left(-C_1 n t^2\right) & ~~\mbox{if}~~ d=1 \\
%       \exp\left(-C_1 n \left( \frac{t}{\log(2+1/t)} \right)^2\right)
%       & ~~\mbox{if}~~ d=2 \\
%       \exp\left(-C_1 n t^d\right)
%       & ~~\mbox{if}~~ d\ge 3
%     \end{cases}.
%   \end{align*}
% \end{lemma}

Let
$B_{\epsilon} \defeq LR + \epsilon^{-1} 2^{\kexp-1} L\left( 2\lbound +
  (\kexp-1)LR \right) + \epsilon^{-\kexp} 2^{\kexp-1} R\lbound L^2 +
\epsilon^{\kexp-2} (\kexp-1) 2^{\kexp-2} L$ to ease notation. From
Lemmas~\ref{lemma:plug-in-rates} and~\ref{lemma:wass-conv}, for any fixed
$\epsilon > 0$, with probability at least
$1-\frac{\gamma}{2}$
\begin{align*}
  \sup_{\pworst(x) \in \mixdist}
    \E_{X\sim \pworst}[ \E[\loss(\empparam; (X,Y)) \mid X] ]
    & \le \inf_{\eta \in [0, \lbound]} \left\{ 
    \frac{1}{\alpha_0 }
    \left( \popobj(\empparam, \eta) \vee \epsilon^{\kexp-1}\right) + \eta
    \right\}  \\
  & \le \inf_{\eta \in [0, \lbound]} \left\{ 
    \frac{1}{\alpha_0 }
    \left( \empobjnoarg(\empparam, \eta)
    \vee \epsilon^{\kexp-1}\right) + \eta
    \right\} +  \frac{B_{\epsilon}t}{\alpha_0}  \\
  & \le \inf_{\eta \in [0, \lbound]} \left\{ 
    \frac{1}{\alpha_0 }
    \left( \empobjnoarg(\theta, \eta)
    \vee \epsilon^{\kexp-1}\right) + \eta
    \right\} +  \frac{B_{\epsilon}t}{\alpha_0}
\end{align*}
for any $\theta \in \Theta$, where we used the fact that $\empparam$ is an
empirical minimizer.

Applying uniform convergence of $\empobjshort$ to $\popobj$ again
(Lemmas~\ref{lemma:plug-in-rates} and~\ref{lemma:wass-conv}), we get
\begin{align*}
  & \sup_{\pworst(x) \in \mixdist}
    \E_{X\sim \pworst}[ \E[\loss(\empparam; (X,Y)) \mid X] ]  \\
  & \le \inf_{\eta \in [0, \lbound]} \left\{ 
    \frac{1}{\alpha_0 }
    \left( \popobj(\theta, \eta) \vee \epsilon\right) + \eta
    \right\} +  \frac{2B_{\epsilon}t}{\alpha_0}  \\
  & \le \inf_{\eta \in [0, \lbound]} \left\{ 
    \frac{1}{\alpha_0 }
    \left( \E_{X\sim \pmarg}\left[\hinge{\condrisk - \eta}^\kdual\right]
    \right)^{\kdualinv}
    + \eta
    \right\} + \frac{\epsilon^{\kexp-1}}{\alpha_0}
    + \frac{2B_{\epsilon}t}{\alpha_0} 
\end{align*}
with probability at least
$1-\gamma$, where we
used the second bound of Lemma~\ref{lemma:eps-lip-bound}.  Taking infimum over
$\theta \in \Theta$, we obtain the result.

\subsection{Proof of Proposition~\ref{prop:fast-rate-ub}}
\label{section:proof-of-fast-rate-ub}

Since our desired bound holds trivially if
$ \left( \E\hinge{\condrisk - \eta}^\kdual\right)^{\kdualinv} \le
\epsilon^{\kexp-1}$, we assume
$ \left( \E\hinge{\condrisk - \eta}^\kdual\right)^{\kdualinv} \ge
\epsilon^{\kexp-1}$. First, we rewrite the left hand side as
\begin{align*}
  \left( \E\hinge{\condrisk - \eta}^\kdual\right)^{\kdualinv}
  = \frac{\E\hinge{\condrisk - \eta}^\kdual}
  {  \left( \E\hinge{\condrisk - \eta}^\kdual\right)^{\kexpinv}}  
  = \frac{\E[\sepobj]}
  {\left( \E\hinge{\condrisk - \eta}^\kdual\right)^{\kexpinv}}
\end{align*}
where for convenience we defined
\begin{equation*}
  \sepobj \defeq (\lossvar - \eta)\hinge{\condrisk - \eta}^{\kdual-1}.
\end{equation*}
Now, note that $\eta \mapsto \sepobj$ is $\kdual\zbound$-Lipschitz.  Applying
a standard bracketing number argument for uniform concentration of Lipschitz
functions~\cite[Theorem 2.7.11]{VanDerVaartWe96}
\begin{align*}
  \sup_{\eta \in [0, \zbound]}
  \left|
  \E[\sepobj] - \E_{\emp}[\sepobj]
  \right|
  \le c_1 M^2 \sqrt{\frac{\log\frac{1}{\gamma}}{n}}
\end{align*}
with probability at least $1-\gamma$, where $c_1$ is some universal constant.
We conclude that with probability at least $1-\gamma$, for all
$\eta \in [0, \zbound]$
\begin{align}
  \left( \E\hinge{\condrisk - \eta}^\kdual\right)^{\kdualinv}
  & \le \left( \E\hinge{\condrisk - \eta}^\kdual\right)^{-\kexpinv}
  \E_{\emp}[\sepobj] \nonumber \\
  & \qquad + \left( \E\hinge{\condrisk - \eta}^\kdual\right)^{-\kexpinv}
    c_1 M^2 \sqrt{\frac{\log\frac{1}{\gamma}}{n}}.
    \label{eq:right-concentration}
\end{align}

Next, we upper bound the first term by our empirical objective
$\empobjshort$
\begin{align*}
  & \left( \E\hinge{\condrisk - \eta}^\kdual\right)^{-\kexpinv}
    \E_{\emp}[\sepobj] \\
  & = (1+\tau_n(\gamma, \epsilon))^{\kexpinv}
    ~\E_{\emp}\left[ \frac{h_{\eta}\opt(X)}{(1+\tau_n(\gamma, \epsilon))^{\kexpinv}}
    (\lossvar - \eta) \right],
\end{align*}
where we used the definition of $\condrisk$ in Eq.~\eqref{eq:opt-hstar}
(we now make the dependence on $\eta$ explicit). Uniform concentration of
Lipschitz functions~\cite[Theorem 2.7.11]{VanDerVaartWe96} implies that there
exists a universal constant $c_{2}> 0$ such that with probability at least
$1-\gamma$
\begin{equation*}
  \E_{\emp}\hinge{\condrisk - \eta}^\kdual
  \le \E\hinge{\condrisk - \eta}^\kdual
  + c_{2} \zbound^2 \sqrt{\frac{1}{n}\log\frac{1}{\gamma}}
\end{equation*}
for all $\eta \in [0, \zbound]$.  Thus, we have
\begin{equation}
  \label{eq:second-mom-hopt}
  \E_{\emp}[h_{\eta}\opt(X)^\kexp] \le 1 + c_{2}\zbound^2
  \left( \E\hinge{\condrisk - \eta}^\kdual\right)^{-1}
  \sqrt{\frac{1}{n}\log\frac{1}{\gamma}}.
\end{equation}
with probability at least $1-\gamma$.

Recalling the definition~\eqref{eq:emplipspace} of $\empholderbound$, since
$x \mapsto h_{\eta}\opt(x)$ is $\frac{L}{\epsilon}$-Lipschitz, we get
\begin{align*}
  & \left( \E\hinge{\condrisk - \eta}^\kdual\right)^{-\kexpinv}
    \E_{\emp}[\sepobj] \\
  & \le (1+\tau_n(\gamma, \epsilon))^{\kexpinv}
    \sup_{h \in \mc{H}_{L_n(\gamma), n}}
    \left\{ \E_{\emp}\left[ \frac{h(X)}{\epsilon} (\lossvar - \eta)\right]
    ~~\Bigg|~~
    \E_{\emp}[h^\kexp(X)] \le \epsilon^\kexp
    \right\}
\end{align*}
with probability at least $1-\gamma$, where we used the
bound~\eqref{eq:second-mom-hopt} in the second inequality.  Combining the
preceding display with the bound~\eqref{eq:right-concentration}, with
probability at least $1-2\gamma$,
\begin{align*}
  \left( \E\hinge{\condrisk - \eta}^\kdual\right)^{\kdualinv}
  \le (1+\tau_n(\gamma, \epsilon))^{\kexpinv}~
    \empupperbd
    + \frac{c_1\zbound^2}{\epsilon^{\kexp-1}} \sqrt{\frac{1}{n}\log\frac{1}{\gamma}}
\end{align*}
for all $\eta \in [0, \zbound]$.

To show the uniform result over $\theta \in \Theta$, we note that
\begin{equation*}
  |Z(\theta, \eta; X, Y) -   Z(\theta', \eta'; X, Y)|
  \le pM |\eta - \eta'| + p M^{p-1}K \ltwo{\theta - \theta'}~\mbox{for all}~\eta \in [0, M], \theta \in \Theta.
\end{equation*}
Setting
$\tau_n \defeq c_2(\zbound^2 +  pM^{p-1} KD \epsilon^{-\kexp} \sqrt{\frac{1}{n}
  \log\frac{1}{\gamma}}$, a similar argument as above, \emph{mutandis mutatis},
gives the desired result.

\subsection{Proof of Theorem~\ref{theorem:lower-bound}}
\label{section:proof-lower-bound}

We follow the frequent approach in modern statistics of reducing
esitmation problems to testing problems, then applying information-theoretic
lower bounds on test error rates, by
reducing the minimax marginal DRO problem
to a composite hypothesis testing problem between two classes
of distributions $\mc{P}_0$ and $\mc{P}_1$.
Following the approach \citet[Ch.~5]{Duchi18} suggests,
define the optimization distance
between two distributions $P_0$ and $P_1$ by
\begin{align*}
  \distopt{P_0}{P_{1}} \defeq 
  \sup \left\{ \delta \ge 0 \mid
  \begin{aligned}
    & \risk(\param; P_0) \le \risk(\param_0^*; P_0) + \delta
    \mbox{~~implies~~}
      \risk(\param; P_1) \ge \risk(\param_{1}^*; P_1) + \delta \\
      & \risk(\param; P_1) \le \risk(\param_{1}^*; P_1) + \delta
    \mbox{~~implies~~}
    \risk(\param; P_0) \ge \risk(\param_{0}^*; P_0) + \delta
  \end{aligned}
  \right\}
\end{align*}
where $\param_v^*\in \argmin_{\param \in \paramdomain} \risk(\param; P_v)$.
We have the following reduction from distributionally robust optimization to
composite hypothesis testing. Its proof is similar to the Le Cam's convex hull
method for estimation---we give it in Section~\ref{section:proof-lecam} for
completeness.
\begin{lemma}[{\cite{LeCam73, Yu97}}]
  \label{lemma:lecam}
  Let $\mc{P}_0, \mc{P}_1 \subseteq \mathfrak{P}_{\holdersmooth}$ be two sets
  of distributions such that $\distopt{P_0}{P_1} \ge 2\delta$ for all
  $P_v \in \mc{P}_v$, $v \in \{0, 1\}$. Then,
  \begin{align*}
    \mc{M}_n
    & \ge \delta \cdot \sup\left\{ 1 - \tvnorm{\bar{P}_0 - \bar{P}_1}:
      \bar{P}_v \in \conv(\mc{P}_v^n), v \in \{0, 1\} \right\}
  \end{align*}
  where $\mc{P}_v^n$ is the set of $n$-product distributions of $\mc{P}_v$ and
  $\conv(\cdot)$ denotes the convex hull of a set.
\end{lemma}

Our approach using Lemma~\ref{lemma:lecam} is then apparent: we construct
families of distributions $\mc{P}_0$  and $\mc{P}_1$ such that their
optimization distances are large, while their variation distances are
small enough that testing between them is impossible.
For simplicity in what follows, we restrict attention to
odd-valued dimensions; the result for even-valued dimensions follow from an
identical construction where we do not consider any variation in the last
covariate dimension, so that the effective dimension of the problem is
$d-1$. We divide the remainder of the proof into preliminaries, separation,
and closeness in variation distance.

\subsubsection*{Preliminaries}

We always consider a uniformly distributed covariate vector
$X \sim \uniform[0, 1]^d$. Our
construction proceeds by concatenating a large number of
\emph{bump} functions together (across dimensions
and space $[0, 1]^d$). In general, we can allow any
differentiable function $\varphi : [0, 1] \to \R$ satisfying
$\linf{\varphi} \le 1$ and $-\varphi(x) = \varphi(1 - x)$,
so that $\int_0^1 \varphi(x) dx = 0$, though to address our
smoothness desiderata we make the specific choice
\begin{equation}
  \label{eqn:bump}
  \varphi(x) \defeq
  \begin{cases}
    \left( 1- (4x-1)^2 \right)^{\holdersmooth} & \mbox{for}~
    0 \le x \le \half \\
    - \left( 1- (4x-3)^2 \right)^{\holdersmooth} &
    \mbox{for}~ \half \le x \le 1.
  \end{cases}
\end{equation}
It is immediate that $\varphi \in C^{\holdersmooth}([0,1])$ and
$\varphi(x) = -\varphi(1 - x)$. Given this function, we can
define the product function
\begin{equation*}
  g : [0, 1]^d \to \R,
  ~~
  g(x) \defeq \prod_{k = 1}^d \varphi(x^k)
\end{equation*}
and let $\sigma^2(\beta, d)
\defeq \int g(x)^2 dx
= (\int_0^1 \varphi^2(u) du)^d$. Letting
\begin{equation*}
  q_{1-\alpha_0} \defeq \inf\{ q \mid \P(g(X) \le q ) \ge 1-\alpha_0\}
\end{equation*}
be the $(1-\alpha_0)$-th quantile of $g(X)$ for $X \sim \uniform[0,1]^d$,
the symmetry of $\varphi$ guarantees that
$q_{1 - \alpha_0} > 0$ whenever $\alpha_0 < \half$. We may then
define the tail average
\begin{equation*}
  \Delta_{\alpha_0, \holdersmooth, d}
  \defeq \int_{[0, 1]^d} g(x) \indic{g(x) \ge q_{1-\alpha_0}} dx,
\end{equation*}
which because of the bump construction~\eqref{eqn:bump} depends only on
$\alpha_0, \holdersmooth$, and $d$, and by symmetry of $\varphi$ we have
$\Delta_{\alpha_0, \beta, d} > 0$ for any $\alpha_0$.
Our coming construction of $\mc{P}_0$ and $\mc{P}_1$ will show the following
result: there exist $c, N$ depending on $d, \holdersmooth, \alpha_0$ only
such that for
$n \ge N$,
\begin{equation}
  \label{eqn:real-lower-bound}
  \mc{M}_n \ge c \frac{\Delta_{\alpha_0, \holdersmooth,d}}{\alpha_0}
  \left( \sigma^2(\holdersmooth,d) n \right)^{\frac{-2\holdersmooth}{2\holdersmooth + d}}.
\end{equation}

As a brief remark, the terms $\Delta_{\alpha_0,\holdersmooth,d}$
and $\sigma^2(\holdersmooth, d)$ depend strongly on the dimension; indeed,
if $\sigma^2_\varphi = \int_0^1 \varphi^2(u) du < 1$, then
$\sigma^2(\holdersmooth,d) = \sigma^{2d}_\varphi$. Similarly,
$\varphi(X^k)$ is a symmetric
random variable on $[-1, 1]$ (and so sub-Gaussian); the product
$g(X) = \prod_{k = 1}^d \varphi(X^k)$ then concentrates very quickly to zero, and
moreover, we have
$q_{1 - \alpha_0} \le \E[|g(X)| \mid g(X) \ge q_{1 - \alpha_0}]
= \E[|g(X)| \indic{g(X) \ge q_{1 - \alpha_0}}] / \alpha_0
\le \E[|g(X)|] / \alpha_0$, and $\E[|g(X)|] = \E[|\varphi(X^1)|]^d$ where
$\E[|\varphi(X^1)|] < 1$. (A concentration argument for
$\log |g(X)| = \sum_{k = 1}^d \log |\varphi(X^k)|$ shows that this
exponential scaling is tight to within the factor $1/\alpha_0$.)
By
considering problems of smaller dimension $k \le d$ (and ignoring all
higher dimensions)
we can replace the bound~\eqref{eqn:real-lower-bound}
by the bound
\begin{equation*}
  \max_{k \le d} \frac{\Delta_{\alpha_0,\holdersmooth,k}}{\alpha_0}
  \left(\sigma^{2k}_\varphi n \right)^\frac{-2 \beta}{2 \beta + k},
\end{equation*}
which allows finite sample guarantees for smaller $n$.

\subsubsection*{Separation in objectives}

We now construct the families of distributions $\mc{P}_0$ and
$\mc{P}_1$ to guarantee sufficient optimization distance
separation $\distopt{P_0}{P_1}$ in Lemma~\ref{lemma:lecam}.
Consider hyperrectangles formed by partitioning each side in the hypercube
$[0, 1]^d$ as
$\left\{ \left[\frac{l-1}{b}, \frac{l}{b} \right] \right\}_{l = 1}^b$ for some
$b \in \N$ to be chosen later. Using a lexicographic ordering, denote the
hyperrectangles as
\begin{equation}
  \label{eqn:rectangles}
  R_j \defeq \prod_{k=1}^d \left[ \frac{l_{jk}-1}{b}, \frac{l_{jk}}{b}\right],
  ~j=1, \ldots, b^d \eqdef m
\end{equation}
where $l_{jk}$'s are defined implicitly. For each
of these $m = b^d$ hyperrectangles,
define the localized bump function
$g_j$ on $R_j$ by
\begin{align}
  g_j(x) \defeq \prod_{k=1}^d \varphi\left(
  b \left( x^k - \frac{l_{jk}-1}{b} \right) \right)
  \indic{x^k \in \left[ \frac{l_{jk}-1}{b}, \frac{l_{jk}}{b}\right]},
  ~j=1, \ldots, m = b^d.
  \label{eqn:gj-def}
\end{align}
Now, fix $t > 0$, to be chosen later when we optimize our
separation. Recalling that $X \sim \uniform[0, 1]^d$, let $P_0$ be such that
\begin{align}
  \label{eqn:lower-P0-dist}
  Y\mid X = -\frac{t\Delta_{\alpha_0, \holdersmooth, d}}{2\alpha_0}
  + \begin{cases}
    1 &~\mbox{w.p.}~\half \\
    - 1 &~\mbox{w.p.}~\half
    \end{cases}~~~\mbox{under}~P_0,
\end{align}
and let the distributions $P_{tv}$ indexed by $v \in \{-1, +1\}^m$
be
\begin{equation*}
  Y\mid X = -\frac{t\Delta_{\alpha_0, \holdersmooth, d}}{2\alpha_0}
  + \begin{cases}
    1 &~\mbox{w.p.}~\half + \frac{t}{2} \sum_{j=1}^m v_j g_j(x) \\
    - 1 &~\mbox{w.p.}~\half -  \frac{t}{2} \sum_{j=1}^m v_j g_j(x)
  \end{cases}~~~  \mbox{under}~P_{tv}.
\end{equation*}
By inspection, we have
\begin{equation*}
  \E_{P_0}[Y \mid X] \equiv - \frac{t\Delta_{\alpha_0, \holdersmooth, d}}{2\alpha_0}
  ~~\mbox{and}~~ \E_{P_{tv}}[Y \mid X] = - \frac{t\Delta_{\alpha_0, \holdersmooth, d}}{2\alpha_0}
  + t \sum_{j=1}^m v_j g_j(X).
\end{equation*}
We will use Le Cam's convex hull method (Lemma~\ref{lemma:lecam}) on the classes
of distributions
\begin{equation*}
  \mc{P}_0 = \{P_0\},~~\mc{P}_1 = \{P_{tv}: v \in \{\pm 1\}^m\}.
\end{equation*}

The next lemma allows us to show
separation between the sets $\mc{P}_0$ and $\mc{P}_1$ in
optimization distance.
See Section~\ref{section:proof-cvar-bound} for a proof.
\begin{lemma}
  \label{lemma:cvar-bound}
  Let $\varphi : [0, 1] \to \R$ be a differentiable function such that
  $\norm{\varphi}_{\infty} \le 1$ and $-\varphi(x^1) = \varphi(1-x^1)$, and
  let $g_j$ be the localized products~\eqref{eqn:gj-def}. If
  $d$ is odd, then
  \begin{equation*}
    \sup_{\pworst \in \mixdist(P)}\EX{X\sim \pworst}\left[
      \sum_{j=1}^m v_j g_j(X)\right]
    \ge \frac{\Delta_{\alpha_0, \holdersmooth, d}}{\alpha_0}.
  \end{equation*}
\end{lemma}

We claim that Lemma~\ref{lemma:cvar-bound} guarantees the
separation
\begin{equation}
  \label{eqn:dopt-sep}
  \distopt{P_{tv}}{P_0} \ge \frac{t\Delta_{\alpha_0, \holdersmooth, d}}{4\alpha_0}
  ~~~\mbox{for all}~v \in \{-1, +1\}^m.
\end{equation}
To see this, note that under $P_0$, we have
$\E_{P_0}[Y \mid X] = -t \frac{\delta_{\alpha_0, \holdersmooth, d}}{2 \alpha_0}$,
independent of $X$, and so
$\risk(\theta; P_0) = -\theta\frac{t\Delta_{\alpha_0, \holdersmooth, d}}{2\alpha_0}$.
For the set $\mc{P}_1$, we observe that any $P_{tv} \in \mc{P}_1$ has
\begin{align*}
  \risk(\theta; P_{tv})
  = \sup_{\pworst \in \mixdist(P)}
  \E_{X \sim \pworst}\left[\theta \E_{P_{tv}}[Y \mid X]\right]
  & = \sup_{\pworst \in \mixdist(P)}
  \E_{X \sim \pworst}\left[\theta\left(-\frac{t \Delta_{\alpha_0, \holdersmooth, d}}{2
      \alpha_0} + t \sum_{j = 1}^m v_j g_j(X)\right)\right] \\
  & \stackrel{(\star)}{\ge}
  \theta \frac{t\Delta_{\alpha_0, \holdersmooth, d}}{2\alpha_0},
\end{align*}
where inequality~$(\star)$ follows from Lemma~\ref{lemma:cvar-bound}.
Consequently, we have
$\inf_{\theta \in [0, 1]} \risk(\theta; P_0) = -
\frac{t\Delta_{\alpha_0, \holdersmooth, d}}{2\alpha_0}$ while
$\inf_{\theta \in [0, 1]} \risk(\theta; P_{tv}) = 0$.
Combining these infimal risk values, we see that for any
$\delta \le \frac{t\Delta_{\alpha_0, \holdersmooth, d}}{4\alpha_0}$,
$ \risk(\theta; P_0) - \inf_{\theta' \in \Theta} \risk(\theta'; P_0) = 
\frac{t\Delta_{\alpha_0, \holdersmooth, d}}{2\alpha_0} (1-\theta) \le \delta$ implies
\begin{equation*}
  \risk(\theta; P_{tv}) - \inf_{\theta' \in \Theta} \risk(\theta'; P_{tv})
  \ge  \theta \frac{t\Delta_{\alpha_0, \holdersmooth, d}}{2\alpha_0}
  \ge \frac{t\Delta_{\alpha_0, \holdersmooth, d}}{4\alpha_0}
  \ge \delta.
\end{equation*}
Similarly, for any $\delta \le \frac{t\Delta_{\alpha_0, \holdersmooth,
    d}}{4\alpha_0}$, $ \risk(\theta; P_{tv}) - \inf_{\theta' \in \Theta}
\risk(\theta'; P_{tv}) \le \delta$ implies $\theta \frac{t\Delta_{\alpha_0,
    \holdersmooth, d}}{2\alpha_0} \le \delta$ and hence
\begin{equation*}
  \risk(\theta; P_0) - \inf_{\theta' \in \Theta} \risk(\theta'; P_0)
  =   \frac{t\Delta_{\alpha_0, \holdersmooth, d}}{2\alpha_0} (1-\theta)
  \ge \frac{t\Delta_{\alpha_0, \holdersmooth, d}}{4\alpha_0}
  \ge \delta.
\end{equation*}
This is the desired separation~\eqref{eqn:dopt-sep}.

\subsubsection*{Closeness in variation distance}

It remains to bound the total
variation distance in Lemma~\ref{lemma:lecam}. For shorthand
in this section, let $\sigma^2 = \sigma^2(\holdersmooth, d)$.
Let $\rho_{\rm hel}$ be the
Hellinger affinity between two distributions
\begin{align*}
  \rho_{\rm hel}(P, Q) \defeq \int \sqrt{\frac{dQ}{dP}} dP,
\end{align*}
and define the following shorthand for the mixture distribution
\begin{equation*}
  \bar{P}_{1, t}  \defeq \frac{1}{2^m} \sum_{v \in \{\pm1\}^m} P_{tv}^n.
\end{equation*}
Le Cam's inequality bounds the total variation distance between the convex
hulls of $\mc{P}_0$ and $\mc{P}_1$
\begin{equation}
  \label{eqn:tv-hellinger}
  \inf\left\{\tvnorm{\bar{P}_0 - \bar{P}_1}:
    \bar{P}_1 \in \conv(\mc{P}_1^n) \right\}
  \le \tvnorm{P_0^n  - \bar{P}_{1, t}}
  \le \sqrt{2 (1-\rho_{\rm hel}(P_0^n, \bar{P}_{1, t}))}.
\end{equation}
Here, a key technical difficulty here is the mixture of the product
distributions $\bar{P}_{1, t}$.  In the rest of the proof, we bound
$\rho_{\rm hel}(P_0^n, \bar{P}_{1, t})$ following the approach
of~\citet{BirgeMa95}. As the distributions of $X$ are identical across
$P_0$ and $P_{tv}$ in our construction, the subsequent derivations subtly
differ from the original proof of~\citet{BirgeMa95}, as we must also
consider the conditional distribution $Y \mid X$, and we detail it below for
completeness.

Let $N = (N_1, \ldots, N_m)$ be a multinomial distribution counting the number
of observations $(X_i, Y_i)$ such that $X_i \in R_j$ for $j = 1, \ldots,
m$. To bound the Hellinger affinity $\rho_{\rm hel}(P_0^n, \bar{P}_{1, t})$,
we start with the fact that conditional on $N$, the likelihood ratio between
$P_{tv}^n$ and $P_0^n$ can be simplified.
% \begin{align*}
%   \frac{dP_{tv}}{dP_0}(x, y) = 1 + y t \sum_{j=1}^m v_j g_j(x).
% \end{align*}
We use the shorthand
\begin{equation}
  \label{eqn:tilted-distribution}
  dP_{\pm, j}(x, y) \defeq (1\pm y t g_j(x)) dP_{0, j}(x, y).
\end{equation}
Note that in our setting, we have bounded $y$ and
$|g_j| \le 1$, so that for small $t$, $P_{\pm, j}$ are valid distributions.
For any fixed $\mathfrak{n} = (n_1, \ldots, n_m) \ge 0$ such that
$\sum_{j=1}^m n_j = n$, denote
$(\bar{X}_{ij}, \bar{Y}_{ij}) \simiid \P_0(\cdot \mid X \in R_j) \eqdef \P_{0,
  j}$ for $i = 1, \ldots, n_j$. Notice that conditional on $N$
\begin{align*}
  \prod_{i=1}^n  \frac{dP_{tv}}{dP_0}(X_i, Y_i) \mid N = \mathfrak{n}
  ~~\eqd \prod_{j: n_j > 0}  \frac{dP_{v_j, j}^{n_j}}{dP_{0, j}^{n_j}}(\{X_i, Y_i\}_{i=1}^{n_j})
  \eqd \prod_{j: n_j > 0} \prod_{i=1}^{n_j}
  \left(1 + \bar{Y}_{ij} t v_j g_j(\bar{X}_{ij})\right)
\end{align*}
Writing $\{j: n_j > 0\} = \{j_1 < \ldots < j_s\}$ for convenience
\begin{align*}
  & \frac{1}{2^m} \sum_{v \in \{\pm 1\}^m} \prod_{j: n_j > 0} \prod_{i=1}^{n_j}
  \left(1 + \bar{Y}_{ij} t v_j g_j(\bar{X}_{ij})\right) \\
  & = \half \prod_{i=1}^{n_{j_1}}
    \left(1 + \bar{Y}_{ij_1} t v_{j_1} g_{j_1}(\bar{X}_{ij_1})\right)
    \cdot \frac{1}{2^{m-1}} \sum_{v: v_{j_1} = +1} \prod_{a = 2}^s \prod_{i=1}^{n_{j_a}}
    \left(1 + \bar{Y}_{ij_a} t v_{j_a} g_{j_a}(\bar{X}_{ij_a})\right) \\
  & \qquad + \half \prod_{i=1}^{n_{j_1}}
    \left(1 - \bar{Y}_{ij_1} t v_{j_1} g_{j_1}(\bar{X}_{ij_1})\right)
    \cdot \frac{1}{2^{m-1}} \sum_{v: v_{j_1} = -1} \prod_{a = 2}^s \prod_{i=1}^{n_{j_a}}
    \left(1 + \bar{Y}_{ij_a} t v_{j_a} g_{j_a}(\bar{X}_{ij_a})\right) \\
  & = \left\{
    \half \prod_{i=1}^{n_{j_1}}
    \left(1 + \bar{Y}_{ij_1} t v_{j_1} g_{j_1}(\bar{X}_{ij_1})\right)
    + \half \prod_{i=1}^{n_{j_1}}
    \left(1 - \bar{Y}_{ij_1} t v_{j_1} g_{j_1}(\bar{X}_{ij_1})\right) \right\}
    \cdot \frac{1}{2^{m-1}} \sum_{v: v_{j_1} = -1} \prod_{a = 2}^s \prod_{i=1}^{n_{j_a}}
    \left(1 + \bar{Y}_{ij_a} t v_{j_a} g_{j_a}(\bar{X}_{ij_a})\right) 
\end{align*}
noting that summands in the final term do not depend on $v_{j_1}$. Induct
through $a = 2, \ldots, s$ to conclude that the preceding display is equal to
\begin{align*}
  \prod_{a=1}^s \left\{
    \half \prod_{i=1}^{n_{j_a}}
    \left(1 + \bar{Y}_{ij_a} t v_{j_a} g_{j_a}(\bar{X}_{ij_a})\right)
    + \half \prod_{i=1}^{n_{j_a}}
    \left(1 - \bar{Y}_{ij_a} t v_{j_a} g_{j_a}(\bar{X}_{ij_a})\right) \right\} \\
  = \prod_{j: n_j  > 0} \left\{
    \half \prod_{i=1}^{n_{j}}
    \left(1 + \bar{Y}_{ij} t v_{j} g_{j}(\bar{X}_{ij})\right)
    + \half \prod_{i=1}^{n_{j}}
    \left(1 - \bar{Y}_{ij} t v_{j} g_{j}(\bar{X}_{ij})\right) \right\}.
\end{align*}
Thus
\begin{align*}
  & \E_{P_0^n}\left[\left( \frac{1}{2^m} \sum_{v \in \{\pm 1\}^m}
  \prod_{i=1}^n \frac{dP_{tv}}{P_0}(X_i, Y_i) \right)^{\half} \mid N = \mathfrak{n}\right] \\
  & = \E_{P_0^n}  \prod_{j: n_j  > 0} \left\{
    \half \prod_{i=1}^{n_{j}}
    \left(1 + \bar{Y}_{ij} t v_{j} g_{j}(\bar{X}_{ij})\right)
    + \half \prod_{i=1}^{n_{j}}
    \left(1 - \bar{Y}_{ij} t v_{j} g_{j}(\bar{X}_{ij})\right) \right\}^\half \\
  & = \prod_{j: n_j  > 0}     \E_{P_{0, j}^{n_j}}   \left\{
    \half \prod_{i=1}^{n_{j}}
    \left(1 + \bar{Y}_{ij} t v_{j} g_{j}(\bar{X}_{ij})\right)
    + \half \prod_{i=1}^{n_{j}}
    \left(1 - \bar{Y}_{ij} t v_{j} g_{j}(\bar{X}_{ij})\right) \right\}^\half  
   = \prod_{j: n_j  > 0}     \E_{P_{0, j}^{n_j}}   \left\{
    \half \frac{dP_{+, j}^{n_j}}{dP_{0, j}^{n_j}} + \half \frac{dP_{-, j}^{n_j}}{dP_{0, j}^{n_j}}
  \right\}^\half,
\end{align*}
where $dP_{\pm, j}$ are the tilted densities~\eqref{eqn:tilted-distribution}.

The following lemma, which we prove in Section~\ref{section:proof-birge},
controls the individual terms in the product.
(See also \citet[Lemma 2]{BirgeMa95}.)
\begin{lemma}
  \label{lemma:birge}
  Let $\sigma^2 = \sigma^2(\holdersmooth, d)$. Then for any $j$ such that
  $n_j > 0$,
  \begin{align*}
    \E_{P_{0, j}^{n_j}}   \left\{
    \half \frac{dP_{+, j}^{n_j}}{P_{0, j}^{n_j}} + \half \frac{dP_{-, j}^{n_j}}{P_{0, j}^{n_j}}
    \right\}^\half
    \ge 1 - \half \left[ (1+t^2 \sigma^2)^{n_j}
      + (1-t^2 \sigma^2)^{n_j} - 2\right].
  \end{align*}
\end{lemma}
\noindent Using $\prod_{j=1}^m (1-a_j) \ge 1- \sum_{j=1}^m a_j$ for any
$a_j \in [0, 1]$, the lemma gives
\begin{align*}
  \E_{P_0^n}\left[\left( \frac{1}{2^m} \sum_{v \in \{\pm 1\}^m}
  \prod_{i=1}^n \frac{dP_{tv}}{P_0}(X_i, Y_i) \right)^{\half} \mid N = \mathfrak{n}\right]
  \ge 1 - \half \sum_{j=1}^m \left[ (1+t^2 \sigma^2)^{n_j} + (1-t^2 \sigma^2)^{n_j} - 2\right].
\end{align*}
Taking expectations on both sides and using
$N_j \sim \bindist(n, \frac{1}{m})$, we get
\begin{align*}
  \rho_{\rm hel}\left( P_0^n, \frac{1}{2^m} \sum_{v \in \{\pm 1\}^m} P_{tv}^n \right)
  & \ge 1 - \half \sum_{j=1}^m
    \E\left[ (1+t^2 \tau_j^2)^{N_j} + (1-t^2 \tau_j^2)^{N_j} - 2\right] \\
  & = 1 - \half \sum_{j=1}^m
    \left[ \left(1+t^2 \frac{\sigma^2}{m} \right)^{n}
    + \left(1-t^2 \frac{\sigma^2}{m}\right)^{n} - 2\right],
\end{align*}
where the last line uses that if $N \sim \bindist(n, p)$ then
$\E[a^N] = ((1 - p) + pa)^n$.
Finally, an elementary calculation using that
$e^x \le 1 + x + x^2$ for $|x| \le 1$ and that $1 + x \le e^x$ for all $x$
shows that
\begin{equation*}
  \left(1+t^2 \frac{\sigma^2}{m} \right)^{n}
  + \left(1-t^2 \frac{\sigma^2}{m}\right)^{n}
  \le \exp\left(\frac{t^2 \sigma^2 n}{m}\right)
  + \exp\left(-\frac{t^2 \sigma^2 n}{m}\right)
  \le
  2 + 2 \left(\frac{t^2 \sigma^2 n}{m}\right)^2
\end{equation*}
whenever $\frac{nt^2 \sigma^2}{m} \le 1$,
and therefore
\begin{equation}
  \label{eqn:hellinger-nice-bound}
  \rho_{\rm hel}\left( P_0^n, \frac{1}{2^m} \sum_{v \in \{\pm 1\}^m} P_{tv}^n \right)
  \ge 1 -  \frac{t^4 n^2 \sigma^4}{m}~~\mbox{whenever}~~\frac{nt^2 \sigma^2}{m} \le 1.
\end{equation}

\subsubsection*{Finalizing the bound}

To show the result~\eqref{eqn:real-lower-bound}, it remains to choose the
separation parameter $t$ to be as large as possible while satisfying that
the mapping $x \mapsto \E_{P_{tv}}[Y \mid X = x] = -\frac{t
  \Delta_{\alpha_0, \holdersmooth, d}}{2\alpha_0} + t \sum_{j = 1}^m v_j
g_j(x)$ is $\beta$-H\"older in $x$ (i.e.\ in $\holderball{\holdersmooth}$)
and that the Hellinger affinity~\eqref{eqn:hellinger-nice-bound} is at least
a constant, which depends on the number of hyperrectangles $b$ via
definitions~\eqref{eqn:rectangles}--\eqref{eqn:gj-def}.

We begin with the H\"older condition. For shorthand, let
$h(x) = t \sum_{j = 1}^m v_j g_j(\cdot)$, omitting the dependence on $t$. We
claim that for any $b \in \N$, the choice $t = c(\holdersmooth)
d^{-\holdersmooth/2} b^{-\holdersmooth}$, where $c(\holdersmooth)$ depends
only on $\holdersmooth$, is sufficient to guarantee that $h \in
\holderball{\holdersmooth}$.
For simplicity assume $\holdersmooth \in \N$
(the calculation is similar but more tedious otherwise), so that $h \in
\holderball{\holdersmooth}$ is equivalent to the $\holdersmooth$th order
tensor $\nabla^\holdersmooth h(x)$ having operator norm at most $c(\holdersmooth)
d^{\holdersmooth/2}$. To that end, let $k \in \N$ and $I = (i_1, \ldots,
i_k) \subset [d]^k$, and let $U = \{u_j\}$ be a (non-repeated) list of the
indices appearing at least once in $I$, while $S_1, \ldots, S_{k'}$ are the
collections of unique indices in $I$ (e.g., if $i_1 = 1, \ldots, i_k = 1$,
then $S_1 = (1, \ldots, 1) \in \N^k$). The
$(i_1, \ldots, i_k)$ entry of the tensor $\nabla^k h(x)$ has the form
\begin{equation*}
  [\nabla^k h(x)]_{i_1, \ldots, i_k}
  = b^k \prod_{j = 1}^{|U|} \varphi^{(|S_j|)}(z^{u_j})
  \prod_{j \not (i_1, \ldots, i_k)} \varphi(z^j)
\end{equation*}
for some values $z^j \in [0, 1]$, by the
definition~\eqref{eqn:gj-def}. Setting $k = \holdersmooth$, the
construction~\eqref{eqn:bump} of the bumps $\varphi$ evidently guarantees that
each entry satisfies
$|[\nabla^\holdersmooth h(x)]_{I}| \le c(\holdersmooth) b^\holdersmooth$ for
$c(\holdersmooth) = O(\holdersmooth!)$.  Making the observation that for an
$r$th order tensor $T$ on $(\R^d)^{\otimes r}$ with entries $\linf{T} \le C$
we have $\opnorm{T} \le C d^{r/2}$, we see that
$\opnorm{\nabla^\holdersmooth h(x)} \le c(\holdersmooth) b^\holdersmooth
d^{\holdersmooth/2}$.  Thus, there are
$N(\alpha_0, \holdersmooth, d)$ and $c(\holdersmooth, d)$ such that for
$n \ge N(\alpha_0, \holdersmooth, d)$, choosing
$t = c(\holdersmooth, d) b^{-\holdersmooth}$ guarantees
$x \mapsto \E_{tv}[Y \mid X = x] \in \holderball{\holdersmooth}$.

It remains to choose $b$ so that the Hellinger
affinity~\eqref{eqn:hellinger-nice-bound} is at least $\frac{3}{4}$, so that
we can apply Le Cam's method (Lemma~\ref{lemma:lecam}). For this,
we require that $\frac{t^4 n^2 \sigma^4}{m} \le \frac{1}{4}$ (which certainly
implies that $\frac{n t^2 \sigma^2}{m} \le 1$). Substituting the
$t = c b^{-\holdersmooth}$ and recalling that $m = b^d$, we see
that it is sufficient that
$\frac{n^2 c^2 b^{-2 \holdersmooth} \sigma^4}{b^d} \le \frac{1}{4}$, i.e.,
$b^{d + 2 \holdersmooth} \ge 4 n^2 \sigma^4$, so that the choice
$b = \ceil{(2 n c \sigma^2)^\frac{2}{d + 2 \holdersmooth}}$ suffices. 
Using the bound~\eqref{eqn:hellinger-nice-bound}, we
conclude
$\helaff(P_0^n, \frac{1}{2^m} \sum_{v \in \{\pm 1\}^m} P_{tv}^n) \ge
\frac{3}{4}$. By combining the separation~\eqref{eqn:dopt-sep} with Le Cam's
convex hull method (Lemma~\ref{lemma:lecam}) and the
bound~\eqref{eqn:tv-hellinger} relating variation distance to Hellinger
affinity, we obtain
\begin{equation*}
  \mc{M}_n  \ge c'(\holdersmooth, d)
  \frac{\Delta_{\alpha_0, \holdersmooth, d}}{\alpha_0}
  (\sigma^2 n)^{-\frac{2\holdersmooth}{2\holdersmooth + d}}
\end{equation*}
for some factor $c'(\holdersmooth, d)$ and all suitably large $n$.

\subsubsection{Proof of Lemma~\ref{lemma:lecam}}
\label{section:proof-lecam}

For any $P_v \in \mc{P}_v$, $v\in \{0, 1\}$, we have
\begin{align*}
  \sup_{P \in \mathfrak{P}_{\holdersmooth}} \E_{P^n}\left[
  \risk(\what{\theta}; P) - \inf_{\theta \in \Theta} \risk(\theta; P) \right]
  \ge \frac{1}{2} \left(
  \E_{P_0^n}\left[
  \risk(\what{\theta}; P_0) - \inf_{\theta \in \Theta} \risk(\theta; P_0) \right]
  + \E_{P_1^n}\left[
  \risk(\what{\theta}; P_1) - \inf_{\theta \in \Theta} \risk(\theta; P_1) \right]
    \right).
\end{align*}
For $v \in \{0, 1\}$, if we define
\begin{equation*}
  \lambda_v(\what{\theta}) \defeq \frac{1}{2\delta} \inf_{P_v \in \mc{P}_v} \left\{
    \risk(\what{\theta}; P_v) - \inf_{\theta \in \Theta} \risk(\what{\theta}; P_v) 
  \right\},
\end{equation*}
we have the following lower bound on the first display.
\begin{align*}
  \delta \cdot \sup_{P_v^n \in \mc{P}_v^n, v \in \{0, 1\}} \left\{
    \E_{P_0^n} \lambda_0(\what{\theta})
    + \E_{P_1^n} \lambda_1(\what{\theta}) \right\}.
\end{align*}
Since this supremum problem is linear in $P_v^n$, we may replace it with a
supremum over the convex hull spanned by $\mc{P}_v^n$.
\begin{equation}
  \label{eqn:linear-programs}
  \delta \cdot \sup_{\bar{P}_v^n \in \conv(\mc{P}_v^n), v \in \{0, 1\}} \left\{
    \E_{\bar{P}_0^n} \lambda_0(\what{\theta})
    + \E_{\bar{P}_1^n} \lambda_1(\what{\theta}) \right\}
\end{equation}

By the definition of $\distopt{P_0}{P_1}$, we have
\begin{equation*}
  \risk(\what{\theta}; P_0) - \inf_{\theta \in \Theta} \risk(\theta; P_0)
  + \risk(\what{\theta}; P_1) - \inf_{\theta \in \Theta} \risk(\theta; P_1)
  \ge 2 \delta
\end{equation*}
for all $P_v \in \mc{P}_v$, $v \in \{0, 1\}$. Taking infimum over
$P_v \in \mc{P}_v$, conclude
$\lambda_0(\what{\theta}) + \lambda_1(\what{\theta}) \ge 1$ almost surely.
From the variational representation of the total variation distance
\begin{equation*}
  1  - \tvnorm{Q -P}  = \inf_{f_0 + f_1 \ge 1} \{ \E_{Q} f_0 + \E_{P} f_1\},
\end{equation*}
we have
\begin{equation*}
  \E_{\bar{P}_0^n} \lambda_0(\what{\theta})
  + \E_{\bar{P}_1^n} \lambda_1(\what{\theta})
  \ge 1 - \tvnorm{\bar{P}_0^n - \bar{P}_1^n}
\end{equation*}
for any $\bar{P}_0^n$ and $\bar{P}_1^n$. Using this to lower bound
the expression~\eqref{eqn:linear-programs} gives our result.

\subsubsection{Proof of Lemma~\ref{lemma:cvar-bound}}
\label{section:proof-cvar-bound}

We construct a particular distribution $\pworst$ taking the form
$d\pworst(x) = L(x) dP_0(x)$, where $L$ is a likelihood ratio we construct
as
\begin{align*}
  L(x) \defeq \frac{1}{\alpha_0} \indic{\sum_{j=1}^m v_j g_j(x) \ge q_{1-\alpha_0}}.
\end{align*}
To see that $\E[L(X)] = 1$, note that the disjointness of $R_j$ yields
\begin{align*}
  \int_{[0, 1]^d} \indic{\sum_{j=1}^m v_j g_j(x) \ge q_{1-\alpha_0}} dx
  = \sum_{j=1}^m \int_{R_j} \indic{ v_j g_j(x) \ge q_{1-\alpha_0}} dx.
\end{align*}
Using the change of variables
$b(x^k - \frac{l_{jk} -1 }{b}) \mapsto x^k$ in the final
display and recalling $m \defeq b^d$, we replace $g_j$ with
$g(x) = \prod_{k = 1}^d \varphi(x^k)$ and find
\begin{align*}
  & \frac{1}{m} \sum_{j=1}^m
    \int_{[0, 1]^d} \indic{ v_j g(x) \ge q_{1-\alpha_0}} dx \\
  & = \frac{1}{m} \sum_{v_j = +1}
  \int_{[0, 1]^d} \indic{g(x) \ge q_{1-\alpha_0}} dx
  + \frac{1}{m} \sum_{v_j = -1}
    \int_{[0, 1]^d} \indic{g(x) \le - q_{1-\alpha_0}} dx \\
  & = \frac{1}{m} \sum_{v_j = +1}
  \int_{[0, 1]^d} \indic{g(x) \ge q_{1-\alpha_0}} dx
  + \frac{1}{m} \sum_{v_j = -1}
  \int_{[0, 1]^d} \indic{g(\onevec - x) \le - q_{1-\alpha_0}} dx,
\end{align*}
where in the last equality, we used the change of variables
$x^k \mapsto 1-x^k$ when $v_j = -1$. As $d$ is
odd,
\begin{align*}
  g(\onevec - x) = \prod_{k=1}^d \varphi(1-x^k)
  = - \prod_{k=1}^d \varphi(x^k) = -g(x).
\end{align*}
This implies 
$\int_{[0, 1]^d} \indic{g(\onevec-x) \le -q_{1-\alpha_0}} dx = \int_{[0, 1]^d}
\indic{-g(x) \le - q_{1-\alpha_0}} dx = \alpha_0$ by the continuity of
$\varphi$. We conclude $\E[L(X)] = 1$.

We now show that
the choice $d\pworst = L dP_0$ satisfies the conclusion of the lemma.
As $\linfstatnorm{L} \le \alpha_0^{-1}$, it is sufficient to show that
$\E[L(X) \sum_{j=1}^m v_j g_j(X)] = \frac{\Delta_{\alpha_0, \holdersmooth, d}}{\alpha_0}$.
Indeed, we have
\begin{align*}
  & \int_{[0, 1]^d} \sum_{j=1}^m v_j g_j(x)
  \indic{\sum_{j=1}^m v_j g_j(x) \ge q_{1-\alpha_0}} dx \\
  & = \sum_{j=1}^m \int_{R_j} v_j g_j(x) \indic{ v_j g_j(x) \ge q_{1-\alpha_0}} dx 
  = \frac{1}{m} \sum_{j=1}^m
    \int_{[0, 1]^d} v_j g(x) \indic{ v_j g(x) \ge q_{1-\alpha_0}} dx \\
  & =  \frac{1}{m} \sum_{v_j = +1}
  \int_{[0, 1]^d} g(x) \indic{g(x) \ge q_{1-\alpha_0}} dx
  - \frac{1}{m} \sum_{v_j = -1}
    \int_{[0, 1]^d} g(x) \indic{g(x) \le - q_{1-\alpha_0}} dx
\end{align*}
where we used the change of variables
$b(x^k - \frac{l_{jk} -1 }{b}) \mapsto x^k$ in the final
equality. When $v_j = -1$, again use the change of variables
$x^k \mapsto 1-x^k$ and use $g(\onevec - x) = -g(x)$ to arrive at
\begin{align*}
  & \frac{1}{m} \sum_{v_j = +1}
  \int_{[0, 1]^d} g(x) \indic{g(x) \ge q_{1-\alpha_0}} dx
    - \frac{1}{m} \sum_{v_j = -1}
  \int_{[0, 1]^d} g(\onevec - x) \indic{g(\onevec - x) \le - q_{1-\alpha_0}} dx \\
  & =   \frac{1}{m} \sum_{v_j = +1}
  \int_{[0, 1]^d} g(x) \indic{g(x) \ge q_{1-\alpha_0}} dx
  + \frac{1}{m} \sum_{v_j = -1}
  \int_{[0, 1]^d} g(x) \indic{g(x) \ge q_{1-\alpha_0}} dx = \Delta_{\alpha_0, \holdersmooth, d}.
\end{align*}

\subsubsection{Proof of Lemma~\ref{lemma:birge}}
\label{section:proof-birge}

We begin with the simple observation that $\sigma^2 \equiv
\sigma^2(\holdersmooth, d)
= \E_{P_{0,j}}[g_j^2(\bar{X}_{ij})]$:
we have
$\E_{P_{0,j}}[g_j^2(\bar{X}_{ij})]
= \E[g_j^2(X) \mid X \in R_j]
= (b \int_0^{1/b} \varphi^2(bx) dx)^d
= (\int_0^1 \varphi^2(x) dx)^d$. Now,
denoting $n = n_j$ for simplicity, odd terms cancel to give
\begin{align*}
  L_{n, j} & \defeq \half \prod_{i=1}^{n}
  \left(1 + \bar{Y}_{ij} t v_{j} g_{j}(\bar{X}_{ij})\right)
  + \half \prod_{i=1}^{n}
  \left(1 - \bar{Y}_{ij} t v_{j} g_{j}(\bar{X}_{ij})\right) \\
  & = 1 + \underbrace{
    \sum_{k \in 2\N, 2 \le k\le n} t^k \sum_{i_1 < \cdots < i_k}
    \bar{Y}_{i_1j} g_j(\bar{X}_{i_1j}) \cdots \bar{Y}_{i_kj} g_j(\bar{X}_{i_kj})
  }_{\eqdef Z_{n, j}}.
\end{align*}
As $\sqrt{1+y} \ge 1+ \frac{y}{2} - \frac{y^2}{2}$ if $y \ge -1$, we have
\begin{equation*}
  \sqrt{L_{n, j}} \ge 1 + \half Z_{n, j} - \half Z_{n, j}^2. 
\end{equation*}
Taking expectations and noting $\E_{P_{0, j}^{n}} Z_{n, j} = 0$ as
$\E_{P_{0, j}}[\bar{Y}_{ij} \mid \bar{X}_{ij}] = 0$,  we get
\begin{align*}
  \E_{P_{0,j}^n} \sqrt{L_{n, j}}
  & \ge 1 - \half \E _{P_{0, j}^n} \left[ Z_{n, j}^2\right]
    = 1 - \half \sum_{k \in 2\N, 2 \le k \le n} t^{2k} \sum_{i_1 < \cdots < i_k}
    \E_{P_{0, j}^n}[ g_j(\bar{X}_{i_1j})^2\cdots g_j(\bar{X}_{i_kj})^2] \\
  & = 1 - \half \sum_{k \in 2\N, 2 \le k \le n} \choose{n}{k} t^{2k} \sigma^{2k},
\end{align*}
where we used the fact that $\bar{X}_{ij}$'s are i.i.d. in the final equality.
Apply the binomial theorem to conclude
\begin{equation*}
  \sum_{k \in 2\N, 2 \le k \le n} \choose{n}{k} t^{2k} \sigma^{2k}
  = (1+t^2 \sigma^2)^{n} + (1-t^2 \sigma^2)^{n} - 2.
\end{equation*}

\subsection{Proof of Corollary~\ref{cor:lower-bound-higher-order}}
\label{sec:proof-lower-bound-higher-order}

Recalling that
$\risk_{\kdual}(\theta; P) \ge \risk(\theta; P)$ for any $p > 1$,
while for the construction~\eqref{eqn:lower-P0-dist}
$\risk_{\kdual}(\theta; P_0) = \risk(\theta; P_0)$ because
$\E_{P_0}[Y \mid X]$ is constant,
we simply recognize that the
optimization distance $\distopt{\cdot}{\cdot}$ is larger for
$\risk_{\kdual}$. The proof of Theorem~\ref{theorem:lower-bound}
then gives the result.

\subsection{Proof of Lemma~\ref{lemma:conf}}
\label{section:proof-conf}

First, note that variational form for the $L^\kdual(P)$-norm gives
\begin{align}
  &     \left( \E_{(X, C) \sim \pdist_{X,C}} \left[ \hinge{\confcondrisk-\eta}^\kdual
    \right]\right)^{\kdualinv}
    \label{eq:var-form-conf} \\
  & = \sup_{h} \left\{
    \E[\bar{h}(X, C) (\lossvar - \eta)]:
    ~\bar{h}: \mc{X} \times \mc{C} \to \R~\mbox{measurable},~\bar{h} \ge 0,
    ~\E[\bar{h}(X, C)^\kexp] \le 1
    \right\}. \nonumber
\end{align}
For ease of notation, let
\begin{equation*}
  e(x) \defeq \E[\loss(\param; (X,Y)) \mid X = x],
  ~~ e(x, c) \defeq \E[\loss(\param; (X,Y)) \mid X = x, C = c].
\end{equation*}

We first show the equality~\eqref{eq:resbound}. To see that ``$\ge$''
direction holds, let
$\epsilon \defeq \left( \E\hinge{e(X, C) - \eta}^\kdual \right)^{\kexpinv}$.
Then, we have
\begin{align}
  \confpopobjshort
  &  \le \sup_{h, f~{\scriptsize \mbox{measurable}}}\Bigg\{
    \E\left[(h(X) + f(X, C)) (\lossvar - \eta)\right]:
    h + f \ge 0,~\E[(h(X) + f(X, C))^\kexp] \le 1\Bigg\} \nonumber \\
  & = \left( \E \hinge{\confcondrisk-\eta}^\kdual \right)^{\kdualinv}
    \label{eq:ub-conf}
\end{align}
where we used the variational form~\eqref{eq:var-form-conf} in the last
inequality.

For the ``$\le$'' inequality, fix an arbitrary $\epsilon > 0$.  If
$\left( \E\hinge{e(X, C) - \eta}^\kdual \right)^{1/\kexp} \le \epsilon$, then
the bound follows. Otherwise, consider
$\left( \E\hinge{e(X, C) - \eta}^\kdual \right)^{1/\kexp} > \epsilon >
0$. Note that the supremum in the variational form~\eqref{eq:var-form-conf} is
attained by
\begin{equation*}
  \bar{h}\opt(x, c) \defeq
  \frac{\hinge{e(x, c) - \eta}^{\kdual-1}}{\left( \E\hinge{e(X, C) - \eta}^\kdual
    \right)^{\kexpinv}}.
\end{equation*}
Now, define
\begin{align*}
  h\opt(x)
  & \defeq
    \frac{\hinge{e(x) - \eta}^{\kdual-1}}
    {\left( \E\hinge{e(X, C) - \eta}^\kdual \right)^{\kexpinv}}, \\
  f\opt(x, c)
  & \defeq
    \frac{1}{\left( \E\hinge{e(X, C) - \eta}^\kdual \right)^{\kexpinv}}
    \left(
    \hinge{e(x, c) - \eta}^{\kdual-1} -
    \hinge{e(x) - \eta}^{\kdual-1} \right)
\end{align*}
so that $\bar{h}\opt = h\opt + f\opt$. Since $\epsilon h\opt \in \holderbound$
and
$\norm{f\opt(X, C)}_{L^{\infty}(P)} \le
\frac{\confthresh^{\kdual-1}}{\epsilon}$, $h\opt$ and $f\opt$ are in the
feasible region of the maximization problem that defines $\confpopobjshort$. We
conclude that
\begin{align*}
  & \left( \E \hinge{\confcondrisk-\eta}^\kdual \right)^{\kdualinv} \\
  & \le \inf_{\epsilon \ge 0}
    \Bigg\{
    \epsilon \vee
    \sup_{h, f~{\scriptsize \mbox{meas.}}}\Big\{
    \E\left[(h(X) + f(X, C)) (\lossvar - \eta)\right]: \nonumber \\
  & \hspace{50pt} h + f \ge 0, ~\E[(h(X) + f(X, C))^\kexp] \le 1,
    ~\epsilon h \in \holderbound,
    ~\norm{f(X, C)}_{L^{\infty}(P)} \le \frac{\confthresh^{\kdual-1}}{\epsilon}
    \Big\} \Bigg\}. \nonumber 
\end{align*}
Rescaling the supremum problem by $1/\epsilon$, we obtain the first
result~\eqref{eq:resbound}.

To show the second result, fix an arbitrary $\epsilon > 0$. If
$\left( \E\hinge{e(X, C) - \eta}^\kdual \right)^{\kexpinv} \le \epsilon$, then
from our upper bound~\eqref{eq:ub-conf}
\begin{align*}
  (\confpopobjshort \vee \epsilon^{\kexp-1}) - \epsilon^{\kexp-1}
  \le 0
\end{align*}
so that our desired result trivially holds. On the other hand, if
$\left( \E\hinge{e(X, C) - \eta}^\kdual \right)^{\kexpinv} > \epsilon$, then
\begin{align*}
  \confpopobjshort =
  \left( \E_{(X, C) \sim p_{X, C}} \left[ \hinge{\confcondrisk-\eta}^\kdual
  \right]\right)^{\kdualinv}
\end{align*}
from our argument above, so the desired result again holds.

\subsection{Proof of Proposition~\ref{prop:fast-rate-ub-conf}}
\label{section:fast-conf}

We proceed similarly as in the proof of Proposition~\ref{prop:fast-rate-ub}.
Letting
\begin{equation*}
  \confsepobj \defeq (\lossvar - \eta)\hinge{\confcondrisk - \eta}^{\kdual-1},
\end{equation*}
rewrite the $L^\kdual$-norm as 
\begin{align*}
  \left( \E\hinge{\confcondrisk - \eta}^\kdual\right)^{\kdualinv}
  & = \frac{\E\hinge{\confcondrisk - \eta}^\kdual}
    {  \left( \E\hinge{\confcondrisk - \eta}^\kdual\right)^{\kexpinv}}  \\
  & = \frac{\E[\confsepobj]}{\left( \E\hinge{\confcondrisk - \eta}^\kdual\right)^{\kexpinv}}.
\end{align*}
% Now, note that $\eta \mapsto \confsepobj$ is monotone. Since monotone
% functions concentrate by standard covering arguments (e.g.~\citep[Theorem
% 2.7.5]{VanDerVaartWe96}), we have
Since $(\theta, \eta) \mapsto \confsepobj$ is $\kdual \zbound$-Lipschitz, we
again get from a standard bracketing number argument for uniform concentration
of Lipschitz functions~\cite[Theorem 2.7.11]{VanDerVaartWe96}
\begin{align}
  \label{eq:lip-concentration}
  \sup_{\eta \in [0, \zbound]}
  \left|
  \E[\confsepobj] - \E_{\emp}[\confsepobj]
  \right|
  \le c_1 \zbound^2  \sqrt{\frac{\log\frac{1}{\gamma}}{n}}
\end{align}
with probability at least $1-\gamma$, where $c_1$ is some universal constant.
Hence, with probability at least $1-\gamma$, for all
$\theta \in \Theta, \eta \in [0, \zbound]$
\begin{align}
  \left( \E\hinge{\confcondrisk - \eta}^\kdual\right)^{\kdualinv}
  & \le \left( \E\hinge{\confcondrisk - \eta}^\kdual\right)^{-\kexpinv}
    \E_{\emp}[\confsepobj] \nonumber \\
  & \qquad + \left( \E\hinge{\confcondrisk - \eta}^\kdual\right)^{-\kexpinv}
    c_1 \zbound^2 \sqrt{\frac{\log\frac{1}{\gamma}}{n}}.
    \label{eq:conf-right-concentration}
\end{align}

Next, we upper bound $\E_{\emp}[\confsepobj]$ by our empirical objective
$\confempobjshort$. To this end, uniform concentration of Lipschitz
functions~\cite[Theorem 2.7.11]{VanDerVaartWe96} again yields
\begin{equation}
  \label{eq:lip-concentration-moment}
  \E_{\emp}\hinge{\confcondrisk - \eta}^{\kdual-1}
  \le \E\hinge{\confcondrisk - \eta}^{\kdual-1}
  + c_{2} \zbound^2 \sqrt{\frac{\log\frac{1}{\gamma}}{n}}
\end{equation}
for all $\theta \in \Theta, \eta \in [0, \zbound]$, with probability at least
$1-\gamma$.  Define the functions
\begin{align*}
  h_{\eta}\opt(x)
  & \defeq
    \frac{\hinge{e(x) - \eta}^{\kdual-1}}{\left( \E\hinge{e(X, C) - \eta}^\kdual \right)^{\kexpinv}}, \\
  f\opt(x, c)
  & \defeq
    \frac{1}{\left( \E\hinge{e(X, C) - \eta}^\kdual \right)^{\kexpinv}}
    \left(
    \hinge{e(x, c) - \eta}^{\kdual-1} -
    \hinge{e(x) - \eta}^{\kdual-1} \right),
\end{align*}
and note that
\begin{equation}
  \label{eq:conf-second-mom-hopt}
  \E_{\emp}[(h_{\eta}\opt(X) + f\opt(X, C))^\kexp] \le 1 + c_{2} \zbound^2
  \left( \E\hinge{\confcondrisk - \eta}^\kdual\right)^{-\kexpinv}
  \sqrt{\frac{\log\frac{1}{\gamma}}{n}}.
\end{equation}
with probability at least $1-\gamma$. 

Since our desired bound holds trivially if
$ \left( \E\hinge{\confcondrisk - \eta}^\kdual\right)^{\kexpinv} \le
\epsilon$, we now assume that
$ \left( \E\hinge{\confcondrisk - \eta}^\kdual\right)^{\kexpinv} \ge
\epsilon$.  Since $\epsilon h_{\eta}\opt(x) \in \holderbound$, we have
\begin{align*}
  & \left( \E\hinge{\confcondrisk - \eta}^\kdual\right)^{-\kexpinv}
    \E_{\emp}[\confsepobj] \\
  & = (1+\tau_n(\gamma, \epsilon))^{\kexpinv}
  \E_{\emp}\left[ \frac{h_{\eta}\opt(X) + f\opt(X, C)}{(1+\tau_n(\gamma, \epsilon))^{\kexpinv}}
  (\lossvar - \eta) \right] \\
  & \le   (1+\tau_n(\gamma, \epsilon))^{\kexpinv} \sup_{h \in \mc{H}_{L_n(\gamma), n},
    f \in \mc{F}_{\delta_n(\gamma), \kdual, n}}
    \Bigg\{ \E_{\emp}\left[ \frac{h(X) + f(X, C)}{\epsilon}
    (\lossvar - \eta)\right]
    ~~\Bigg|~~ \\
    & \hspace{220pt} \E_{\emp}[(h(X) + f(X, C))^\kexp] \le \epsilon^\kexp
    \Bigg\}
\end{align*}
with probability at least $1-\gamma$, where we used the
bound~\eqref{eq:conf-second-mom-hopt} in the second inequality. Combining the
preceding display with the bound~\eqref{eq:conf-right-concentration}, with
probability at least $1-2\gamma$,
\begin{align*}
  \left( \E\hinge{\confcondrisk - \eta}^\kdual\right)^{\kdualinv}
  \le (1+\tau_n(\gamma, \epsilon))^{\kexpinv}~
  \confempupperbd
  + \frac{c_1\zbound^2}{\epsilon^{\kexp-1}} \sqrt{\frac{\log\frac{1}{\gamma}}{n}}.
\end{align*}
for all $\theta \in \Theta, \eta \in [0, \zbound]$.

\subsection{Proof of Lemma~\ref{lemma:conf-empirical-dual}}
\label{section:proof-conf-empirical-dual}

We take the dual of the optimization problem
\begin{align*}
  & \maximize_{h, f \in \R^n}
    \frac{1}{n} \sum_{i=1}^n \frac{h_i + f_i}{\epsilon} (\lossitem - \eta) \\
  & \subjectto ~~h_i + f_i \ge 0~~\mbox{for all}~~i \in [n], 
    ~~\frac{1}{n} \sum_{i=1}^n (h_i + f_i)^\kexp \le \epsilon^\kexp,\\
  & \hspace{45pt}
    ~~h_i - h_j \le L^{\kdual-1} \norm{X_i - X_j}^{\kdual-1}~~\mbox{for all}~~i,j \in [n], \\
  & \hspace{50pt} |f_i| \le \delta^{\kdual-1}~~\mbox{for all}~~i \in [n]
\end{align*}
where $h_i \defeq h(X_i)$ and $f_i = f(X_i, C_i)$. To ease notation, we do a
change of variables $h_i \gets \frac{h_i}{\epsilon}$,
$f_i \gets \frac{f_i}{\epsilon}$ and $q_i \gets h_i + f_i$ which gives
\begin{align}
  \label{eq:primal-again}
  & \maximize_{q, h \in \R^n}
    \frac{1}{n} \sum_{i=1}^n q_i (\lossitem - \eta) \\
  & \subjectto ~~q_i \ge 0
    ~~\mbox{for all}~~i \in [n], 
    ~~\frac{1}{n} \sum_{i=1}^n q_i^\kexp \le 1, \nonumber \\
  & \hspace{45pt}
    ~~h_i - h_j
    \le \frac{L^{\kdual-1}}{\epsilon} \norm{X_i - X_j}^{\kdual-1}~~\mbox{for all}~~i,j \in [n],
    \nonumber
    \\
  & \hspace{50pt} |q_i - h_i| \le \frac{\delta^{\kdual-1}}{\epsilon}
    ~~\mbox{for all}~~i \in [n].
\end{align}
For $\gamma \in \R^n_+$, $\lambda \ge 0$, $B \in \R^{n \times n}_+$,
$\xi^+, \xi^- \in \R^n_+$, the associated Lagrangian is
\begin{align*}
  \mc{L}(q, h, \gamma, \lambda, B, \xi^+, \xi^-)
  & \defeq
    \frac{1}{n} \sum_{i=1}^n q_i (\lossitem-\eta)
    + \frac{1}{n} \gamma^\top q
    + \frac{\lambda}{2} \left( 1- \frac{1}{n} \sum_{i=1}^n q_i^\kexp\right) \\
  & \qquad  + \frac{1}{n^2}
    \left( \frac{L^{\kdual-1}}{\epsilon} \tr(B^\top D) - h^\top
    (B\onevec - B^\top \onevec)
    \right) \\
  & \qquad + \frac{\xi^{+\top}}{n}
    \left(  \frac{\confthresh^{\kdual-1}}{\epsilon} \onevec
    - (q-h) \right)
    + \frac{\xi^{-\top}}{n}
    \left(  \frac{\confthresh^{\kdual-1}}{\epsilon} \onevec
    + (q-h) \right)
\end{align*}
where $D \in \R^{n \times n}$ is a matrix with entries
$D_{ij} = \norm{X_i - X_j}^{\kdual-1}$. From strong duality, the primal
optimal value~\eqref{eq:primal-again} is
$\inf_{\gamma \in \R^n_+, \lambda \ge 0, B \in \R^{n \times n}_+,
  \xi^+, \xi^- \in \R^n_+} \sup_{q, h}
\mc{L}(q, h, \gamma, \lambda, B, \xi^+, \xi^-)$.

The first order conditions for the inner supremum give
\begin{align*}
  q\opt_i \defeq \frac{1}{n \lambda} \left(
  \lossitem-\eta + n\gamma - \xi^+_i + \xi^-_i
            \right),~~~
  \frac{1}{n} (B\onevec - B^\top \onevec)
           = \left( \xi^+ -
            \xi^-\right).
\end{align*}
By nonnegativity of $B$ and $\xi^+, \xi^-$, the second equality implies
that $\xi^+ = \frac{1}{n} B\onevec$ and
$\xi^- = \frac{1}{n} B^\top \onevec$. Substituting these values and
infimizing out $\lambda, \gamma \ge 0$ as in Lemma~\ref{lemma:empirical-dual},
we obtain
\begin{align*}
  \inf_{\lambda \ge 0, \gamma \in \R^n_+}
  \sup_{q, h} \mc{L}(q,, h, \gamma, \lambda, B, \xi^+, \xi^-)
  & =  \bigg( \frac{\kdual-1}{n} \sum_{i=1}^n
      \hingeBig{ \loss(\param; (X_i,Y_i))
      - \frac{1}{n} \sum_{j=1}^n (B_{ij} - B_{ji}) -\eta}^\kdual
  \bigg)^{\kdualinv} \\
& \hspace{40pt}
  + \frac{L^{\kdual-1}}{\epsilon n^2} \sum_{i,j = 1}^n \norm{X_i - X_j} B_{ij}
  + \frac{2 \confthresh^{\kdual-1}}{\epsilon n^2} \sum_{i, j = 1}^n |B_{ij}|.
\end{align*}
Taking the infimum with respect to $B \in \R^{n\times n}_+$
gives the lemma.

%%% Local Variables:
%%% mode: latex
%%% TeX-master: "main.tex"
%%% End:

\else
\newpage
\appendix

\fi

\end{document}